\providecommand{\llamaindexbranded}{}
\documentclass[10pt,twocolumn,letterpaper]{article}

\usepackage[margin=1in]{geometry}

\usepackage[utf8]{inputenc}
\usepackage[T1]{fontenc}
\usepackage{times}

\usepackage{amsmath,amssymb,amsthm}

\usepackage{graphicx}
\graphicspath{{figures/}}
\usepackage{subcaption}
\usepackage{capt-of}   %

\usepackage[normalem]{ulem}   %
\usepackage{booktabs}
\usepackage{tabularx}
\newcolumntype{L}[1]{>{\raggedright\arraybackslash\hspace{0pt}}X}
\usepackage{colortbl}  %
\usepackage{longtable}   %
\usepackage{multirow}
\usepackage{makecell}
\usepackage{enumitem}
\usepackage{numprint}
\npstyleenglish
\usepackage{etoc}

\usepackage{algorithm}
\usepackage{algorithmic}

\usepackage{xcolor}
\ifdefined\cleanbuildenabled
\newcommand{\boyang}[1]{}
\newcommand{\simon}[1]{}
\newcommand{\adrian}[1]{}
\newcommand{\todo}[1]{}
\newcommand{\eli}[1]{}
\newcommand{\zhaoqi}[1]{}
\newcommand{\antonio}[1]{}
\else
\newcommand{\boyang}[1]{{\color{red}\textbf{[Boyang:} #1\textbf{]}}}
\newcommand{\simon}[1]{{\color{blue}\textbf{[Simon:} #1\textbf{]}}}
\newcommand{\adrian}[1]{{\color{teal}\textbf{[Adrian:} #1\textbf{]}}}
\newcommand{\todo}[1]{{\color{orange}\textbf{[TODO:} #1\textbf{]}}}
\newcommand{\eli}[1]{{\color{violet}\textbf{[Eli:} #1\textbf{]}}}
\newcommand{\zhaoqi}[1]{{\color{magenta}\textbf{[Zhaoqi:} #1\textbf{]}}}
\newcommand{\antonio}[1]{{\color{green}\textbf{[Antonio:} #1\textbf{]}}}
\fi

\newcommand{\cmark}{$\checkmark$}

\newcommand{\pmark}{$\circ$}
\newcommand{\yes}{\ensuremath{\bullet}}
\newcommand{\half}{\ensuremath{\circ}}
\newcommand{\no}{{}}
\definecolor{oursband}{HTML}{E8F1FB}
\definecolor{oursink}{HTML}{2B6CB8}
\newcommand{\yesours}{{\color{oursink}\ensuremath{\bullet}}}
\definecolor{resultdroplight}{HTML}{FCE8E6}
\definecolor{resultdropmedium}{HTML}{F6C7C3}
\definecolor{resultdropstrong}{HTML}{EFA8A3}
\definecolor{qualok}{HTML}{14843E}
\definecolor{qualpartial}{HTML}{DE8A0C}
\definecolor{qualbad}{HTML}{C62828}
\definecolor{qualnone}{HTML}{969696}

\usepackage[colorlinks=true,linkcolor=blue,citecolor=blue,urlcolor=blue]{hyperref}
\usepackage[nameinlink,capitalize]{cleveref}
\ifdefined\llamaindexbranded
  \usepackage{llamapaper}
\fi

\usepackage[numbers,sort&compress]{natbib}
\usepackage{xfp}

\newcommand{\statnum}[1]{\numprint{#1}}

\newcommand{\shortdocs}{68}
\newcommand{\shortpages}{244}

\newcommand{\mediumdocs}{64}
\newcommand{\mediumpages}{1682}

\newcommand{\longdocs}{17}
\newcommand{\longpages}{1522}

\newcommand{\autodocs}{14}
\newcommand{\autopages}{193}

\newcommand{\formdocs}{169}
\newcommand{\formpages}{594}

\newcommand{\formtypes}{17}

\newcommand{\formreviewed}{13}

\newcommand{\formeligiblerules}{16534}
\newcommand{\formbboxev}{13867}

\newcommand{\corruptdocs}{38}
\newcommand{\corruptpages}{634}
\newcommand{\corruptreal}{30}
\newcommand{\corruptsynth}{8}

\newcommand{\evaldocs}{\fpeval{\shortdocs+\mediumdocs+\longdocs+\autodocs+\formdocs+\corruptdocs}}
\newcommand{\evalpages}{\fpeval{\shortpages+\mediumpages+\longpages+\autopages+\formpages+\corruptpages}}

\newcommand{\formshortdocs}{155}
\newcommand{\mediumformdocs}{14}     %
\newcommand{\autoshortdocs}{6}
\newcommand{\automediumdocs}{8}
\newcommand{\corruptlenshort}{23}
\newcommand{\corruptlenmedium}{12}
\newcommand{\corruptlenlong}{3}
\newcommand{\lonedocs}{\fpeval{\shortdocs+\formshortdocs+\autoshortdocs+\corruptlenshort}}
\newcommand{\ltwodocs}{\fpeval{\mediumdocs+\mediumformdocs+\automediumdocs+\corruptlenmedium}}
\newcommand{\lthreedocs}{\fpeval{\longdocs+\corruptlenlong}}
\newcommand{\lonepages}{615}
\newcommand{\ltwopages}{2438}
\newcommand{\lthreepages}{1816}

\newcommand{\ndomains}{8}          %
\newcommand{\ndoctypes}{67}        %

\newcommand{\formbboxpct}{\fpeval{round(100*\formbboxev/\formeligiblerules,0)}}

\newcommand{\creditorrows}{8624}     %
\newcommand{\unclaimedrows}{26725}   %
\newcommand{\secthirteenfrows}{3063} %

\newcommand{\gmShort}{87.9}

\newcommand{\gmLong}{27.9}

\newcommand{\gmCostCents}{1.0}

\newcommand{\czOverall}{93.6}

\newcommand{\czFthree}{95.4}

\newcommand{\czScan}{93.4}
\newcommand{\czHand}{93.6}

\newcommand{\czDegraded}{81.0}

\newcommand{\czCostCents}{27.8}

\newcommand{\ccOverall}{87.1}

\newcommand{\ccLong}{88.1}

\newcommand{\ccFthree}{82.4}

\newcommand{\ccCostCents}{16.2}

\newcommand{\rdLong}{92.0}

\newcommand{\rdFthree}{87.5}

\newcommand{\rdScan}{81.1}

\newcommand{\ceOverall}{86.8}

\newcommand{\ceCostCents}{1.0}

\newcommand{\agOverall}{89.5}

\newcommand{\agCostCents}{3.1}

\newcommand{\apOverall}{95.6}
\newcommand{\apShort}{96.6}
\newcommand{\apMedium}{93.3}
\newcommand{\apLong}{94.4}

\newcommand{\apFthree}{95.5}

\newcommand{\apCostCents}{8.1}

\newcommand{\ccSfour}{87.8}
\newcommand{\rdSfour}{95.3}
\newcommand{\apSfour}{95.9}
\newcommand{\dlSfour}{32.7}
\newcommand{\exmSfour}{24.8}
\newcommand{\nsystems}{14}

\newcommand{\rdGPageLone}{72.6}

\newcommand{\rdGPageLthree}{67.3}

\newcommand{\apGOverall}{46.4}
\newcommand{\apGPageOverall}{84.9}

\newcommand{\exmGPageLone}{61.7}

\newcommand{\exmGPageLthree}{0.0}
\newcommand{\dlGOverall}{2.0}
\newcommand{\dlGPageOverall}{48.5}

\newcommand{\cmpdocs}{332}

\newcommand{\cmpdegraded}{38}
\newcommand{\cmpratiospan}{7021}
\newcommand{\cmpratiolo}{0.19}
\newcommand{\cmpratiohi}{1359}
\newcommand{\cmpsparsepages}{31}
\newcommand{\cmpsparsefields}{6}
\newcommand{\cmpmaxfields}{86242}
\newcommand{\cmpmaxfieldspages}{114}
\newcommand{\cmpoverthousand}{48}
\newcommand{\cmpovertenk}{12}

\newcommand{\cmpfoneratio}{62.6}
\newcommand{\cmpfonepages}{17}

\newcommand{\cmpftworatio}{1.6}
\newcommand{\cmpftwopages}{25}
\newcommand{\cmpftwofields}{35}

\newcommand{\cmpfthreeratio}{50.0}
\newcommand{\cmpfthreepages}{1}

\theoremstyle{definition}

\title{ExtractBench: A Benchmark for Schema-Guided Enterprise Document Extraction}

\author{
  Boyang Zhang \quad
  Adrian Lyjak \quad
  Eli Stewart \quad
  Zhaoqi Li \quad
  Simon Suo \\[4pt]
  \normalsize \texttt{\{boyang, adrian, eli, zhaoqi, simon\}@runllama.ai}
}

\date{}

\ifdefined\llamaindexbranded
\LIPaperType{Technical report}
\LIPaperDate{July 2026}
\LIPDFAuthor{Boyang Zhang, Adrian Lyjak, Eli Stewart, Zhaoqi Li, Simon Suo}
\LIPDFSubject{A benchmark for schema-guided enterprise document extraction}
\LIAbstract{Enterprise workflows increasingly rely on agents for \emph{schema-guided extraction}: given a document and a user-defined schema, the agent faithfully follows the schema to produce the correct output with source evidence as grounding metadata.
We present ExtractBench, a benchmark for schema-guided extraction, and, to our knowledge, conduct the first evaluation to report value accuracy, record completeness at scale, grounding, and measured cost together.
The benchmark contains \statnum{\evalpages}{} pages across \evaldocs{} enterprise documents, \ndomains{} business domains, and \ndoctypes{} document types, with clear tags differentiating their challenge scenarios.
The scalable schema and ground-truth curation pipeline combines independent-system agreement and adjudication for real documents, known values for synthetic lists, and human verification for forms.
We report order-insensitive value F1 for value accuracy, plus two grounding metrics for source traceability: word- and page-level F1.
Commercial VLMs perform well on short documents but often truncate record lists on long ones, while coding agents retain higher accuracy at much higher cost.
LlamaExtract Agentic Plus ranks first on all three metrics, with higher accuracy than the coding agents at \apCostCents\,\textcent/page versus their \ccCostCents{}--\czCostCents\,\textcent/page.
Dataset and evaluation code are available on \href{https://huggingface.co/datasets/llamaindex/ExtractBench}{HuggingFace} and \href{https://github.com/run-llama/ExtractBench}{GitHub}.
}
\LIPaperLinks{%
  \href{https://huggingface.co/datasets/llamaindex/ExtractBench}{Dataset}
  \hspace{1.1em}
  \href{https://github.com/run-llama/ExtractBench}{Evaluation code}
}
\LIHero{%
  \begin{center}
  \includegraphics[width=\textwidth]{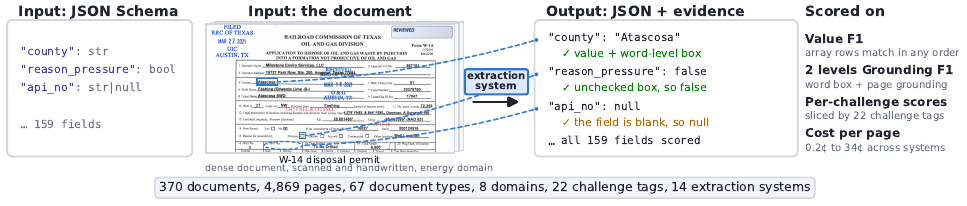}
  \captionof{figure}{ExtractBench scores schema-guided extraction on real enterprise documents: given a document and its schema, a system returns schema-valid JSON with evidence, and is measured on value accuracy, grounding, per-challenge tags, and cost.}
  \label{fig:hero}
  \end{center}
}
\fi

\begin{document}

\ifdefined\llamaindexbranded
\maketitle
\else
\twocolumn[{%
  \maketitle
  \vspace{-1.0em}
  \begin{center}
  \includegraphics[width=\textwidth]{figures/hero_a2_w14.pdf}
  \captionof{figure}{ExtractBench scores schema-guided extraction on real enterprise documents: given a document and its schema, a system returns schema-valid JSON with evidence, and is measured on value accuracy, grounding, per-challenge tags, and cost.}
  \label{fig:hero}
  \end{center}
  \vspace{0.2em}
}]

\begin{abstract}
Enterprise workflows increasingly rely on agents for \emph{schema-guided extraction}: given a document and a user-defined schema, the agent faithfully follows the schema to produce the correct output with source evidence as grounding metadata.
We present ExtractBench, a benchmark for schema-guided extraction, and, to our knowledge, conduct the first evaluation to report value accuracy, record completeness at scale, grounding, and measured cost together.
The benchmark contains \statnum{\evalpages}{} pages across \evaldocs{} enterprise documents, \ndomains{} business domains, and \ndoctypes{} document types, with clear tags differentiating their challenge scenarios.
The scalable schema and ground-truth curation pipeline combines independent-system agreement and adjudication for real documents, known values for synthetic lists, and human verification for forms.
We report order-insensitive value F1 for value accuracy, plus two grounding metrics for source traceability: word- and page-level F1.
Commercial VLMs perform well on short documents but often truncate record lists on long ones, while coding agents retain higher accuracy at much higher cost.
LlamaExtract Agentic Plus ranks first on all three metrics, with higher accuracy than the coding agents at \apCostCents\,\textcent/page versus their \ccCostCents{}--\czCostCents\,\textcent/page.
Dataset and evaluation code are available on \href{https://huggingface.co/datasets/llamaindex/ExtractBench}{HuggingFace} and \href{https://github.com/run-llama/ExtractBench}{GitHub}.

\end{abstract}
\fi

\section{Introduction}
\label{sec:intro}

\begin{table*}[t!]
\centering
\footnotesize
\setlength{\tabcolsep}{2.5pt}
\begin{tabular*}{\textwidth}{@{\extracolsep{\fill}}clllcccccc@{}}
\toprule
&
\textbf{Benchmark}
  & \textbf{Corpus}
  & \textbf{Schemas}
  & \textbf{Domains}
  & \makecell{\textbf{Real}\\\textbf{docs}}
  & \textbf{Long records}
  & \makecell{\textbf{Scans /}\\\textbf{handwriting}}
  & \textbf{Grounding}
  & \textbf{Cost} \\
\midrule
\multirow{3}{*}{\rotatebox[origin=c]{90}{\makecell[c]{\scriptsize\textsc{Fixed}\\[-1pt]\scriptsize\textsc{KIE}}}}
  & \mbox{SROIE~\cite{huang2019sroie}}
  & 1,000
  & fixed
  & 1
  & \cmark
  &
  & \pmark
  & \pmark
  & \\
  & \mbox{DocILE~\cite{simsa2023docile}}
  & 6,680
  & fixed
  & 1
  & \cmark
  &
  & \pmark
  & \cmark
  & \\
  & \mbox{RealKIE~\cite{townsend2025realkienoveldatasetsenterprise}}
  & 1,867
  & fixed
  & 5
  & \cmark
  & \pmark
  & \pmark
  &
  & \\
\midrule
\multirow{5}{*}{\rotatebox[origin=c]{90}{\makecell[c]{\scriptsize\textsc{Schema}\\[-1pt]\scriptsize\textsc{guided}}}}
  & \mbox{Contextual EB~\cite{ferguson2026extractbench}}
  & 35
  & 5
  & 5
  & \cmark
  & \pmark
  &
  &
  & \\
  & \mbox{Extend LongArray~\cite{extend2026longarray}}
  & 45
  & 3
  & 3
  &
  & \cmark
  &
  &
  & \\
  & \mbox{Micro1 LongExtract-50~\cite{micro1longextract}}
  & 50
  & per-doc
  & 7
  & \cmark
  & \cmark
  &
  &
  & \\
  & \mbox{VAREX~\cite{varex2026}}
  & 1,798
  & per-doc
  & 1
  &
  &
  &
  &
  & \\
  & \textbf{ExtractBench (ours)}
  & \textbf{370}
  & \textbf{67}
  & \textbf{8}
  & \cmark
  & \cmark
  & \cmark
  & \cmark
  & \cmark \\
\bottomrule
\end{tabular*}

\vspace{2pt}
{\scriptsize
\raggedright
\cmark\,=\,covered and scored \quad \pmark\,=\,partial or incidental coverage; blank means absent.\par
}
\caption{Comparison of representative fixed-ontology document-IE benchmarks (upper block) and modern schema-guided extraction benchmarks (lower block). ExtractBench is the only benchmark that jointly evaluates long-record completeness, real scans and handwriting, word- and page-level grounding, and measured cost. The full capability matrix appears in \Cref{tab:benchmark-comparison} (\Cref{app:comparison-detail}).}
\label{tab:benchmark-capabilities}
\end{table*}

Until recently, extracting structured data from business documents was performed by humans: knowledge workers read financial filings, insurance claims, purchase orders, and government forms, then keyed the relevant values into a system of record for downstream workflows.
This work is usually highly manual and repetitive, and any mistakes can be costly~\cite{holt2018extracting,chivers2022ants}.
With the recent development of large language models and autonomous agents, we see fast-growing demand from enterprises to deploy agents to complete extraction-focused document workflows historically performed by humans.

Schema definition is at the center of the extraction workflow.
A schema defines one extraction task, shared across all documents of the same type.
For example, one invoice schema covers invoices from every vendor regardless of how different each invoice may look.
Given that enterprises write a new schema for almost every new workflow, a system cannot be tuned to just one fixed template.
We define the extraction task as \emph{schema-guided extraction} (defined precisely in \Cref{sec:task-def}): given a document and a user-defined schema as input, the agent faithfully follows the schema to produce the correct output along with source evidence as grounding metadata.

In real-world use cases faced by enterprises, there are many sources of challenges and failure cases in an extraction workflow, with common ones including missing rows in long lists, selecting the wrong occurrence of a sparse fact, overfilling dense forms, and confusing similar dates, identifiers, or amounts.
There are also particular challenges in accurately understanding the structure of the document, which we call \emph{perception challenges}: scan or handwriting noise, hierarchical headers, cross-page continuation, and large or irregular tables.
Length creates a separate problem: a system can read local values correctly but still truncate a long schedule.

Visual grounding and traceability are another critical element for making agent-powered extraction effective and reliable at production scale for enterprises.
Given that there are always inevitable failures --- such as when an agent fails in reconciling fund holdings because a long schedule is truncated and rows are missing from the output --- it requires a human in the loop to use visual grounding signals to quickly identify and correct the issues.
Additionally, the cost of extraction per page also matters at production volume~\cite{lin2025visual}.
In high-volume, document-intensive enterprise workflows, a cost difference of one cent per page may determine if an AI initiative is financially viable.

Although there have been attempts from multiple existing benchmarks to tackle this challenge, they all have critical limitations (\Cref{tab:benchmark-capabilities}; \Cref{sec:related-work} and \Cref{app:comparison-detail} give the detailed comparison).
Classic information extraction filled fixed, hand-built templates with per-task systems~\cite{grishman1996muc6}.
Fixed KIE benchmarks, such as SROIE~\cite{huang2019sroie} and DocILE~\cite{simsa2023docile}, do not handle user-specified schemas.
More recent schema-guided benchmarks~\cite{ferguson2026extractbench,varex2026} cover only a narrow dimension of the problem.
The three closest benchmarks each cover one slice of these requirements: schema-conformant JSON against enterprise-scale schemas~\cite{ferguson2026extractbench}, row completeness on synthetic rendered arrays~\cite{extend2026longarray}, and long statistical reports and filings~\cite{micro1longextract}.
None of the three measures cost, scores grounding, or contains a scanned or handwritten document.
For example, Contextual AI's ExtractBench~\cite{ferguson2026extractbench}\footnote{The unrelated academic benchmark by Contextual AI shares the ExtractBench name~\cite{ferguson2026extractbench}; \Cref{sec:related-work} details how the two differ in scope.} does not cover any handwritten documents, nor does it take visual grounding or cost into consideration.

To bridge the gap, we introduce ExtractBench, a comprehensive benchmark for schema-guided enterprise document extraction that carries broad task coverage, evaluates traceability, and measures cost.
ExtractBench contains \evaldocs{} documents (\statnum{\evalpages}{} pages) across \ndomains{} business domains and \ndoctypes{} document types.
Each document type has one schema shared across its documents.
Each document is tagged by task challenge, perception challenge, table structure, domain, and length (\Cref{sec:taxonomy}).
The benchmark is composed of real born-digital documents, synthetic long lists based on real layouts, and real regulatory and tax forms with schemas authored from blank templates.
To establish high-quality ground truth at scale, we design a scalable pipeline: independent-system proposals are adjudicated for real documents, values are set before rendering for synthetic lists, and humans verify both values and grounding on scanned forms (\Cref{sec:ground-truth}).
We evaluate accuracy with order-insensitive value F1 over the values in the extracted JSON.
To evaluate visual grounding ability, we also score whether a correct value points to its source for fields with human-verified boxes, so reviewers can audit the answer without searching the document (\Cref{sec:grounded}).

We evaluate \nsystems{} frontier methods spanning commercial VLMs, open-source extraction, coding agents, and specialized APIs.
We noticed significant performance variance in out-of-the-box frontier models across different challenge dimensions.
For example, Gemini 3.5 Flash accuracy dropped significantly from \gmShort\% on short documents to \gmLong\% on long ones (\Cref{sec:results}).
LlamaExtract Agentic Plus shows much more consistent performance, with \apShort\% on short and \apLong\% on long documents.
It also outperforms Codex GPT-5.5 (\apOverall\% versus \czOverall\%) at a much lower cost (\apCostCents\,\textcent/page versus \czCostCents\,\textcent/page).
Additionally, commercial VLMs and coding agents do not return word-level boxes, so workflows that require source evidence need specialized extraction APIs (\Cref{sec:grounding-gap}).

Our contributions include:
\begin{itemize}[leftmargin=1.2em]
  \item \textbf{A challenge-tagged benchmark with broad coverage.}
  \evaldocs{} documents and \statnum{\evalpages}{} pages span \ndomains{} business domains and \ndoctypes{} document types, with tags for task challenge, perception challenge, table structure, domain, and length that support per-challenge analysis.
  \item \textbf{A scalable pipeline for schema and ground-truth curation.}
  To produce high-quality, well-specified schema--ground-truth pairs without labeling every field by hand, we combine frontier-model ensembles for real documents, programmatic generation for synthetic long lists, and human verification for scanned forms.
  \item \textbf{A broad evaluation of frontier extraction methods.}
  We compare \nsystems{} systems spanning commercial VLMs, OSS pipelines, coding agents, and specialized APIs, reporting accuracy, grounding, cost, and the quality--cost tradeoff.
\end{itemize}

\section{ExtractBench}
\label{sec:extractbench}

This section defines schema-guided extraction precisely, then describes how ExtractBench applies it to build a challenge-tagged corpus.

\subsection{Task Definition}
\label{sec:task-def}

Given a document and a schema, a system returns structured data with evidence (\Cref{fig:hero}):
\[
  f:\,(\text{document},\ \text{schema}) \longmapsto (\text{structured data},\ \text{evidence}).
\]

\paragraph{Input.}
The input is a full document, born-digital or scanned, and a schema written by the user.
The user specifies the extraction task via a schema: it lists the fields to extract, and each field has a name, a type, and a natural-language description of what belongs in it.
It is expressed as a \emph{JSON Schema}, the industry-standard way to specify structured
output, and may contain scalar fields, nested objects, arrays of records, nullable fields, and value
constraints.
A schema defines one extraction task and guides all documents of the same type, even though the documents can vary significantly in structure, layout, and styling.
For example, insurance claims from different providers vary greatly in length and layout.
A document usually holds more than the schema asks for --- and sometimes less: any field the document leaves unanswered must come back as null.

\paragraph{Output.}
The output is a schema-valid JSON object, with the source page and a bounding box for each value as \emph{evidence}.
It must return correct, exhaustive values (including repeated records), correctly use null for absent information, and ground each extracted value.
\Cref{sec:metrics} makes these expectations precise.

\subsection{Taxonomy and Coverage}
\label{sec:taxonomy}
\label{sec:task-families}  %
\label{sec:corpus}  %
\begin{figure*}[!t]
\centering
\includegraphics[width=\textwidth]{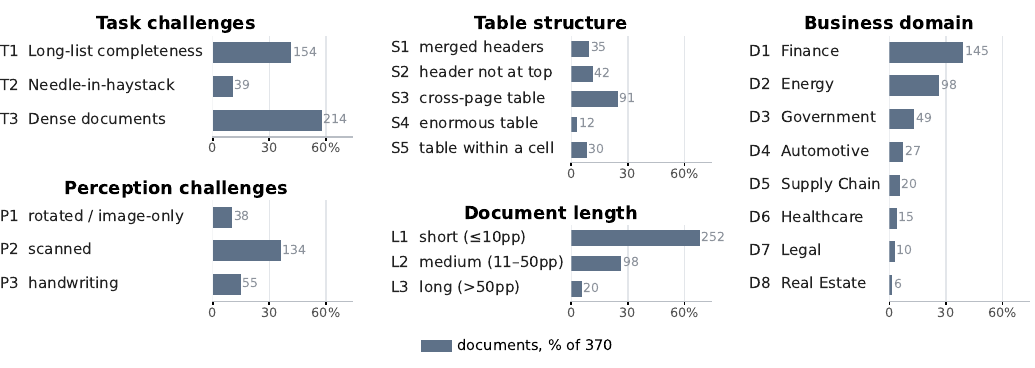}
\caption{ExtractBench coverage across the five tag axes. Each bar is
the share of the \evaldocs{} documents carrying the tag, with its document count shown beside the bar. The task panel
reports the three task challenges; a document tagged with several of one challenge's sub-tags counts once.
Tags may overlap across panels; \Cref{tab:tax-challenge} (\Cref{app:taxonomy}) defines every tag
and sub-tag.
}
\label{fig:taxonomy-coverage}
\end{figure*}

Existing benchmarks for document extraction often report one aggregate score over a narrow set of
document types.
An aggregate score does not show whether a system missed a third of a list or got one label wrong,
and it cannot distinguish a hard extraction task from a bad scan.
ExtractBench instead tags each document along five independent axes: task challenge (what makes
extraction hard), perception challenge (how the page was captured), table structure, length, and
business domain.
Because the axes are independent, a low score can be traced to its actual cause.
\Cref{fig:taxonomy-coverage} shows how the corpus distributes over the five axes.
The paragraphs below briefly explain each axis; \Cref{tab:tax-challenge} in \Cref{app:taxonomy}
gives additional details and representative document types for every tag.

\paragraph{Task challenges.}
A \emph{task challenge} defines the nature of the extraction task and what makes it difficult.
\begin{itemize}[leftmargin=1.2em]
  \item \textbf{T1: long-list completeness.}
  Recover \emph{every} record of a repeated structure that can span many pages.
  Typical failures are truncation, duplicated or merged rows, hallucinated records, and values
  attached to the wrong record.
  \item \textbf{T2: needle-in-haystack.}
  Find a small number of requested facts in a long document.
  T2 has few target records but many plausible mentions, only one of which is canonical; failures are
  missed targets, wrong occurrences, and unnormalized paraphrases.
  It is also the only task challenge that asks for far less than the document holds: a median of
  just \cmpftworatio{} fields per page (\Cref{app:compression}).
  \item \textbf{T3: dense documents.}
  Fill many fields from a document dense with labels, blanks, checkboxes, handwriting, and
  scan artifacts.
  The characteristic failure is over-extraction, inventing a value for a field that is actually
  blank, compounded by missed checkboxes and mislabeled fields.
  A dense document also repeats identifiers, dates, and amounts of the same format, so a plausible
  value can end up in the wrong field.
  Dense forms are the most common case (T3.a); receipts, invoices, and regulatory filings belong
  here too.
  T3.e marks schemas with more than 150 leaf fields. It includes multi-page tax bundles whose
  schemas are large even when individual pages are not especially dense.
\end{itemize}

\paragraph{Perception challenges.}
A \emph{perception challenge} records how the page was captured.
The tags are rotated or image-only capture (P1), scanned page images (P2), and handwriting (P3)
(\Cref{tab:tax-challenge}).
They are assigned independently of the task challenge, so the same extraction task can appear under
more than one perception challenge.

\paragraph{Table structure.}
Tables earn a dedicated axis for two reasons. First, most of the values enterprises extract live in
tables, from holdings schedules to invoice line items.
Second, tables fail in a way no other page element does: a complex table can be read correctly
value by value and still be assembled into the wrong structure, a failure that the task and
perception axes cannot isolate.
The structure tags mark the layouts where this happens: merged or hierarchical headers
(S1), a header that does not sit above its data (S2), a table that continues across pages (S3), a
table beyond a thousand rows (S4), and a table packed inside a single cell (S5).
Each layout has its own failure: a merged header attaches values to the wrong column, a pivoted
header transposes the record, a cross-page table loses its continuation, a very large table stops
early, and a packed cell comes back as one string instead of its fields.

\paragraph{Document length.}
Documents fall into three length buckets: short (L1, up to 10 pages), medium (L2, 11 to 50), and
long (L3, more than 50).
Length gets its own axis for the same reason the other axes are separate: the same task challenge
can appear at any length, and length adds a failure of its own, since a system can read every value
on a page correctly and still stop before the end of a long schedule.

\paragraph{Business domains.}
An enterprise extraction system needs to work across domains, for two reasons: teams want one
system rather than a separate tool per document type, and businesses often do not control what
arrives and must process whatever their customers, vendors, and regulators send.
ExtractBench therefore spans \ndomains{} domains and \ndoctypes{} document types (\Cref{tab:tax-domain}):
finance and fund holdings (D1), energy-sector regulatory forms (D2), government procurement and
customs (D3), auto valuation (D4), supply-chain and other transactional documents (D5),
healthcare remittance (D6), legal and bankruptcy filings (D7), and real-estate closing disclosures
(D8).
Prior benchmarks for structured extraction usually cover fewer domains and a handful of real
document types (\Cref{tab:benchmark-breadth}).

\begin{figure*}[t!]
  \centering
  \includegraphics[width=\textwidth]{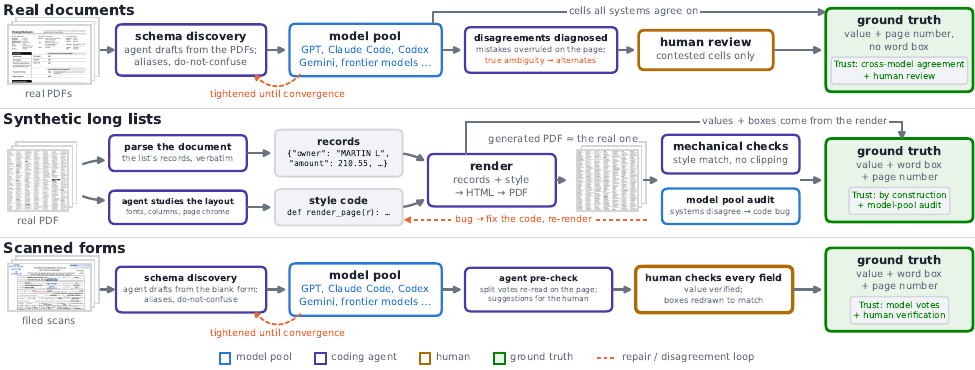}
  \caption{How ground truth is constructed, one strip per source type.
  A card's border marks who runs the step (blue: the extraction-model pool; violet: coding-agent pipeline code; amber: a human); the filled green card is the resulting ground truth, and orange dashes mark repair and disagreement loops.}
  \label{fig:gt-pipelines}
\end{figure*}

\subsection{Schema and Ground-Truth Construction}
\label{sec:ground-truth}

To properly evaluate a schema-guided extraction system, the extraction task itself needs to be well specified: each schema must be coherent with the documents it applies to, and every field must have a clear expected value.
If a schema is poorly aligned with its document family, or its instructions leave the extraction goal ambiguous, errors can no longer be attributed and a low score may reflect a defect in the benchmark rather than in the system being tested.

Creating well-defined schemas and ground truth that corresponds to them is labor-intensive, particularly when documents are long and the data is dense.
Checking every field by hand is prohibitive in time and cost at this scale, and using a single extractor's output as ground truth would repeat its mistakes and bias the results toward that extractor.
This motivates a scalable pipeline that produces high-quality schema and ground-truth pairs without fully manual annotation.

To this end, we combine three sources of documents, each annotated by the method that fits it: frontier-model ensembles for real documents, programmatic generation for synthetic long lists, and human labelers for scanned forms (\Cref{fig:gt-pipelines}).
Real documents supply the layouts, scan noise, and domain range we want to test, but drawing a box on
every one of their fields is prohibitively slow.
Synthetic long lists cover documents too large to label by hand: thousands of similar records are slow
to annotate and easy to misread.
Scanned forms are real documents that need a person to decide each value, because handwriting is
unclear and a mark can sit between two fields.
Documents from these sources are also re-captured as degraded scans, which needs no new
annotation: the values do not change, so the clean document's ground truth carries over.
We use this methodology to build ExtractBench.
\Cref{app:methodology} gives the full procedures.

\paragraph{Schemas.}
A document type is a family of documents that carry the same kind of information --- SEC 13F filings, utility bills, mortgage closing disclosures --- however much their layouts differ.
In ExtractBench, each document type has exactly one schema, shared by all of its documents.
The schema takes the form a user writes in production: field names, types, and a natural-language description for each field (\Cref{sec:task-def}).
How each schema is authored depends on its source and is described with each pipeline below.
Every field is written to have a deterministic expected value in the document, so the ground truth is the same no matter which system is being scored.

\paragraph{Real documents.}
\label{sec:gt-real}
The schema is drafted from sample documents, then several extraction systems from different model and pipeline families run against this candidate schema (\Cref{fig:gt-pipelines}, top strip): no single extractor is reliable enough, and same-family systems share mistakes.
A value on which every system agrees, including null for absent fields, becomes candidate ground truth.
Disagreements are classified by cause: if more than one reading of the field is defensible, the schema is at fault, and we tighten its description with aliases, format requirements, location hints, and do-not-confuse guidance until reruns converge; if only one reading is defensible, it is a model failure, and a reviewer settles the contested cells against the page.

\paragraph{Synthetic long lists.}
\label{sec:gt-synthetic}
We build each synthetic document backwards, data first and document second, so no human labeling is needed.
From a real filing (a fund schedule, holdings register, or creditor matrix), we produce the records, parsed verbatim or generated in its style, and rendering code that a coding agent writes after studying the layout's fonts, columns, and page chrome (\Cref{fig:gt-pipelines}, middle strip); the family keeps the real filing's schema.
That code renders the records into a PDF closely matching the real one, with page breaks placed by measurement.
Every value is known before the PDF exists, and its page and word-level box are read back from the render, so the ground truth stays exact however long the list grows.
Mechanical checks catch style mismatches and clipping, and an extraction-system pool audits the finished document, its disagreements exposing rendering-code bugs that are fixed before the family ships.

\begin{figure*}[!t]
\centering
\includegraphics[width=\textwidth]{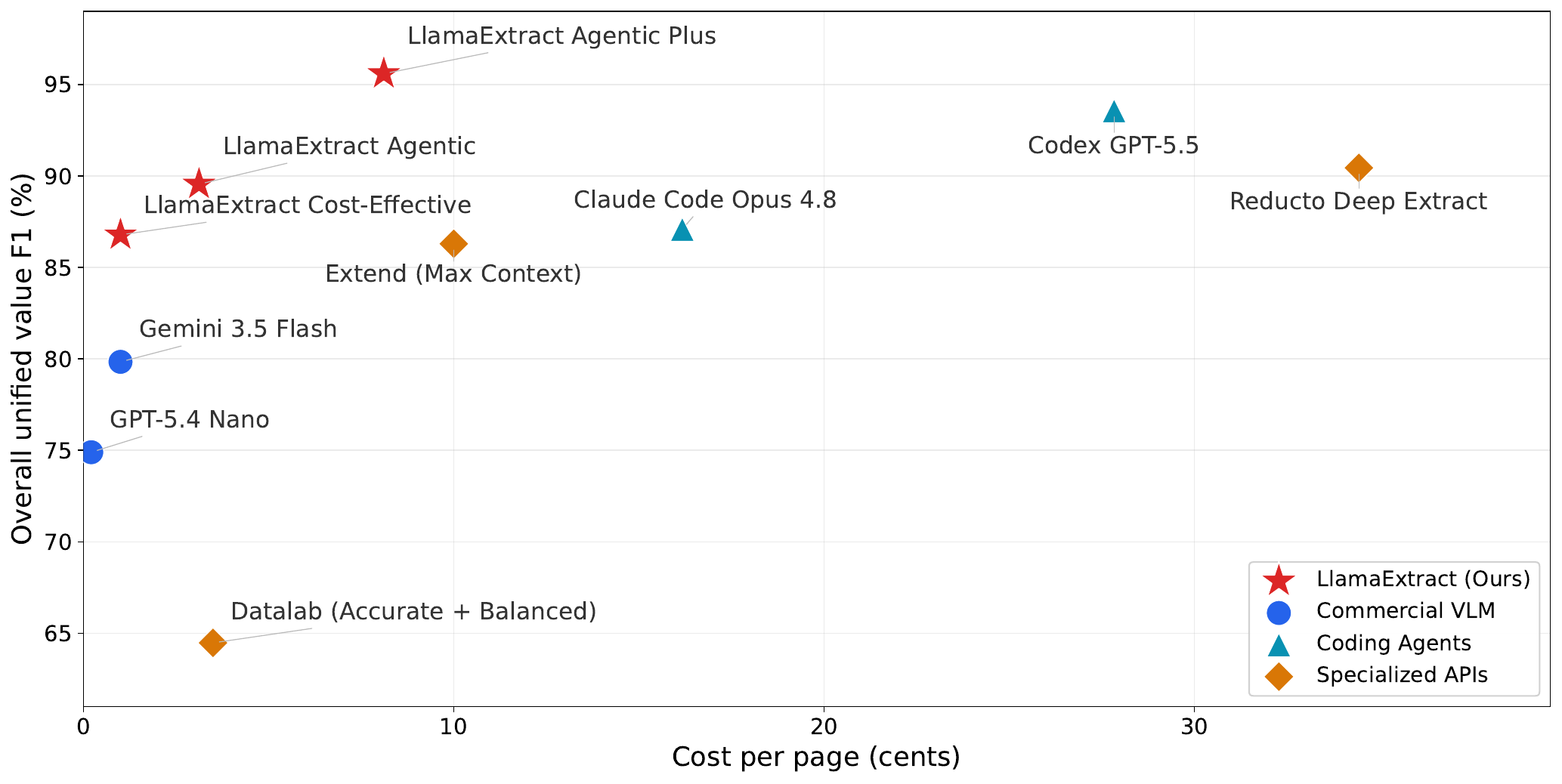}
\caption{Overall unified value F1 versus mean document-level cost per page, pooled over every scored document. Marker shape and color follow the system grouping of \Cref{sec:setup}. The four open-weight pipelines have no vendor API price and are omitted. Per-length results appear in \Cref{app:cost-detail}.}
\label{fig:quality-cost}
\end{figure*}
\begin{table*}[!t]
\centering
\normalsize
\setlength{\tabcolsep}{0pt}
\renewcommand{\arraystretch}{0.85}
\begin{tabular}{@{}l>{\centering\arraybackslash}p{2.5em}>{\centering\arraybackslash}p{2.5em}>{\centering\arraybackslash}p{2.5em}>{\centering\arraybackslash}p{2.5em}>{\centering\arraybackslash}p{2.5em}>{\centering\arraybackslash}p{2.5em}>{\centering\arraybackslash}p{2.5em}>{\centering\arraybackslash}p{2.5em}>{\centering\arraybackslash}p{2.5em}>{\centering\arraybackslash}p{2.5em}>{\centering\arraybackslash}p{2.5em}>{\centering\arraybackslash}p{2.5em}>{\centering\arraybackslash}p{2.5em}>{\centering\arraybackslash}p{2.5em}@{}}
\toprule
 & \multicolumn{6}{c}{\makebox[0pt][c]{\makecell{\textbf{Specialized APIs}}}} & \multicolumn{2}{c}{\makebox[0pt][c]{\makecell{\textbf{Coding}\\\textbf{Agents}}}} & \multicolumn{4}{c}{\makebox[0pt][c]{\makecell{\textbf{OSS}}}} & \multicolumn{2}{c}{\makebox[0pt][c]{\makecell{\textbf{Commercial}\\\textbf{VLM}}}} \\
\cmidrule(lr){2-7}\cmidrule(lr){8-9}\cmidrule(lr){10-13}\cmidrule(lr){14-15}
\textbf{Dimension} & \rotatebox{80}{\textbf{LE Agentic Plus}} & \rotatebox{80}{\textbf{LE Agentic}} & \rotatebox{80}{\textbf{LE Cost-Eff.}} & \rotatebox{80}{\textbf{Datalab A+B}} & \rotatebox{80}{\textbf{Extend Max}} & \rotatebox{80}{\textbf{Reducto Deep}} & \rotatebox{80}{\textbf{Codex GPT-5.5}} & \rotatebox{80}{\textbf{CC Opus 4.8}} & \rotatebox{80}{\textbf{Gemma4 26B}} & \rotatebox{80}{\textbf{Qwen3.6 35B-A3B}} & \rotatebox{80}{\textbf{NuExtract3}} & \rotatebox{80}{\textbf{Lift 9B}} & \rotatebox{80}{\textbf{GPT-5.4 Nano}} & \rotatebox{80}{\textbf{Gemini 3.5 Flash}} \\
\midrule
\textbf{Overall} & \textbf{95.6} & 89.5 & 86.8 & 64.5 & 86.3 & 90.4 & \underline{93.6} & 87.1 & 66.2 & 87.3 & 47.9 & 77.3 & 74.9 & 79.8 \\
\midrule
\multicolumn{15}{@{}l}{\textbf{Document Length}} \\
L1 Short ($\le$10 pp) & \textbf{96.6} & 92.0 & 90.8 & 62.8 & 92.0 & 94.2 & \underline{95.7} & 90.1 & 80.5 & 93.1 & 54.4 & 87.2 & 77.4 & 87.9 \\
L2 Medium (11--50 pp) & \textbf{93.3} & 85.4 & \cellcolor{resultdroplight}80.1 & 73.8 & \cellcolor{resultdroplight}78.8 & \cellcolor{resultdroplight}80.5 & \underline{91.2} & \cellcolor{resultdroplight}79.2 & \cellcolor{resultdropstrong}40.5 & 84.8 & \cellcolor{resultdroplight}39.3 & \cellcolor{resultdroplight}62.6 & 76.4 & \cellcolor{resultdroplight}69.8 \\
L3 Long ($>$50 pp) & \textbf{94.4} & \cellcolor{resultdroplight}78.6 & \cellcolor{resultdropmedium}69.2 & \cellcolor{resultdropmedium}40.5 & \cellcolor{resultdropstrong}51.3 & \underline{92.0} & \cellcolor{resultdroplight}78.9 & 88.1 & \cellcolor{resultdropstrong}12.2 & \cellcolor{resultdropstrong}26.8 & \cellcolor{resultdropstrong}8.9 & \cellcolor{resultdropstrong}25.3 & \cellcolor{resultdropstrong}35.8 & \cellcolor{resultdropstrong}27.9 \\
\midrule
\multicolumn{15}{@{}l}{\textbf{Task Challenge}} \\
T1 Long-list completeness & \textbf{96.1} & 85.9 & \cellcolor{resultdroplight}81.8 & 80.2 & 87.2 & \underline{94.8} & 91.7 & 93.6 & \cellcolor{resultdropmedium}51.1 & \cellcolor{resultdroplight}79.0 & \cellcolor{resultdropmedium}31.8 & \cellcolor{resultdroplight}68.6 & 72.2 & 78.8 \\
T2 Needle-in-haystack & \textbf{93.6} & 88.3 & 82.3 & 73.9 & 90.3 & \underline{92.5} & 91.7 & 89.1 & 63.0 & 85.3 & \cellcolor{resultdropmedium}25.2 & 78.0 & 74.0 & 87.9 \\
T3 Dense documents & \textbf{95.5} & 92.1 & 90.5 & \cellcolor{resultdroplight}54.4 & 85.7 & 87.5 & \underline{95.4} & 82.4 & 76.8 & 93.1 & 58.7 & 82.9 & 76.4 & 80.5 \\
\midrule
\multicolumn{15}{@{}l}{\textbf{Perception Challenge}} \\
P1 Rotated / image-only & \textbf{95.9} & 88.2 & 85.0 & 78.9 & 89.0 & \underline{93.9} & \cellcolor{resultdroplight}81.0 & 91.2 & 66.5 & 86.8 & \cellcolor{resultdropmedium}28.9 & 80.7 & \cellcolor{resultdroplight}64.7 & 88.6 \\
P2 Scanned & \textbf{93.9} & 89.8 & 87.8 & \cellcolor{resultdropmedium}47.6 & \cellcolor{resultdroplight}80.9 & \cellcolor{resultdroplight}81.1 & \underline{93.4} & \cellcolor{resultdroplight}74.2 & 69.1 & 92.0 & 62.6 & 76.0 & \cellcolor{resultdroplight}67.4 & \cellcolor{resultdroplight}71.1 \\
P3 Handwriting & \textbf{93.8} & 90.9 & 87.6 & \cellcolor{resultdropmedium}47.2 & \underline{93.8} & 92.3 & 93.6 & \cellcolor{resultdroplight}74.7 & 73.9 & 92.3 & 75.8 & 86.0 & \cellcolor{resultdroplight}67.7 & \cellcolor{resultdroplight}74.2 \\
\midrule
\multicolumn{15}{@{}l}{\textbf{Table Structure}} \\
S1 Merged headers & \underline{94.5} & \cellcolor{resultdroplight}78.5 & \cellcolor{resultdroplight}79.4 & 82.8 & 91.0 & 94.2 & \textbf{95.0} & 94.0 & \cellcolor{resultdropmedium}44.6 & \cellcolor{resultdroplight}81.8 & \cellcolor{resultdroplight}33.3 & \cellcolor{resultdroplight}69.9 & \cellcolor{resultdroplight}68.9 & 80.2 \\
S2 Pivoted / header not at top & \underline{95.0} & 86.1 & 87.2 & 84.4 & 91.4 & \textbf{95.3} & 94.9 & 94.3 & 63.0 & 89.6 & \cellcolor{resultdropstrong}20.9 & 76.4 & 77.7 & 88.4 \\
S3 Cross-page table & \textbf{95.8} & \cellcolor{resultdroplight}84.3 & \cellcolor{resultdroplight}79.0 & 78.5 & 85.1 & \underline{94.4} & 89.4 & 92.5 & \cellcolor{resultdropstrong}40.5 & \cellcolor{resultdroplight}73.8 & \cellcolor{resultdroplight}37.6 & \cellcolor{resultdroplight}64.5 & 72.3 & \cellcolor{resultdroplight}73.6 \\
S4 Enormous table & \textbf{95.9} & \cellcolor{resultdroplight}78.1 & \cellcolor{resultdropmedium}67.8 & \cellcolor{resultdropstrong}32.7 & \cellcolor{resultdropstrong}24.8 & \underline{95.3} & \cellcolor{resultdroplight}78.9 & 87.8 & \cellcolor{resultdropstrong}0.0 & \cellcolor{resultdropstrong}1.3 & \cellcolor{resultdropstrong}3.7 & \cellcolor{resultdropstrong}1.2 & \cellcolor{resultdropstrong}7.2 & \cellcolor{resultdropstrong}1.5 \\
S5 Table within a cell & \textbf{97.2} & 87.3 & \cellcolor{resultdroplight}78.2 & 71.7 & \cellcolor{resultdroplight}75.9 & \underline{95.4} & \cellcolor{resultdroplight}86.8 & 93.9 & \cellcolor{resultdropstrong}37.1 & \cellcolor{resultdropstrong}56.6 & 50.1 & \cellcolor{resultdropstrong}51.3 & \cellcolor{resultdroplight}67.3 & \cellcolor{resultdropstrong}53.4 \\
\midrule
\multicolumn{15}{@{}l}{\textbf{Business Domain}} \\
D1 Finance & \textbf{96.3} & 91.3 & 87.2 & 62.1 & \cellcolor{resultdroplight}79.2 & \cellcolor{resultdroplight}85.1 & \underline{96.2} & 84.7 & \cellcolor{resultdroplight}59.5 & 85.9 & 48.4 & \cellcolor{resultdroplight}71.8 & 77.5 & 75.4 \\
D2 Energy & \textbf{95.0} & 90.7 & 88.5 & \cellcolor{resultdropmedium}49.0 & \underline{94.5} & 93.7 & 94.1 & 82.9 & 78.7 & 92.4 & 76.6 & 87.9 & 73.0 & 78.9 \\
D3 Government & \textbf{93.4} & \cellcolor{resultdroplight}83.4 & 83.0 & 77.9 & 86.5 & 92.5 & \underline{92.7} & 91.0 & \cellcolor{resultdroplight}58.1 & 83.7 & \cellcolor{resultdropstrong}18.0 & \cellcolor{resultdroplight}71.5 & 74.3 & 81.4 \\
D4 Automotive & 97.9 & 95.0 & 95.2 & 85.0 & 91.9 & 97.3 & 95.2 & \underline{98.0} & 83.9 & 96.6 & \cellcolor{resultdropstrong}14.8 & 85.9 & 79.3 & \textbf{98.0} \\
D5 Supply Chain & 97.9 & 93.2 & 92.0 & 82.5 & 96.8 & 96.1 & 95.9 & \textbf{99.0} & 87.4 & 95.4 & \cellcolor{resultdropmedium}31.6 & 93.5 & 83.7 & \underline{98.2} \\
D6 Healthcare & 92.6 & \cellcolor{resultdropmedium}74.1 & \cellcolor{resultdropmedium}70.7 & 77.9 & \underline{93.3} & 90.3 & \cellcolor{resultdroplight}82.8 & \textbf{95.1} & \cellcolor{resultdropstrong}39.4 & \cellcolor{resultdropmedium}69.0 & \cellcolor{resultdroplight}34.3 & \cellcolor{resultdropmedium}61.8 & \cellcolor{resultdropmedium}55.2 & \cellcolor{resultdroplight}74.5 \\
D7 Legal & \textbf{96.7} & \cellcolor{resultdroplight}81.5 & \cellcolor{resultdropmedium}69.1 & 66.6 & \cellcolor{resultdropstrong}56.1 & \underline{92.8} & \cellcolor{resultdropstrong}62.0 & \cellcolor{resultdroplight}74.3 & \cellcolor{resultdropstrong}17.9 & \cellcolor{resultdropstrong}58.8 & 56.0 & \cellcolor{resultdropstrong}42.2 & \cellcolor{resultdropmedium}51.7 & \cellcolor{resultdropmedium}58.3 \\
D8 Real Estate & 94.0 & 93.7 & 93.0 & 73.8 & 93.2 & \textbf{95.9} & 93.8 & 93.4 & 90.8 & 93.6 & \cellcolor{resultdroplight}37.8 & 89.8 & 85.8 & \underline{94.7} \\
\bottomrule
\end{tabular}%
\caption{Unified value F1 (\%) by dimension and system. Models are columns, grouped by system type, and dimensions are rows. Within each row, \textbf{bold} and \underline{underlined} mark the highest and second-highest scores. Red shading marks drops of more than 5/15/25 points from each system's overall score (darker means larger). Overall aggregates each system's evaluated documents. \Cref{tab:res-challenge-detail} (\Cref{app:challenge-detail}) reports every sub-tag.}
\label{tab:results-by-dimension}
\label{tab:res-length}
\label{tab:res-challenge}
\label{tab:res-modality}
\label{tab:res-structure}
\label{tab:res-domain}
\end{table*}

\begin{table*}[!t]
\centering
\small
\setlength{\tabcolsep}{5pt}
\begin{tabular*}{\textwidth}{@{\extracolsep{\fill}}lcccccccc@{}}
\toprule
 & \multicolumn{4}{c}{\textbf{Word-level grounding F1}} & \multicolumn{4}{c}{\textbf{Page-level grounding F1}} \\
\cmidrule(lr){2-5}\cmidrule(l){6-9}
\textbf{System} & Overall & Short & Medium & Long & Overall & Short & Medium & Long \\
\midrule
LE Agentic Plus & \textbf{46.4} & \textbf{43.7} & \textbf{54.0} & \textbf{54.7} & \textbf{84.9} & \textbf{89.7} & \textbf{72.2} & \textbf{87.1} \\
LE Agentic & \underline{44.1} & 42.3 & \underline{50.5} & \underline{45.7} & 66.1 & 69.7 & 56.6 & \underline{67.6} \\
LE Cost-Eff. & 40.4 & 40.2 & 42.3 & 36.7 & 64.2 & 68.9 & 53.7 & 56.5 \\
Datalab A+B & 2.0 & 2.7 & 0.2 & 0.0 & 48.5 & 56.9 & 38.6 & 0.0 \\
Extend Max & 25.1 & 33.9 & 0.2 & 0.0 & 48.9 & 61.7 & 27.7 & 0.0 \\
Reducto Deep & 43.3 & \underline{42.8} & 45.6 & 41.1 & \underline{71.7} & \underline{72.6} & \underline{70.4} & 67.3 \\
\midrule
\emph{All other systems} & 0.0 & 0.0 & 0.0 & 0.0 & 0.0 & 0.0 & 0.0 & 0.0 \\
\bottomrule
\end{tabular*}
\caption{Grounding score (\%). Word-level grounding F1 requires a correct value and word-level box at IoU 0.5; page-level grounding F1 requires a correct value and the source page. The final row spans all systems not named above. \textbf{Bold} marks each column's best value, \underline{underlined} the second best. LE Agentic and LE Cost-Effective return word-level boxes only when the caller enables the granular bounding-box option; both are run with it on.}
\label{tab:grounding}
\end{table*}

\paragraph{Scanned forms.}
\label{sec:gt-forms}
Scanned forms are the one source where a person checks every field (\Cref{fig:gt-pipelines}, bottom strip).
The schema is authored against the blank form template and frozen before any document is labeled.
An ensemble of up to five systems votes on every schema leaf; contested votes go to an adjudication agent that must inspect the page before ruling.
A designated pipeline proposes a box per field, and a human annotator accepts, edits, nulls, or redraws each one.
This yields \formdocs{} human-verified documents, with \formbboxpct\% of verified fields carrying a human-placed box; the rest are mostly blank fields, with nothing on the page to box.

\label{sec:trust}
The three pipelines back their ground truth differently: real-document values are confirmed by agreement across independent systems, synthetic values and boxes are exact by construction, and form values and boxes are checked by a person.
This determines which metrics each document supports: values are scored everywhere, box-level grounding only where the boxes are verified (\Cref{sec:metrics}).

\subsection{Metrics}
\label{sec:metrics}

ExtractBench measures two things.
Value accuracy asks whether a system returned the right values, and is scored with the unified value F1 on every document.
Grounding asks whether the system can show where each value came from, and is scored only on documents whose box ground truth is verified (\Cref{sec:trust}).

\paragraph{Value accuracy.}
\label{sec:headline}
The unified value F1 scores whether the extracted \emph{values} match the expected output, under one definition for scalar fields and arrays of records.
Each output is flattened into cells, one per scalar field and per aligned record subfield, and a cell is correct when it matches its expected counterpart after normalization.
Precision, recall, and F1 are computed over these cells per document, and slices report unweighted document means (\Cref{app:metric-detail} gives the exact scoring rules).
\begin{itemize}[leftmargin=1.2em]
  \item \textbf{Array alignment.} A repeated structure is compared as an unordered set of records: records are paired by the Hungarian algorithm to minimize mismatched cells, following how prior extraction benchmarks align line items and arrays~\cite{simsa2023docile}. Unmatched expected records lower recall; extra predictions lower precision.
  \item \textbf{Normalization.} Values are normalized before comparison: dates to ISO format, strings by collapsing whitespace; everything else requires exact equality, with no numeric tolerance and no LLM judge. The few exceptions are listed in \Cref{app:normalization}.
  \item \textbf{Missing values.} An omitted key scores as an explicit null, so every scalar field counts toward both precision and recall, and a correct null on a blank field is credited (\Cref{tab:null-matrix} lists every case). Only repeated records move precision and recall apart, so a gap between them points to dropped or extra records rather than wrong values (\Cref{app:pr-detail} tabulates both per system and length slice).
\end{itemize}

\paragraph{Grounding.}
\label{sec:grounded}
For fields with a verified ground-truth box, ExtractBench also reports word-level grounding precision, recall, and F1.
A field counts as grounded only when its value is correct and its predicted box overlaps an accepted box for that field, at a fixed IoU threshold of 0.5: a well-placed box around a wrong value earns no credit.
Page-level grounding F1 asks the weaker version of the same question, requiring only the correct source page rather than a box, which many systems satisfy even when they return no boxes at all.
\Cref{app:aggregation} gives the details.

\section{Experiments}
\label{sec:results}

\subsection{Setup}
\label{sec:setup}

We evaluate \nsystems{} extraction systems\footnote{Models and prices reflect those available as of July 1, 2026.} across three high-level approaches:
\begin{itemize}[leftmargin=1.2em]
  \item \textbf{VLMs} treat extraction as direct multimodal generation: they receive the document and schema and generate structured output in a single model call. We evaluate GPT-5.4 Nano~\cite{openai2026gpt54nano} and Google Gemini 3.5 Flash~\cite{google2026gemini35flash}, called through constrained structured-output APIs; and Lift 9B~\cite{datalab2026lift}, NuExtract3~\cite{numind2026nuextract3}, Qwen3.6 35B-A3B~\cite{qwen2026qwen36}, and Gemma4 26B~\cite{google2026gemma4}, which are self-hosted.
  \item \textbf{Coding agents} extract through an iterative tool-use loop: they can inspect the document, write and run parsing code, validate results, and revise the final output. We evaluate Claude Code Opus 4.8~\cite{anthropic2026opus48} and Codex GPT-5.5~\cite{openai2026gpt55}, which receive the document and schema with filesystem and tool access (tool configuration in \Cref{app:eval-config}).
  \item \textbf{Specialized APIs} provide a managed document workflow that handles preprocessing, parsing, and schema-guided extraction, sometimes with source grounding. We evaluate Reducto Deep Extract~\cite{reducto2026deep}, Extend Max Context~\cite{extend2026maxcontext}, Datalab Accurate Parse + Balanced Extract~\cite{datalab2026extract,datalab2026pricing}, and three LlamaExtract tiers~\cite{llamaindex2026llamaextract} (Cost-Effective, Agentic, and Agentic Plus).
\end{itemize}

All systems receive the same document--schema pairs and are evaluated without benchmark-specific tuning; all runs took place in June--July 2026.
We report unweighted document-level means for value F1, word- and page-level grounding F1, and cost, using the metrics defined in \Cref{sec:metrics}.
Per-page costs apply published provider rates to actual token or credit consumption (\Cref{app:pricing}).
Because the four self-hosted VLMs have no directly comparable API price, we omit them from cost comparisons.

\subsection{Quality--Cost Frontier}
\label{sec:cost}

Enterprise extraction workloads often span millions of pages, making per-page cost differences substantial.
At one million pages, each cent per page adds \$10{,}000.
\Cref{fig:quality-cost} compares overall value F1 with measured per-page cost.

The evaluated system families occupy distinct regions of this tradeoff.
The VLMs with reported costs occupy the low-cost region ($\leq$\gmCostCents\,\textcent/page), but neither exceeds 80\% F1.
Coding agents reach \ccOverall\% and \czOverall\% F1, but cost \ccCostCents{} and \czCostCents\,\textcent/page.
Specialized APIs span a much wider range.
Within this group, LlamaExtract traces the quality--cost frontier: Cost-Effective reaches \ceOverall\% F1 at \ceCostCents\,\textcent/page, Agentic reaches \agOverall\% at \agCostCents\,\textcent/page, and Agentic Plus reaches \apOverall\% at \apCostCents\,\textcent/page.
Agentic Plus outperforms both coding agents while costing no more than half as much.
These results show why extraction quality and cost must be evaluated jointly: greater spending does not necessarily produce greater accuracy.
A broader comparison of commercial VLMs is provided in \Cref{app:model-family-quality-cost}.

\subsection{Results Across Dimensions}
\label{sec:main-results}

Overall F1 makes it easy to compare systems, but a single aggregate score cannot show which document characteristics drive their successes and failures.
To expose these failure modes, we use ExtractBench's challenge tags to break down performance across five axes: document length, task challenge, perception challenge, table structure, and business domain (\Cref{tab:res-length}).

\paragraph{Document length.}
Most systems perform well on short documents, with more than half scoring above 90\%, but the spread widens as documents grow longer.
On long documents the commercial VLMs fall below 40\%, while Claude Code Opus 4.8 (\ccLong\%) and Reducto Deep Extract (\rdLong\%) remain close to their short-document scores.
LlamaExtract Agentic Plus is the strongest across all three lengths and never drops below 90\% (\apShort{}/\apMedium{}/\apLong{}).
The long-document failure is concentrated in recall: entire records are dropped rather than misread (\Cref{app:pr-detail} reports precision and recall separately).
We attribute this to context limits: most systems cannot work through a long document in a single pass, and those without a strategy for iterating over it stop early, truncating the remaining records.

\paragraph{Task challenge.}
Long-list completeness (T1) and needle-in-haystack (T2) mirror the document-length results: LlamaExtract Agentic Plus and Reducto Deep Extract are the top two systems on both challenges.
Dense documents (T3) reorder the ranking.
Reducto Deep Extract drops to \rdFthree\%, Claude Code Opus 4.8 to \ccFthree\%, and Datalab Accurate Parse + Balanced Extract to 54.4\%.
The T3 scores combine several sources of difficulty: form layout, document classification, reviewer annotations, and schema size.
LlamaExtract Agentic Plus (\apFthree\%) and Codex GPT-5.5 (\czFthree\%) lead on this challenge.
T3.e is the 35-document subset with schemas above 150 leaf fields.
\Cref{app:challenge-detail} reports its scores and describes which pipelines rejected these documents.
\Cref{app:failure-analysis} reports pipeline-level success rates and failure causes.
Qualitative examples for T1--T3 are provided in \Cref{app:qualitative}.

\paragraph{Perception challenge.}
The perception axis exposes system-specific blind spots.
Codex GPT-5.5 handles rotated or image-only capture (P1) poorly, dropping to \czDegraded\% from \czScan\%--\czHand\% on the other perception challenges.
Reducto Deep Extract shows the complementary weakness: it stays above 90\% on rotated or image-only capture and on handwriting (P3), but falls to \rdScan\% on scanned pages (P2).
Qwen3.6 35B-A3B is stronger on scanned pages and handwriting than the other VLMs, scoring above 92\% on both.
LlamaExtract Agentic Plus remains the strongest system across all three perception challenges.

\paragraph{Table structure.}
Enormous tables (S4, beyond a thousand rows) produce the sharpest separation in \Cref{tab:res-structure}.
Most systems stop early and return only a small fraction of the records: every VLM scores below 10\% on this slice, and Datalab Accurate Parse + Balanced Extract (\dlSfour\%) and Extend Max Context (\exmSfour\%) also fall sharply.
By contrast, LlamaExtract Agentic Plus (\apSfour\%), Reducto Deep Extract (\rdSfour\%), and Claude Code Opus 4.8 (\ccSfour\%) are the top three systems.
Cross-page tables (S3) pose a milder version of the same failure, where the difficulty is carrying the table structure across page breaks.
Pivoted layouts (S2) are the least discriminative of the structure slices, because most leading systems handle them well.

\paragraph{Business domains.}
Domain difficulty largely reflects the mix of task challenges inside each domain (\Cref{tab:res-domain}).
Finance (D1) and government (D3) carry the long-list (T1) and needle-in-haystack (T2) tasks; energy (D2) is dominated by scanned dense forms (T3); and legal (D7) and healthcare (D6) hold long record lists, including creditor matrices, sanctions lists, and clinical event logs.
A domain's score is therefore mostly a reweighting of its task-challenge results, and we read the domain axis as a check on coverage rather than as an independent source of difficulty.

\subsection{The Grounding Gap}
\label{sec:grounding-gap}

Grounding makes extraction auditable by letting a reviewer trace each predicted value back to its source.
ExtractBench measures this capability explicitly,
whereas existing schema-guided extraction benchmarks overlook it (\Cref{tab:benchmark-capabilities}).
We consider a field grounded only when both the extracted value and its citation are correct, at the page- or word-level.
We highlight the grounding gap in \Cref{tab:grounding}.
\begin{itemize}
  \item \textbf{VLMs and coding agents} do not return evidence by default; they therefore score zero at both grounding levels. Users who need auditable outputs must add a separate evidence-localization component or use an extraction API with grounding built in.
  \item \textbf{Granularity challenge.} Locating the exact word is much more challenging than finding the correct page. LlamaExtract Agentic Plus achieves \apGPageOverall\% page-level grounding F1 but only \apGOverall\% word-level F1. Datalab shows a larger gap, at \dlGPageOverall\% versus \dlGOverall\%. Page evidence narrows the search, but still leaves reviewers to locate the value among similar candidates.
  \item \textbf{Robustness to length.} Extend Max Context falls from \exmGPageLone\% page-level grounding F1 on short documents to \exmGPageLthree\% on long documents, and Datalab follows the same pattern. Reducto Deep Extract is more stable, declining from \rdGPageLone\% to \rdGPageLthree\%, while LlamaExtract Agentic Plus remains the strongest system overall.
\end{itemize}

Even the best overall word-level grounding F1 is only \apGOverall\%. Systems are increasingly capable of extracting values and often identifying their source pages, but reliably connecting each value to its exact supporting evidence remains an open problem.

\section{Related Work}
\label{sec:related-work}

\subsection{Benchmarks for Document Extraction}

Document extraction benchmarks cover two main settings: fixed-ontology extraction, where fields are defined in advance, and \emph{schema-guided extraction}, where the user supplies a schema defining the fields and output structure at inference time.

\paragraph{Fixed-ontology document IE.}
Fixed-ontology benchmarks ask how reliably a system can recover a known set of fields as the documents become more challenging.
They have progressively expanded document diversity from forms and receipts to multilingual layouts and enterprise domains~\cite{jaume2019funsd,huang2019sroie,park2019cord,xu2022xfund,townsend2025realkienoveldatasetsenterprise}.
Other work increases structural and contextual complexity through line items, tables, long documents, and unfamiliar templates~\cite{wang2023vrdu,simsa2023docile,stanislawek2021kleister,hendrycks2021cuad,huybrechts2025documenthaystack}.
Some benchmarks additionally annotate spatial positions or study localization~\cite{simsa2023docile,toles2025formgym}.
Across these settings, however, the target fields remain fixed by the benchmark: they test robustness within a known ontology rather than whether a system can follow a new user-supplied schema.

\paragraph{Schema-guided extraction benchmarks.}
Recent document extraction benchmarks have focused on the schema-guided setting, where users specify the extraction task at inference time without retraining the system~\cite{ferguson2026extractbench,sibue2026exstructiny,ji2026unikie}.
Work in this setting has progressively increased task scale and complexity, testing more complex schemas, longer documents, and larger outputs.
ContextualAI's ExtractBench~\cite{ferguson2026extractbench} stresses schema complexity, with schemas containing up to 369 fields, but evaluates only five shared schemas; LongExtractBench-50~\cite{micro1longextract} and VAREX~\cite{varex2026} use a different schema for every document, so they do not test whether one extraction task transfers across diverse document appearances.
LongArray-Extract~\cite{extend2026longarray} and LongExtractBench-50~\cite{micro1longextract} instead stress completeness over long documents and repeated records, though their public test sets contain only dozens of documents.
Several benchmarks~\cite{varex2026,ji2026unikie,extend2026longarray} use synthetic construction to scale these evaluations, but their generated documents do not capture the visual variability and perception challenges found in real enterprise data.
This fragmented coverage makes it difficult to compare system families comprehensively or diagnose why they fail.

To our knowledge, ExtractBench provides the broadest combined coverage of these dimensions, spanning real document families and targeted synthetic stress tests across \ndomains{} business domains (\Cref{tab:benchmark-capabilities}).
ExtractBench is designed around production requirements at scale, jointly measuring value accuracy, source grounding, and per-page cost to capture whether outputs are correct, traceable, and economical to produce.
We also stratify the dataset with challenge tags to ensure coverage across task and perception difficulties and diagnose where different system families fail.

\subsection{Methods for Document Extraction}

Modern schema-guided extraction systems fall into three broad families: general-purpose vision-language models that generate outputs directly, coding agents that inspect documents iteratively with tools, and specialized extraction systems designed around document processing workflows.
These approaches make different tradeoffs in completeness, visual robustness, grounding, and cost.

\paragraph{Vision-language models.}
General-purpose vision-language models are multimodal reasoners that accept text and images and generate flexible outputs.
Document extraction can therefore be reformulated as multimodal generation: document pages are rendered as images, the schema is expressed as text instructions, and the model returns extracted values either as prompted text or as schema-compliant output enforced through a structured-output API.
Systems in this family include closed general-purpose models~\cite{openai2026gpt54nano,google2026gemini35flash}, general-purpose open-weight models~\cite{qwen2026qwen36,google2026gemma4}, and models tuned specifically for extraction~\cite{datalab2026lift,numind2026nuextract3}.
All three groups accept new schemas without task-specific retraining.
This direct, one-pass workflow is simple and efficient, but can miss values in long repeated structures, and the systems evaluated here do not return source evidence.

\paragraph{Coding agents.}
Coding agents such as Codex~\cite{openai2026codex,openai2026gpt55} and Claude Code~\cite{anthropic2026claudecode,anthropic2026opus48} approach extraction through an iterative tool-use loop: given the document and schema as files, they can inspect pages, write parsing code, run checks, and revise the final JSON.
This loop is more flexible than one-pass generation and can help with long documents or repeated records, where the agent can revisit the document rather than rely on a single model response.
The same flexibility creates cost and reliability risks: even short documents may trigger many inspection, coding, debugging, and validation steps, and unconstrained agents can run for many steps before producing a small extraction.
Coding agents also require an agent runtime with filesystem access and validation, depend on how documents are rendered and which tools are available, and are not designed around extraction-specific grounding metadata.

\paragraph{Specialized document extractors.}
Commercial platforms~\cite{reducto2026deep,extend2026maxcontext,datalab2026extract,llamaindex2026llamaextract} expose extraction as a purpose-built API rather than a raw model prompt or coding environment.
As managed services, they let users upload files directly and handle format support, preprocessing, document parsing, and schema-guided extraction with little configuration.
They can also expose visual grounding and other extraction metadata, such as source pages and, in some cases, word-level boxes, making outputs easier to audit than raw model responses.
However, specialized APIs still differ substantially in completeness, robustness, grounding quality, and cost.

\section{Conclusion}
\label{sec:conclusion}

We introduced ExtractBench, a challenge-tagged benchmark that brings the core requirements of real schema-guided extraction into one evaluation: correct and complete outputs, source traceability, robustness across document challenges, and cost at scale.
By measuring these dimensions together, ExtractBench shows not only which systems perform well, but where and why they fail.

Direct VLM extraction is inexpensive but often truncates long record lists; coding agents are more robust on these workloads but substantially more expensive.
Specialized APIs span the quality--cost frontier, with LlamaExtract Agentic Plus achieving the strongest overall performance at a lower cost than the coding agents.
Challenge-tagged results further show that systems degrade differently on long documents, dense schemas, scans, handwriting, and enormous tables.

Grounding remains the clearest area for improvement.
The evaluated VLMs and coding agents do not return source evidence by default, while word-level grounding F1 remains at \apGOverall\% even for specialized systems that return boxes.
Together, these results set a clear bar for reliable extraction: complete outputs, traceable evidence, and sustainable cost at scale.

\newpage
\begingroup
\small
\bibliographystyle{plainnat}
\bibliography{references}
\endgroup

\newpage

\onecolumn
\appendix

\etocsetnexttocdepth{subsection}
\etocsettocstyle{\section*{Appendix Contents}\small}{}
\etocsetlocaltop.toc{part}
\localtableofcontents

\section{Benchmark Details}
\label{app:benchmark-details}
\subsection{Taxonomy and Coverage Reference}
\label{app:taxonomy}

The consolidated taxonomy table defines every tag: what it stresses,
representative document types, and tagged document and page coverage.
The tags are shared with the capability comparison
(\Cref{tab:benchmark-comparison}) and with every result slice in
\Cref{sec:results}; \Cref{fig:taxonomy-coverage} in the main text shows the
challenge-level coverage as a distribution.

\begin{table}[p]
\centering
\footnotesize
\setlength{\tabcolsep}{4pt}
\renewcommand{\arraystretch}{1.0}
\setlength{\abovecaptionskip}{4pt}
\begin{tabular}{@{}>{\raggedright\arraybackslash}p{0.21\textwidth}>{\raggedright\arraybackslash}p{0.24\textwidth}>{\raggedright\arraybackslash}p{0.29\textwidth}rr@{}}
\toprule
\textbf{Tag} & \textbf{What it stresses} & \textbf{Representative document types} & \textbf{Docs} & \textbf{Pages} \\
\midrule
\multicolumn{5}{@{}l}{\textbf{Task Challenges}} \\
\addlinespace[1pt]
\textbf{T1.a}~ single long table & one homogeneous table spanning pages $\rightarrow$ array & SEC 13F holdings, fund schedules, registers & 99 (26.8\%) & 2681 (55.1\%) \\
\textbf{T1.b}~ cross-page continuation & records continue past page breaks under repeated headers & GSA labor schedules, IRS Schedule I, clinical logs & 91 (24.6\%) & 3081 (63.3\%) \\
\textbf{T1.c}~ repeated complex region & each record a multi-field block, not a table row & OFAC SSI list, service lists, bankruptcy E/F & 27 (7.3\%) & 492 (10.1\%) \\
\textbf{T1.d}~ pivoted / matrix & entities down and values across columns, with the header not at top & auto valuations, census cross-tabs, election pivots & 42 (11.4\%) & 717 (14.7\%) \\
\textbf{T1.e}~ packed / multi-row cell & one record spans sub-rows, or one cell packs many fields & FTX / iMedia creditor matrices, CBP 7501 & 30 (8.1\%) & 1173 (24.1\%) \\
\textbf{T2.a}~ sparse in narrative & few target fields buried in long prose & DD1155, CLIN schedules, SF1449 & 15 (4.1\%) & 617 (12.7\%) \\
\textbf{T2.b}~ paraphrased match & pick the canonical occurrence of a paraphrased value & earnings decks, investor presentations & 12 (3.2\%) & 356 (7.3\%) \\
\textbf{T2.c}~ dedup across modalities & reconcile base vs.\ modified copies & contract modification sets & 5 (1.4\%) & 222 (4.6\%) \\
\textbf{T2.d}~ cross-ref / reconciliation & combine / check values across sections & SEFA schedules, audit reconciliations & 24 (6.5\%) & 492 (10.1\%) \\
\textbf{T3.a}~ dense form & labeled cells, checkboxes, blanks on a short form & RRC oil-and-gas forms, CBP 7501, closing disclosures & 194 (52.4\%) & 798 (16.4\%) \\
\textbf{T3.b}~ receipt / invoice & line-item business documents & invoices, receipts, purchase orders & 17 (4.6\%) & 46 (0.9\%) \\
\textbf{T3.c}~ classify then extract & route by document class before extracting & brokerage statement families & 3 (0.8\%) & 61 (1.3\%) \\
\textbf{T3.d}~ filer-reviewer separation & separate original filer entries from later regulator annotations & administratively reviewed W-14 filings & 13 (3.5\%) & 13 (0.3\%) \\
\textbf{T3.e}~ large schema & schema contains more than 150 leaf fields & W-2, reviewed W-14, and Form 1040 bundles & 35 (9.5\%) & 421 (8.6\%) \\
\addlinespace[3pt]
\multicolumn{5}{@{}l}{\textbf{Perception Challenges}} \\
\addlinespace[1pt]
\textbf{P1}~ rotated / image-only & rotated or skewed page image, no text layer & scan-degraded re-captures & 38 (10.3\%) & 634 (13.0\%) \\
\textbf{P2}~ scanned & scanned page image & scanned regulatory forms & 134 (36.2\%) & 654 (13.4\%) \\
\textbf{P3}~ handwriting & handwriting on the page & hand-completed form fields & 55 (14.9\%) & 71 (1.5\%) \\
\addlinespace[3pt]
\multicolumn{5}{@{}l}{\textbf{Table Structure}} \\
\addlinespace[1pt]
\textbf{S1}~ merged headers & hierarchical headers & banded financial statements & 35 (9.5\%) & 721 (14.8\%) \\
\textbf{S2}~ header not at top / pivoted & pivoted layout & valuation and census matrices & 42 (11.4\%) & 717 (14.7\%) \\
\textbf{S3}~ cross-page table & continues across pages & long procurement schedules & 91 (24.6\%) & 3081 (63.3\%) \\
\textbf{S4}~ enormous table & very large table & 13F holdings, unclaimed-property lists & 12 (3.2\%) & 1106 (22.7\%) \\
\textbf{S5}~ table within a cell & nested table & creditor address blocks & 30 (8.1\%) & 1173 (24.1\%) \\
\addlinespace[3pt]
\multicolumn{5}{@{}l}{\textbf{Document Length}} \\
\addlinespace[1pt]
\textbf{L1}~ short & up to 10 pages & receipts, single-page forms, short filings & 252 (68.1\%) & 615 (12.6\%) \\
\textbf{L2}~ medium & 11--50 pages & multi-page statements, mid-size filings & 98 (26.5\%) & 2438 (50.1\%) \\
\textbf{L3}~ long & more than 50 pages & registers, holdings, long schedules & 20 (5.4\%) & 1816 (37.3\%) \\
\addlinespace[3pt]
\multicolumn{5}{@{}l}{\textbf{Business Domain}} \\
\addlinespace[1pt]
\textbf{D1}~ Finance & financial disclosures, holdings, and tax records & 13F / N-PORT, fund schedules, 1040, W-2, K-1, 1099-B & 145 (39.2\%) & 1956 (40.2\%) \\
\textbf{D2}~ Energy & regulatory forms and filer-reviewer annotations & Texas RRC W-1, W-2, W-14, 2A, P-4, P-18, H-5 & 98 (26.5\%) & 145 (3.0\%) \\
\textbf{D3}~ Government & procurement, customs, and public reporting & CBP 7501, GSA labor, IRS-990, SEFA, CLIN / SF-1449 & 49 (13.2\%) & 1328 (27.3\%) \\
\textbf{D4}~ Automotive & valuation reports and comparison tables & CCC / Mitchell total-loss valuations & 27 (7.3\%) & 377 (7.7\%) \\
\textbf{D5}~ Supply Chain & transactional documents and itemized records & invoices, receipts, rate cards, product specs, utility bills & 20 (5.4\%) & 54 (1.1\%) \\
\textbf{D6}~ Healthcare & remittance and clinical-event records & remittance advice, adverse-event / deviation logs & 15 (4.1\%) & 421 (8.6\%) \\
\textbf{D7}~ Legal & filings, creditor matrices, and entity lists & bankruptcy schedules, creditor matrix, sanctions list & 10 (2.7\%) & 562 (11.5\%) \\
\textbf{D8}~ Real Estate & mortgage closing disclosures & TRID mortgage closing disclosure & 6 (1.6\%) & 26 (0.5\%) \\
\bottomrule
\end{tabular}
\caption{ExtractBench taxonomy and coverage. Each row gives a task challenge, perception challenge, table structure, size, or business-domain slice, with representative documents and tagged document and page coverage. Percentages are shares of the 370-document, 4{,}869-page benchmark. Tags may overlap.}
\label{tab:tax-challenge}
\label{tab:tax-modality}
\label{tab:tax-structure}
\label{tab:tax-length}
\label{tab:tax-domain}
\end{table}

\begin{figure}[H]
\centering
\includegraphics[width=\textwidth]{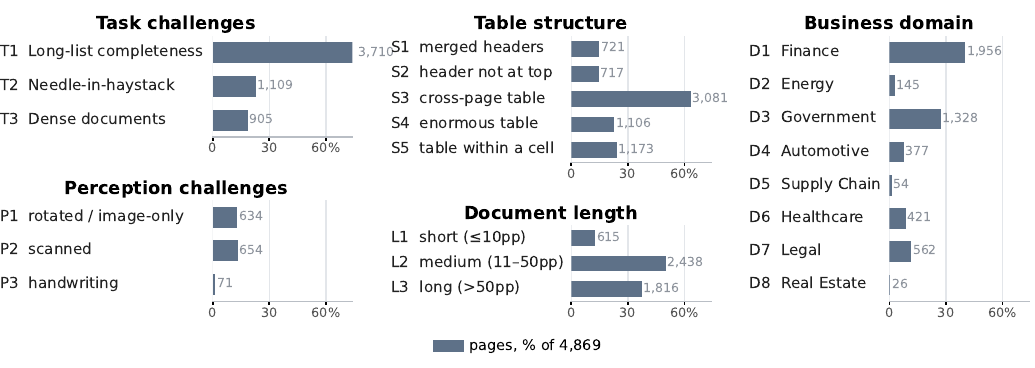}
\caption{ExtractBench coverage by page share. Each bar is the share of the
\statnum{\evalpages}{} corpus pages carrying the tag, with its page count shown beside the bar.
Tags may overlap across panels; \Cref{tab:tax-challenge} defines every tag and sub-tag.}
\label{fig:taxonomy-pages}
\end{figure}

\paragraph{Task-challenge notes.}
Partial credit can hide F1 errors: a system can emit a well-formed array that is missing a third of
its rows.
T1 covers a large share of the benchmark and holds records at large scale: a real SEC 13F table
with \statnum{\secthirteenfrows}{} holdings rows, a bankruptcy creditor matrix with
\statnum{\creditorrows}{} address-block records, and an unclaimed-property list with
\statnum{\unclaimedrows}{} rows.
T2 failures include selecting the canonical value of a KPI that recurs many times under paraphrase,
or a target field buried in procurement narrative.
On T3 documents, field localization remains a major error source even with dedicated
tools~\cite{toles2025formgym}.
The T3.d slice is \formreviewed{} administratively reviewed W-14 filings, whose twin schema asks
for the original filer value and the later regulator annotation separately rather than merging the
two.
For T3.e, the W-2 schema has 152 leaf fields (4 documents), the reviewed W-14 schema has 175
(13 documents), and the Form 1040 schemas have 1{,}343--1{,}368 (18 bundles). Because the threshold
is strictly greater than 150 leaf fields, the tag covers 35 documents and 421 pages and overlaps other T3 tags.

\paragraph{Grounding tags.}
An extraction is auditable only if each value points back to where it came from, so a separate set of
tags records the required evidence: a box around each extracted value (G1) and a box on a checkbox
together with the boolean read from it (G4).
G2 covers a field that expands into many records, and G3 covers deeply nested objects and arrays.
These tags constrain the output rather than the document; \Cref{tab:tax-delivery} gives the coverage
of each.
\Cref{sec:grounding-gap} scores grounding; to support these tags, the ground truth of
\Cref{sec:ground-truth} pairs every value with a location.

\par\medskip\noindent
\begin{minipage}{\columnwidth}
\footnotesize
\setlength{\tabcolsep}{4pt}
\centerline{\begin{tabular}{@{}p{0.35\columnwidth}p{0.40\columnwidth}r@{}}
\toprule
\textbf{Grounding (G)} & \textbf{What it requires} & \textbf{Docs} \\
\midrule
G1~ value-level box & box at each extracted value & 237 (64.1\%) \\
G2~ 1:N cardinality & one field expands to many records & 36 (9.7\%) \\
G3~ deep nesting & deeply nested objects and arrays & 8 (2.2\%) \\
G4~ checkbox / boolean box & read and locate a checkbox & 182 (49.2\%) \\
\bottomrule
\end{tabular}}
\captionof{table}{Grounding tags: what the output must carry besides the value---a box at each value, cardinality, nesting, and checkbox handling. Document counts are shares of the benchmark and may overlap. Grounding is scored in \Cref{sec:grounding-gap}.}
\label{tab:tax-delivery}
\end{minipage}
\par\medskip

\subsubsection{Document and Ground-Truth Size by Task Challenge}
\label{app:compression}

\begin{figure}[t]
\centering
\includegraphics[width=\textwidth]{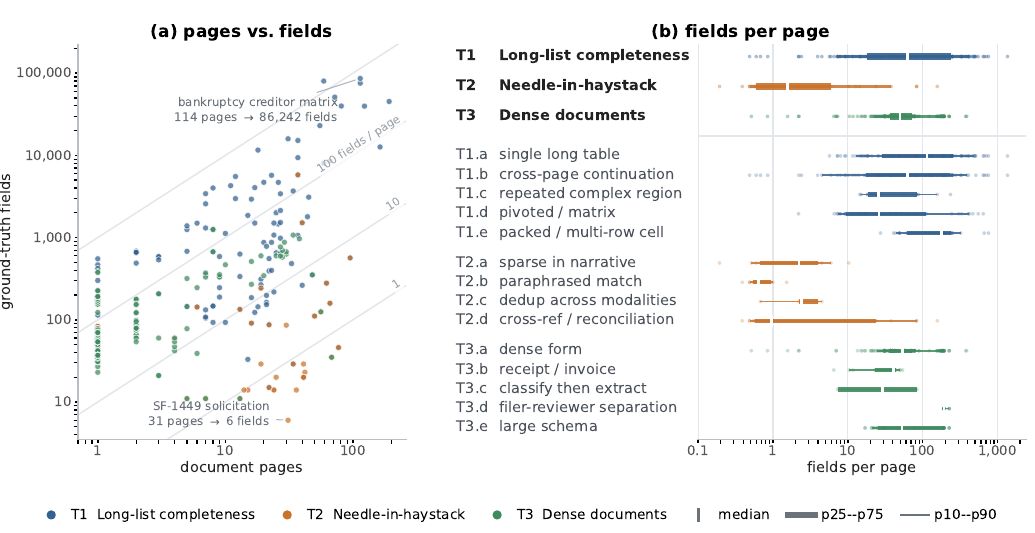}
\caption{Document size against ground-truth size, by task challenge. A
\emph{field} is one cell of the unified value metric: one per scalar field and
per record subfield (\Cref{sec:headline}). A document's field count is therefore
the number of cells a system must return correctly for full recall on it.
\textbf{(a)} One point per document, over the \cmpdocs{} pool documents that are
not scan-degraded re-captures; a document tagged with more than one task challenge
appears once per challenge. The diagonals mark a constant number of fields per page.
\textbf{(b)} The same rate by task challenge, then by sub-tag in code order
beneath a dividing rule, with the p10--p90 range, the p25--p75 range, and the
median; a document carrying several sub-tags appears in each of its rows.
\Cref{tab:tax-challenge} gives the document count behind every row.
Document size is counted in pages rather than in document text because most pages
of the form and scan-degraded slices carry no text layer.
The \cmpdegraded{} scan-degraded re-captures are omitted: a re-capture keeps its
original's schema, values, and page count, so it lands on the same point.}
\label{fig:compression}
\end{figure}

The task challenges are defined by what makes extraction hard
(\Cref{sec:taxonomy}). They also separate quantitatively, by how many fields a
document yields per page (\Cref{fig:compression}). Across the pool that rate
spans a factor of \statnum{\cmpratiospan}, from \cmpratiolo{} fields per page to
\statnum{\cmpratiohi}.
T3.e uses the number of leaf fields in the requested schema. The vertical axis in
\Cref{fig:compression} instead counts scored ground-truth cells, so repeated records add cells without
adding schema fields.

\paragraph{T2 is the only task challenge that compresses.}
Needle-in-haystack documents yield a median of \cmpftworatio{} fields per page: a
median of \cmpftwopages{} pages read for \cmpftwofields{} fields returned, so the
system reads a long document and keeps almost none of it. The extreme case is an
SF-1449 solicitation whose schema asks for \cmpsparsefields{} fields across
\cmpsparsepages{} pages. Both other task challenges return more than an order of
magnitude more fields per page.

\paragraph{T1 and T3 return dozens of fields per page.}
Long-list documents yield a median of \cmpfoneratio{} fields per page and dense
documents \cmpfthreeratio{}; the output re-encodes most of the document rather
than summarizing it. \cmpoverthousand{} documents carry more than
\statnum{1000}{} ground-truth fields and \cmpovertenk{} more than
\statnum{10000}{}, up to \statnum{\cmpmaxfields}{} fields in a
\cmpmaxfieldspages-page bankruptcy creditor matrix. A system that reads such a
document correctly still has to return every one of them, which is the long-list
completeness failure of \Cref{sec:main-results}: precision stays high while
recall falls.

\paragraph{The rate alone does not define the taxonomy.}
T1 and T3 overlap in \Cref{fig:compression}b and separate by scale instead: the
median T1 document is \cmpfonepages{} pages against \cmpfthreepages{} for T3, so
the same field density arrives either as one dense page or as tens of pages of
records. This is why the benchmark tags task challenge and document length on
independent axes (\Cref{sec:taxonomy}) rather than collapsing both into a single
difficulty score.

\subsection{Corpus Composition}
\label{app:composition}

This appendix details the corpus by document length. Every document carries exactly one length class,
so the regulatory and tax forms, the automotive valuations, and the scan-degraded re-captures are
described inside the class their page count puts them in rather than as separate slices.
Each re-capture pairs one-to-one with a clean original that also appears in the benchmark, under the
same schema and the same expected values (\Cref{app:gt-corrupt}).

\begin{table}[H]
\centering
\small
\setlength{\tabcolsep}{8pt}
\begin{tabular}{@{}lrrrr@{}}
\toprule
\textbf{Benchmark} & \textbf{Domains} & \textbf{Schemas/types} & \textbf{Documents} & \textbf{Max pages} \\
\midrule
ContextualAI EB~\cite{ferguson2026extractbench} & 5 & 5 & 35 & 218 \\
Extend LongArray~\cite{extend2026longarray} & 3 & 3 & 45 & 235 \\
Micro1 LongExtract-50~\cite{micro1longextract} & 7 & 50 & 50 & 11{,}622 \\
VAREX~\cite{varex2026} & 1 & 1{,}798$^{\dagger}$ & 1{,}798 & 1 \\
DocILE~\cite{simsa2023docile} & 1 & 1 & 6{,}680 & 3 \\
VRDU~\cite{wang2023vrdu} & 2 & 2 & 2{,}556 & 12 \\
RealKIE~\cite{townsend2025realkienoveldatasetsenterprise} & 5 & 5 & 1{,}867 & 198 \\
Kleister~\cite{stanislawek2021kleister} & 2 & 2 & 3{,}318 & 368 \\
CUAD~\cite{hendrycks2021cuad} & 1 & 1 & 510 & 154 \\
Legacy KIE~\cite{jaume2019funsd,huang2019sroie,park2019cord,xu2022xfund} & 2 & 4 & 3{,}565 & 1 \\
\midrule
\textbf{ExtractBench} & 8 & \textbf{67} & 370 & 192 \\
\bottomrule
\end{tabular}
\caption{Dataset scale and diversity. Document and page counts show scale. Domain and schema counts show breadth. The schema figures are not directly comparable: $^{\dagger}$\,VAREX uses one schema per synthetic single-page form, Micro1 uses a model-drafted schema per document, and ExtractBench reuses a document-type schema across documents.}
\label{tab:benchmark-breadth}
\end{table}

\subsubsection{Short Documents ($\le$10 pages)}
\label{app:comp-short}
\label{app:comp-form}

\lonedocs{} documents, \statnum{\lonepages}{} pages: \shortdocs{} multi-domain business documents,
\formshortdocs{} regulatory and tax forms, \autoshortdocs{} automotive total-loss valuations, and
\corruptlenshort{} scan-degraded re-captures.

Real families span finance, commodity business documents, government and customs forms, healthcare remittance advice, mortgage closing disclosures, product spec sheets, and auto total-loss valuations. Examples include SEC 13F and N-PORT holdings, invoices and receipts, CBP-7501 entry continuations, and census and budget cross-tabs with header-not-at-top pivots.
Synthetic re-renders add a bankruptcy Schedule~E/F with contingent/unliquidated/disputed booleans and election statement-of-votes pivots whose positional vote arrays align to per-contest candidate columns.
These short documents concentrate same-type identifiers and value--label disambiguation: electric-versus-gas meter IDs on one bill, vendor versus customer versus account numbers on one invoice, and structurally identical money fields whose meaning differs by section.
The automotive reports are pivoted comparison documents: the vehicle under appraisal runs down the
page while comparable vehicles run across it, with the header beside the data rather than above it.
They are the densest concentration of the pivoted/matrix structure (S2) in the benchmark.

The forms are Texas Railroad Commission energy filings (drilling permits, well completions,
injection and disposal permits, enhanced-oil-recovery designations, H$_2$S certificates, pressure
tests, plug records, transportation authorities, and skim-oil reports) and federal tax documents
(Form 1040 returns, W-2 wage statements, Schedule K-1, and 1099-B pages). They span \statnum{\formpages}{}
pages, from 1950s typewritten-on-scan filings to current born-digital output, and carry
\formtypes{} distinct frozen schemas across thirteen form families, because Form 1040 has a separate
schema per tax year and Schedule K-1 has separate partnership and S-corporation schemas. All but the
longer Form 1040 bundles are ten pages or fewer; the \mediumformdocs{} that are not appear in
\Cref{app:comp-medium}.
What they stress is dense labelled cells, checkbox banks, blanks that must come back null, handwriting
and scan noise on the older filings, and identifiers of the same shape whose meaning differs by
section. Every document carries the whole filed form, nothing cropped.
A human annotator reviews every form field by field; the tax forms additionally pass through
schema-first adjudication and an audit against the rendered pages.
Of the \statnum{\formeligiblerules}{} human-verified evidence rules across these documents,
\statnum{\formbboxev}{} ($\sim$\formbboxpct\%) carry a human-placed value box. Most of the
remainder are fields the form leaves blank, which have nothing on the page to box.

All form documents are public records.
The energy forms are regulatory filings served by the Texas Railroad Commission's public records. The tax forms come from public releases: 1040 returns released in full by public officials, W-2 wage statements from a public utility district's employer reference-copy register, and K-1 and 1099-B schedule pages from publicly filed documents.
Personal taxpayer identifiers were masked in those releases themselves --- employee and taxpayer SSNs appear on the page as masked strings (e.g., \texttt{XXX-XX-XXXX}), and the ground truth expects the masked string --- while the remaining identifiers, such as employer EINs, are business identifiers printed on public filings.

\subsubsection{Medium Documents (11--50 pages)}
\label{app:comp-medium}

\ltwodocs{} documents, \statnum{\ltwopages}{} pages: \mediumdocs{} multi-domain reports,
\mediumformdocs{} Form 1040 bundles, \automediumdocs{} automotive valuations, and
\corruptlenmedium{} scan-degraded re-captures.

Real families include financial-KPI documents---earnings decks, annual-report extracts, and press releases---scored against a master KPI schema that requires canonical-occurrence selection and GAAP/non-GAAP reconciliation. Other examples are SEFA single-audit schedules with hierarchical headers, SEC N-PORT holdings with nested detail, IRS Schedule I grant tables with multi-line addresses across pages, GSA labor-rate schedules with isolated pivots, merged brokerage/1099 statements that require classification, and auto total-loss valuations with a bookout matrix and option grid.
Synthetic re-renders add clinical protocol-deviation logs with per-entry timestamped comment threads (deep nesting) and county audit lists whose record blocks reconcile to multiple rollup levels.
The Form 1040 bundles that land here are the benchmark's largest schemas, carrying optional schedules
and absent sub-forms that must come back null.

\subsubsection{Long Documents ($>$50 pages)}
\label{app:comp-long}
\label{app:comp-corrupt}

\lthreedocs{} documents, \statnum{\lthreepages}{} pages: \longdocs{} multi-domain filings and
registers, and \corruptlenlong{} scan-degraded re-captures.

Real documents include an SEC 13F information table (\statnum{\secthirteenfrows}{} holdings rows), a government CLIN schedule (62 pages, sparse fields in narrative plus base-versus-modification deduplication), an SF1449 solicitation (95 pages, line items leaking across page breaks), and a DD1155 schedule continuation (56 pages, sparse fields plus handwritten signatures).
Synthetic re-renders include a bankruptcy creditor matrix (\statnum{\creditorrows}{} street-address blocks, packed cells at scale), an unclaimed-property list (\statnum{\unclaimedrows}{} rows, the benchmark's extreme truncation stress), and a rotated-landscape clinical adverse-event listing (163 pages), all with exact page and word-level box evidence by construction.

\subsection{Annotation Methodology in Detail}
\label{app:methodology}

This appendix gives the complete procedures for the three ground-truth methods in \Cref{sec:ground-truth}.

\paragraph{Evidence lists.}
Ground truth records each field as an \emph{evidence list}: the expected value, any alternate defensible readings, and for each reading a source page and, where reviewed, a word-level box. Scoring accepts a prediction matching \emph{any} listed reading (\emph{OR-acceptance}), and every expected record still counts against recall.
A field's evidence list holds more than one entry in two cases: a value can be cited from several places (a drug name that appears in the title, the indications paragraph, and a dosage table), or a genuinely ambiguous field has more than one defensible reading, each carrying its own box.
The annotator ratifies the list, and OR-acceptance scores predictions against it (\Cref{sec:metrics}).

\subsubsection{Real Documents}
\label{app:gt-real}

\paragraph{Pipeline.}
For each document family the workflow is as follows.
(1)~Collect representative PDFs for the family (invoices, remittance advice, KPI reports, \dots) and confirm membership.
(2)~Draft or refine the schema. Every field description carries aliases, format requirements, location hints, and do-not-confuse guidance.
(3)~Run multiple extraction systems from different model and pipeline families against the same schema. A smaller cross-extract trio drives the schema-discovery loop, and a broader ensemble drives verification.
(4)~Compare outputs cell by cell, aligning repeated rows by a declared identity key (invoice line ID, claim number, VIN, check number) rather than by list position.
(5)~Two coding agents independently inspect the PDF and classify each disagreement as schema ambiguity, model failure, or unresolvable.
(6)~Build candidate ground truth from cells the extraction systems agreed on and disputed cells both coding agents confidently classify as model failures. Block all other disputed cells.
(7)~Re-score cached extraction outputs against the candidate ground truth before promoting it.
(8)~\emph{Human check and fix}: An annotator reviews every blocked cell, including any cell for which either coding agent is unsure, the agents disagree, or both classify it as schema-ambiguous or unresolvable. The annotator checks each against the PDF and makes the final classification. A schema ambiguity returns to step (2) for a revised field description and another system run; a model failure is verified or corrected before the cell is added to candidate ground truth; a cell that remains unresolvable stays blocked.

\paragraph{Two operating modes.}
The same machinery serves both: \emph{schema discovery} creates a new family, and \emph{ground-truth review} audits an existing one by hunting cells where many pipelines fail against the current truth, verifying them against the source PDF, and patching a copied dataset.

\subsubsection{Synthetic Long Lists}
\label{app:gt-synthetic}

\paragraph{Pipeline.}
(1)~Choose a real long-list layout pattern (repeated rows, sectioned registers, continuation pages, totals, nested records); the layout may be a fund schedule, a holdings register, or a creditor matrix sampled from real templates.
(2)~Build the full structured content first (records, fields, nulls, totals, hierarchy, normalization rules).
(3)~Render the content into a realistic PDF in that pattern.
(4)~Paginate by \emph{measurement}: rendered blocks are measured and packed into pages by actual size, so a long record takes more space, a heading stays attached to its item, and no page stops early. Fixed rows-per-page pagination creates artificial page breaks and wrong source-page labels.
(5)~Derive ground truth from the render: locate every record and field in the final PDF and assign source pages, quotes, and word-level boxes from what is actually visible.
(6)~Validate consistency across PDF, expected JSON, schema, evidence, page references, and boxes.
(7)~Audit semantically by running several extraction systems and inspecting aggregated disagreements for schema or generation defects that mechanical checks cannot see.

\subsubsection{Scanned Forms}
\label{app:gt-forms}

\paragraph{Five stages.}
(0)~\emph{Corpus}: Select documents per form type.
(1)~\emph{Schema} (agent-only): Draft from the blank form template, check the fit against real scans, tighten over a small refinement loop, and freeze before any document is labeled. A schema revision after aggregation begins requires a version bump and targeted re-review, never silent reinterpretation of old votes.
(2)~\emph{Aggregate and adjudicate}: An ensemble of up to five systems votes per schema leaf and contested leaves go to adjudication against the rendered page.
(3)~\emph{Human review}: An annotator reviews the proposed value and box for each field, then accepts, edits, nulls, or redraws it.
(4)~\emph{Post-QA}: The reviewed ground truth is evaluated against the same systems again to identify remaining inconsistencies.

\paragraph{Value consensus.}
Because the schema is frozen, every pipeline answers the identical question per leaf. Omitting a path is a null vote, not an abstention.
Votes are normalized by the field's comparator (case, whitespace, date, boolean) before grouping, and array rows align across pipelines by declared identity key, never list position.
Each leaf lands in a tier: \emph{unanimous}, \emph{majority}, \emph{split} (plurality proposal, queued for adjudication), or \emph{all-null}.
A null majority over a real minority value is also queued, since the ensemble is built to catch silent field loss.
The ensemble draws on models from several independent families, so no one system decides a value on its own.
Adjudication load scales with the size of the schema: small forms converge with no adjudicated fields at all, while the largest need a handful per document.

\paragraph{Adjudication rules.}
The adjudicator must inspect the rendered page before ruling.
Checkboxes are two-state wherever their page is in the filing: the verdict is \emph{true} (a mark is present) or \emph{false} (no mark, including when the box is not printed on this copy), and a \emph{false} verdict needs no supporting quote since there is nothing on the page to cite. A checkbox on a page the filing does not include is null.

\paragraph{One-source boxes.}
Value votes benefit from redundancy, but agreement among predicted boxes does not verify the cited location.
Bounding boxes are never merged across systems.
Each field's box comes from a single citation-emitting pipeline in the pool. A field with a nullish value needs no box, and a field with no valid citation is handed to the annotator to draw.

\subsubsection{Scan-Degraded Re-Captures}
\label{app:gt-corrupt}

\corruptdocs{} documents are degraded re-captures from the sources above. Their expected values
stay the same, so the difference between clean and degraded scores measures the effect of capture degradation.
This slice is also what populates the P1 rotated / image-only tag of \Cref{tab:tax-modality}: each
document appears twice, clean and degraded, under the same schema and the same expected values, so
only the capture differs.

\paragraph{Pipeline.}
Each of the \corruptdocs{} documents (\corruptreal{} real and \corruptsynth{} synthetic) is rendered to
page images, slightly rotated, given a slight perspective shift, and passed through one
scan recipe from a fixed library. The recipes include photocopier and carbon-copy tone curves, fax
thresholding, sensor and speckle noise, phone-camera capture, aging and bleed-through, low-resolution
resampling, dust, and shadowed copying. Each document's recipe and seed are recorded, so it can be
regenerated byte-for-byte.
Recipes that warp the page non-rigidly (creases, book curvature, elastic deformation) are excluded,
because their effect on a word box cannot be written down in closed form.

\paragraph{Ground truth.}
Values, schema, and tags are copied from the clean document unchanged. Boxes are stored normalized
to the page, which makes them invariant to every photometric effect and to uniform rescaling, so
only the rotation and perspective steps move them; those are applied to the box corners in closed
form.

\subsection{Full Capability Comparison}
\label{app:comparison-detail}

The capability summary of the introduction merges related taxonomy tags into
single rows. The matrix below scores every tag separately, using the tags
defined in \Cref{app:taxonomy}. Benchmark columns are grouped by task
specification: schema-guided extraction supplies the target schema at evaluation
time, whereas fixed-ontology benchmarks predefine the labels. A filled circle
denotes covered and scored, a hollow circle partial or incidental coverage, and
a blank absence. Each mark is verified against the benchmark's primary source.
Legacy KIE combines FUNSD, SROIE, CORD, and XFUND. \Cref{tab:benchmark-capabilities} gives the main-text summary,
including domain breadth and measured cost.

\begin{table}[p]
\centering
\footnotesize
\setlength{\tabcolsep}{4.2pt}
\renewcommand{\arraystretch}{1.15}
\begin{tabular*}{\textwidth}{@{\extracolsep{\fill}}l>{\columncolor{oursband}}ccccccccccc@{}}
\toprule
& \multicolumn{5}{c}{\textbf{Schema-guided}} & \multicolumn{6}{c}{\textbf{Fixed ontology}} \\
\cmidrule(lr){2-6}\cmidrule(lr){7-12}
\textbf{Dimension} & \rotatebox{90}{\textbf{ExtractBench}} & \rotatebox{90}{ContextualAI EB~\cite{ferguson2026extractbench}} & \rotatebox{90}{Extend LongArray~\cite{extend2026longarray}} & \rotatebox{90}{Micro1 LongExtract-50~\cite{micro1longextract}} & \rotatebox{90}{VAREX~\cite{varex2026}} & \rotatebox{90}{RealKIE~\cite{townsend2025realkienoveldatasetsenterprise}} & \rotatebox{90}{DocILE~\cite{simsa2023docile}} & \rotatebox{90}{VRDU~\cite{wang2023vrdu}} & \rotatebox{90}{Legacy KIE~\cite{jaume2019funsd,huang2019sroie,park2019cord,xu2022xfund}} & \rotatebox{90}{Kleister~\cite{stanislawek2021kleister}} & \rotatebox{90}{CUAD~\cite{hendrycks2021cuad}} \\
\midrule
\textit{T1: long-list completeness} & & & & & & & & & & & \\
\quad \textbf{T1.a}~ single long table & \yesours & \yes & \yes & \yes & \half & \half & \yes & \half & \half & \no & \no \\
\quad \textbf{T1.b}~ cross-page continuation & \yesours & \half & \yes & \yes & \no & \no & \half & \no & \no & \no & \no \\
\quad \textbf{T1.c}~ repeated complex region & \yesours & \yes & \half & \half & \half & \no & \no & \yes & \yes & \no & \no \\
\quad \textbf{T1.d}~ pivoted / matrix & \yesours & \no & \no & \half & \no & \no & \no & \no & \no & \no & \no \\
\quad \textbf{T1.e}~ packed / multi-row cell & \yesours & \no & \no & \no & \no & \no & \half & \no & \half & \no & \no \\
\textit{T2: needle-in-haystack} & & & & & & & & & & & \\
\quad \textbf{T2.a}~ sparse in narrative & \yesours & \half & \no & \half & \no & \yes & \no & \no & \no & \yes & \yes \\
\quad \textbf{T2.b}~ paraphrased match & \yesours & \half & \no & \no & \no & \half & \half & \yes & \no & \yes & \yes \\
\quad \textbf{T2.c}~ dedup across modalities & \yesours & \no & \half & \no & \no & \no & \no & \no & \no & \half & \no \\
\quad \textbf{T2.d}~ cross-ref / reconciliation & \yesours & \no & \half & \no & \no & \no & \no & \no & \half & \half & \half \\
\textit{T3: dense documents} & & & & & & & & & & & \\
\quad \textbf{T3.a}~ dense form & \yesours & \half & \no & \no & \yes & \half & \half & \yes & \yes & \no & \no \\
\quad \textbf{T3.b}~ receipt / invoice & \yesours & \no & \no & \no & \no & \yes & \yes & \half & \yes & \no & \no \\
\quad \textbf{T3.c}~ classify then extract & \yesours & \no & \no & \no & \no & \no & \no & \no & \no & \no & \half \\
\quad \textbf{T3.d}~ filer-reviewer separation & \yesours & \no & \no & \no & \no & \no & \no & \no & \no & \no & \no \\
\quad \textbf{T3.e}~ large schema & \yesours & \half & \no & \half & \no & \no & \no & \no & \no & \no & \no \\
\textit{Perception challenges} & & & & & & & & & & & \\
\quad \textbf{P1}~ rotated / image-only & \yesours & \half & \no & \no & \half & \half & \no & \half & \half & \half & \no \\
\quad \textbf{P2}~ scanned & \yesours & \no & \no & \no & \no & \yes & \half & \half & \yes & \yes & \no \\
\quad \textbf{P3}~ handwriting & \yesours & \no & \no & \no & \no & \half & \no & \half & \half & \half & \no \\
\textit{Table structure} & & & & & & & & & & & \\
\quad \textbf{S1}~ merged headers & \yesours & \no & \no & \half & \no & \no & \no & \half & \half & \no & \no \\
\quad \textbf{S2}~ header not at top / pivoted & \yesours & \no & \no & \half & \no & \no & \no & \no & \no & \no & \no \\
\quad \textbf{S3}~ cross-page table & \yesours & \half & \yes & \yes & \no & \half & \half & \yes & \no & \half & \half \\
\quad \textbf{S4}~ enormous table & \yesours & \half & \yes & \yes & \no & \no & \no & \no & \no & \no & \no \\
\quad \textbf{S5}~ table within a cell & \yesours & \no & \no & \no & \no & \no & \no & \no & \yes & \no & \no \\
\textit{Grounding \& output trust} & & & & & & & & & & & \\
\quad \textbf{G1}~ value-level box & \yesours & \no & \no & \no & \no & \half & \yes & \yes & \yes & \no & \half \\
\quad \textbf{G2}~ 1:N cardinality & \yesours & \yes & \yes & \yes & \yes & \yes & \yes & \yes & \yes & \half & \yes \\
\quad \textbf{G3}~ deep nesting & \yesours & \half & \half & \half & \half & \no & \no & \yes & \half & \no & \no \\
\quad \textbf{G4}~ checkbox / boolean box & \yesours & \no & \no & \no & \no & \no & \no & \no & \no & \no & \no \\
\bottomrule
\end{tabular*}
\vspace{2pt}
{\scriptsize\raggedright \yes~= covered and scored \quad \half~= partial or incidental \quad blank = absent.\par}
\caption{Full comparison of benchmark task specification and capability coverage.}
\label{tab:benchmark-comparison}
\end{table}

\clearpage
\section{Metric Details}
\label{app:metric-detail}

This appendix gives the exact matching, normalization, and aggregation rules behind the unified
value F1 of \Cref{sec:headline}. Scoring is fully deterministic: the same predictions and ground
truth always produce the same score, with no model in the loop.

\subsection{Cell Matching and Normalization}
\label{app:normalization}

Two cells are compared under the following rules, in order; the first rule that applies decides.
\begin{enumerate}[leftmargin=1.4em,itemsep=1pt]
  \item \emph{Date canonicalization.} Before any comparison, every string on either side that
  matches one of eight common date formats (\texttt{2019-03-28}, \texttt{3/28/2019},
  \texttt{3/28/19}, \texttt{3-28-2019}, \texttt{March 28, 2019}, \texttt{Thursday March 28 2019},
  \texttt{28 March 2019}, and hyphenated variants) is rewritten to ISO \texttt{YYYY-MM-DD}.
  Guards keep non-dates intact: candidates shorter than 4 or longer than 50 characters,
  all-digit strings, strings containing a run of ten or more digits, and parses outside
  1900--2100 pass through unchanged.
  \item \emph{Long-list comparability.} So that our long-list numbers can be read against
  Extend's LongArray benchmark, we adopt its public reference scorer verbatim, including the two
  free-text fields it compares by edit-distance ratio rather than
  exactly~\cite{extend2026longarray}. No other field in the benchmark uses fuzzy matching.
  \item \emph{Strings.} Whitespace runs collapse to single spaces and outer whitespace is
  trimmed; the comparison is then exact and case-sensitive. There is no punctuation stripping,
  unicode folding, or currency/thousands handling.
  \item \emph{Everything else.} Plain equality. Numbers carry no tolerance, and a number never
  equals its string rendering (\texttt{"1,000"} $\ne$ \texttt{1000}). A list of scalars inside a
  record compares as one opaque, order-sensitive value; order-invariance applies to records, not
  to scalar lists.
\end{enumerate}

Two acceptance layers apply after exact matching. Both are declared in the ground truth before
scoring and applied identically to every system.
First, a scanned-form field can declare \emph{opt-in leniencies} where the printed template makes
one strict reading unfair (\Cref{tab:normalizers}).
Second, by OR-acceptance over the evidence list (\Cref{app:methodology}), a prediction is correct
when it matches the expected value or any recorded alternate reading.

\begin{table}[H]
\centering
\small
\setlength{\tabcolsep}{8pt}
\begin{tabular}{@{}ll@{}}
\toprule
\textbf{Leniency} & \textbf{Rule} \\
\midrule
\texttt{null\_equals\_false} & a blank checkbox may read as \texttt{false} or as null \\
\texttt{case\_insensitive} & casefolded comparison (typewriter and stamp case) \\
\texttt{optional\_terminal\_punctuation} & one trailing \texttt{.\,,\,;\,:} is ignored on each side \\
\texttt{punctuation\_spacing} & whitespace around \texttt{.\,,\,;\,:} is insignificant \\
\texttt{phone\_digits} & phone numbers compare by their last ten digits \\
\texttt{lenient\_date} & split preprinted years rejoin (``19\,55''), two-digit years try both centuries \\
\bottomrule
\end{tabular}
\caption{Opt-in, per-field leniencies for scanned forms. Each is declared in the ground
truth for specific fields and applies only after the exact match misses, so a leniency can never
turn a passing cell into a failure.}
\label{tab:normalizers}
\end{table}

\subsection{Missing-Value Semantics}
\label{app:null-semantics}

Every scalar field of the schema enters both the precision and the recall denominator, and a key
absent from the output is scored identically to an explicit null. \Cref{tab:null-matrix} lists
every case. Two consequences follow: a hallucinated value on a blank field costs both
precision and recall (the cell sits in both denominators), and correctly returning null for a
blank field is credited, so a system cannot be hurt by faithful nulls.

\begin{table}[H]
\centering
\small
\setlength{\tabcolsep}{8pt}
\begin{tabular}{@{}lll@{}}
\toprule
\textbf{Expected} & \textbf{Predicted} & \textbf{Outcome} \\
\midrule
value & matching value & correct (counts toward P and R) \\
value & different value & miss in both P and R \\
value & null or key omitted & miss in both P and R \\
null & null or key omitted & correct (counts toward P and R) \\
null & value & miss in both P and R (hallucination) \\
\addlinespace[2pt]
records & fewer records & each missing record's cells miss in R only \\
records & extra / duplicated records & each extra record's cells miss in P only \\
null or \texttt{[]} & \texttt{[]} or null & no cells contributed \\
\bottomrule
\end{tabular}
\caption{Scoring outcome for every (expected, predicted) state. Scalar cells are
symmetric between precision and recall; only repeated records move the two sides apart, which is
what makes the precision--recall split of \Cref{app:pr-detail} a truncation diagnostic.}
\label{tab:null-matrix}
\end{table}

\subsection{Array Alignment}
\label{app:alignment}

Records pair by a globally optimal one-to-one assignment (\texttt{linear\_sum\_assignment}, the
Hungarian family) whose cost between an expected and a predicted record is the number of
mismatched declared subfield cells; minimizing total cost is equivalent to maximizing total field
agreement over the array. The assignment is not greedy, consults no identity keys, and record
order never affects the value score. Rows whose cells match exactly are pre-paired by a hash join
as a provably score-preserving fast path. Duplicated predictions pair at most once; the surplus
copies count only against precision. A record subfield that is itself a list of records recurses
with an independent assignment per matched pair.

\subsection{Aggregation and Grounding}
\label{app:aggregation}

The per-document score is micro precision/recall/F1 over the document's cell bag; slice and
overall scores are unweighted means of per-document values, so a document with more fields does
not weigh more, and a slice's mean F1 is not the harmonic mean of its mean P and R. A document a system fails to
return, whether it rejects the schema, errors, or returns no JSON object, scores zero rather than
being dropped, so every system is averaged over the same document set and a refusal is penalized
like any other miss.

A word-level box tightly encloses the cited word or a short span of adjacent words, rather than the surrounding table cell.
Grounding metrics gate on value correctness: a cell is grounded-correct when its value is
accepted and a predicted citation box overlaps any evidence box for that field at
IoU $\ge 0.5$ on the correct page. The grounding precision denominator counts only gradeable
claims (citations on cells aligned to box-bearing ground truth), the recall denominator counts
ground-truth cells that carry a verified box, and documents with no verified boxes emit no
grounding score at all rather than zero. Page-level evidence replaces the box test with
page membership. For each document on which all three metrics are defined, unified value F1
$\ge$ page F1 $\ge$ word-level grounding F1.

\clearpage
\section{Evaluation Protocol}
\label{app:evaluation-protocol}
This section documents the prompts, configurations, and cost rules underlying the evaluation in the main text. \Cref{app:eval-config} specifies how each system is run, while \Cref{app:pricing} records the cost-accounting rules and provider rates.
\subsection{System Configuration}
\label{app:eval-config}

\noindent This subsection records the exact prompts and consequential configurations underlying the evaluation setup in \Cref{sec:setup}. The accompanying benchmark release contains the complete integrations and provider-specific options.

\paragraph{VLM APIs.}
GPT-5.4 Nano and Gemini 3.5 Flash receive the document and the same benchmark prompts.
\begin{quote}
\small
\textbf{System prompt.} \emph{``You are extracting structured data from a document according to the provided JSON schema. Return only the JSON that matches the schema. Use null for fields not present in the document. When the schema includes a list field, populate every relevant row visible in the document -- do not return an empty list when rows are present.''}
\end{quote}
\begin{quote}
\small
\textbf{User prompt.} \emph{``Extract every field from the attached document according to the schema. Return JSON only. Use null for fields not present in the document. Whenever the schema declares a list field, enumerate every row visible in the document -- do not collapse rows or return an empty list when rows are present.''}
\end{quote}
GPT-5.4 Nano uses the OpenAI Responses API with the target schema supplied as \texttt{text.format=json\_schema} and \texttt{strict=false}. Gemini 3.5 Flash uses \texttt{response\_json\_schema} with temperature zero and low thinking. For GPT, the harness recursively sets \texttt{additionalProperties=false}; for Gemini, it also promotes repeated-record nodes into ordinary JSON-Schema properties. These mechanical transformations do not change the requested fields or descriptions.

\paragraph{Self-hosted VLMs.}
Lift~9B passes the task schema through the official Lift SDK to vLLM guided-JSON decoding. Qwen3.6 35B-A3B uses vLLM's \texttt{json\_schema} response format with xgrammar-guided decoding. NuExtract3 receives a mechanical conversion of the task schema into its native extraction-template format rather than a generic JSON-Schema constraint. Gemma4~26B receives the schema in the prompt and uses vLLM's \texttt{json\_object} mode, which enforces valid JSON but not the complete schema. Qwen3.6 35B-A3B and Gemma4 use the benchmark prompts above; Lift and NuExtract3 use their model-native interfaces.

\paragraph{Coding agents.}
Claude Code Opus~4.8 and Codex GPT-5.5 run through their vendor CLIs in isolated working directories containing the staged document and target schema. Both may use local computation, are instructed to rely only on the provided materials, and have a 1,200-second per-document timeout. Codex uses low reasoning effort.

The shared task prompt below is quoted verbatim; \texttt{[DOCUMENT]} and \texttt{[SCHEMA]} mark the staged filename and full task schema inserted for each example.

\begin{samepage}
\begin{quote}
\small
\textbf{Shared Claude Code and Codex task prompt.} {\itshape ``Extract structured data from the document file `./[DOCUMENT]' in the current directory.

\medskip
Use only local file inspection and shell commands. Do not use web search, network calls, browser tools, or external services. Temporary scratch files inside the current directory are OK.

\medskip
Return a single JSON object conforming to this schema as your final answer:

\texttt{[SCHEMA]}

\medskip
Rules:

- Use null for fields not present in the document.

- For list/array fields, enumerate every relevant row visible in the document; never collapse rows.

- For large regular tables, prefer writing and running a local script to parse/enumerate rows.

- For forms, prefer direct field extraction from the document content.

- Write the resulting JSON object to ./output.json and validate that it is valid JSON before stopping.

- Do not print the JSON to your assistant output.''\par}
\end{quote}
\end{samepage}

\paragraph{Specialized extraction APIs.}
Specialized extraction systems receive the same document and schema without a benchmark-specific prompt. LlamaExtract Cost-Effective and Agentic use their matching parse tiers, explicit parse-first execution, source citations, and word-level bounding boxes; Agentic Plus uses per-document extraction with source citations and confidence scores. Reducto Deep Extract runs with citations and deep extraction enabled. Extend Max Context uses the \texttt{extraction\_performance} processor with citations, advanced figure parsing, and \texttt{large\_array\_max\_context}. Datalab uses Accurate Parse, Balanced Extract, and JSON output.

\subsection{Cost Accounting and Pricing}
\label{app:pricing}

This appendix records the list prices used to reconstruct commercial-system costs in \Cref{sec:cost}. We apply the public pay-as-you-go or standard-tier rates available as of July 1, 2026 to recorded or reconstructed token, credit, or page usage. Volume, committed-use, and enterprise discounts are not included.

Open-weight pipelines that we self-host (Lift~9B~\cite{datalab2026lift}, NuExtract3~\cite{numind2026nuextract3}, Qwen3.6 35B-A3B~\cite{qwen2026qwen36}, and Gemma4 26B~\cite{google2026gemma4}) have no vendor API price and are omitted. The \textcent/page figures in \Cref{sec:cost} are calculated from the token, credit, or page consumption of the benchmark runs. Because provider pricing can depend on observed usage, these measured costs may differ from a headline per-page rate.

\newpage
\paragraph{Token-metered systems.}
\Cref{tab:pricing-token} reports the token rates used for the VLM API and coding-agent runs. Codex GPT-5.5 is priced at the GPT-5.5 rates and Claude Code Opus~4.8 at the Opus~4.8 rates. OpenAI applies a 2$\times$ input and 1.5$\times$ output surcharge above 272K input tokens~\cite{openai2026gpt55model}. Anthropic's Opus~4.7+ tokenizer emits ${\sim}30\%$ more tokens for the same text~\cite{anthropic2026pricing}.

\begin{table}[H]
\centering
\footnotesize
\setlength{\tabcolsep}{8pt}
\begin{tabular}{@{}lrrr@{}}
\toprule
\textbf{Model} & \textbf{Input} & \textbf{Cached} & \textbf{Output} \\
\midrule
\multicolumn{4}{@{}l}{\textit{OpenAI}~\cite{openai2026pricing}} \\
\quad GPT-5.4 Nano   & 0.20 & 0.02 & 1.25 \\
\quad GPT-5.5        & 5.00 & 0.50 & 30.00 \\
\addlinespace[2pt]
\multicolumn{4}{@{}l}{\textit{Google}~\cite{google2026pricing}} \\
\quad Gemini 3.5 Flash & 1.50 & 0.15 & 9.00 \\
\addlinespace[2pt]
\multicolumn{4}{@{}l}{\textit{Anthropic}~\cite{anthropic2026pricing}} \\
\quad Opus 4.8       & 5.00 & 0.50 & 25.00 \\
\bottomrule
\end{tabular}
\caption{Token-metered list prices in USD per one million tokens, using the standard real-time tier. ``Cached'' denotes cache-hit input.}
\label{tab:pricing-token}
\end{table}

\paragraph{Managed document extraction APIs.}
These APIs typically separate document extraction into two stages: (1) parsing (i.e. transcribing) the document into a machine-readable representation, and (2) extracting the target fields from that representation. \Cref{tab:pricing-page} reports the corresponding parse and extract charges.

\begin{table}[H]
\centering
\footnotesize
\setlength{\tabcolsep}{8pt}
\begin{tabular*}{\textwidth}{@{\extracolsep{\fill}}lrllr@{}}
\toprule
\textbf{System} & \textbf{\textcent/credit} & \makecell{\textbf{Parse}\\\textit{credits/page}} & \makecell{\textbf{Extract}\\\textit{credits/page}} & \textbf{\textcent/page} \\
\midrule
\multicolumn{5}{@{}l}{\textit{LlamaExtract}~\cite{llamaindex2026llamaextract,llamaindex2026pricing}} \\
\quad Cost-Effective  & 0.125 & 3  & 5  & 1 \\
\quad Agentic         & 0.125 & 10 & 15 & 3.1 \\
\quad Agentic Plus    & 0.125 & 10 & 50 & 7.5 \\
\addlinespace[2pt]
\multicolumn{5}{@{}l}{\textit{Reducto}~\cite{reducto2026pricing,reducto2026credit}} \\
\quad Deep Extract    & 1.5 & 1--2 & $\max(30,\,4p + 0.1f)$ credits/document & variable \\
\addlinespace[2pt]
\multicolumn{5}{@{}l}{\textit{Extend}~\cite{extend2026pricing}} \\
\quad Max Context     & 1.25 & 2 & 6 & 10 \\
\addlinespace[2pt]
\multicolumn{5}{@{}l}{\textit{Datalab}~\cite{datalab2026extract,datalab2026pricing}} \\
\quad A$+$B & 1 & 1.0 & 2.5 & 3.5 \\
\bottomrule
\end{tabular*}
\caption{Page- and credit-metered list prices. Here, $p$ is the number of pages and $f$ the number of returned fields.}
\label{tab:pricing-page}
\end{table}

\noindent The provider-specific pricing rules are:
\begin{itemize}[leftmargin=*,itemsep=2pt,topsep=3pt]
    \item \emph{LlamaExtract Agentic Plus.} For large schemas, LlamaIndex applies a schema-size multiplier to the 50-credit \emph{extract} rate~\cite{llamaindex2026pricing}.
    \item \emph{Reducto Deep Extract.} Reducto charges $\max(30,\,4p + 0.1f)$ extract credits per document, plus 1--2 parse credits/page. At the lower parse rate, a one-page document costs at least 31 credits, or 46.5\,\textcent{}.
    \item \emph{Extend Max Context.} Extend's published base rate is 3 extract credits/page plus 2 parse credits/page~\cite{extend2026pricing}. Its \texttt{large\_array\_max\_context} strategy makes multiple passes at approximately twice the extract credits~\cite{extend2026maxcontext}; accordingly, we double the extract component and leave the parse component unchanged.
    \item \emph{Datalab A$+$B.} Datalab prices parse and extract directly in cents/page. To present both components in the same columns, we treat 1\,\textcent{} as one credit.
\end{itemize}

\clearpage
\section{Detailed Results}
\label{app:additional-results}

Using the same document-level aggregation and cost accounting as the main text,
this section extends the headline results in four directions.
\Cref{app:document-length-detail} examines quality, cost, precision, and recall
across document lengths; \Cref{app:challenge-detail} isolates fine-grained task
failures; \Cref{app:failure-analysis} separates completed extractions from
operational failures; and \Cref{app:model-family-quality-cost} compares
quality--cost scaling within commercial model families.

\subsection{Document-Length Analysis}
\label{app:document-length-detail}
\paragraph{Quality and Cost.}
\label{app:cost-detail}

\begin{figure}[H]
\centering
\begin{subfigure}[t]{0.49\textwidth}
  \includegraphics[width=\linewidth]{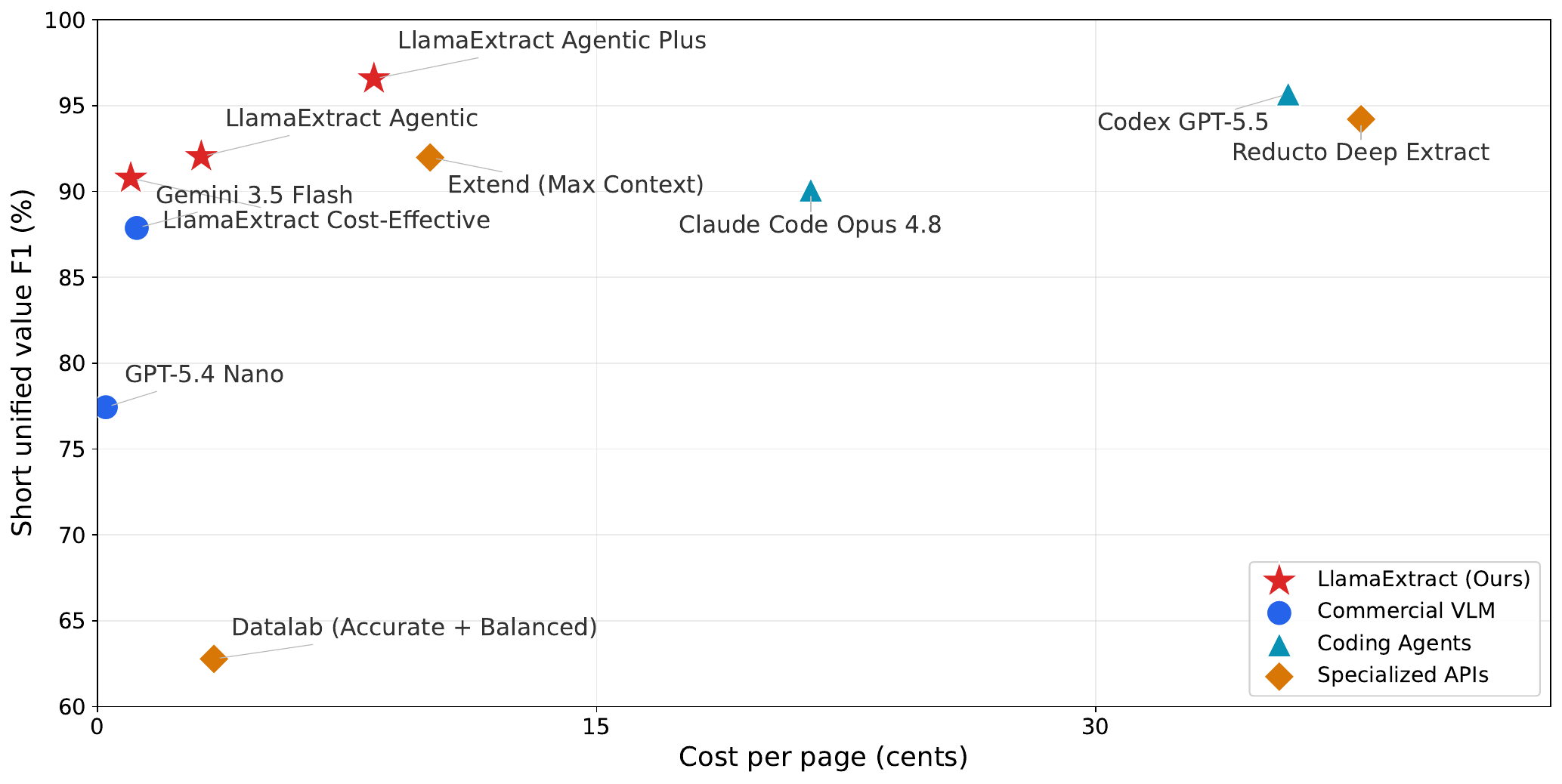}
  \caption{Short documents (L1)}\label{fig:qc-short}
\end{subfigure}\hfill
\begin{subfigure}[t]{0.49\textwidth}
  \includegraphics[width=\linewidth]{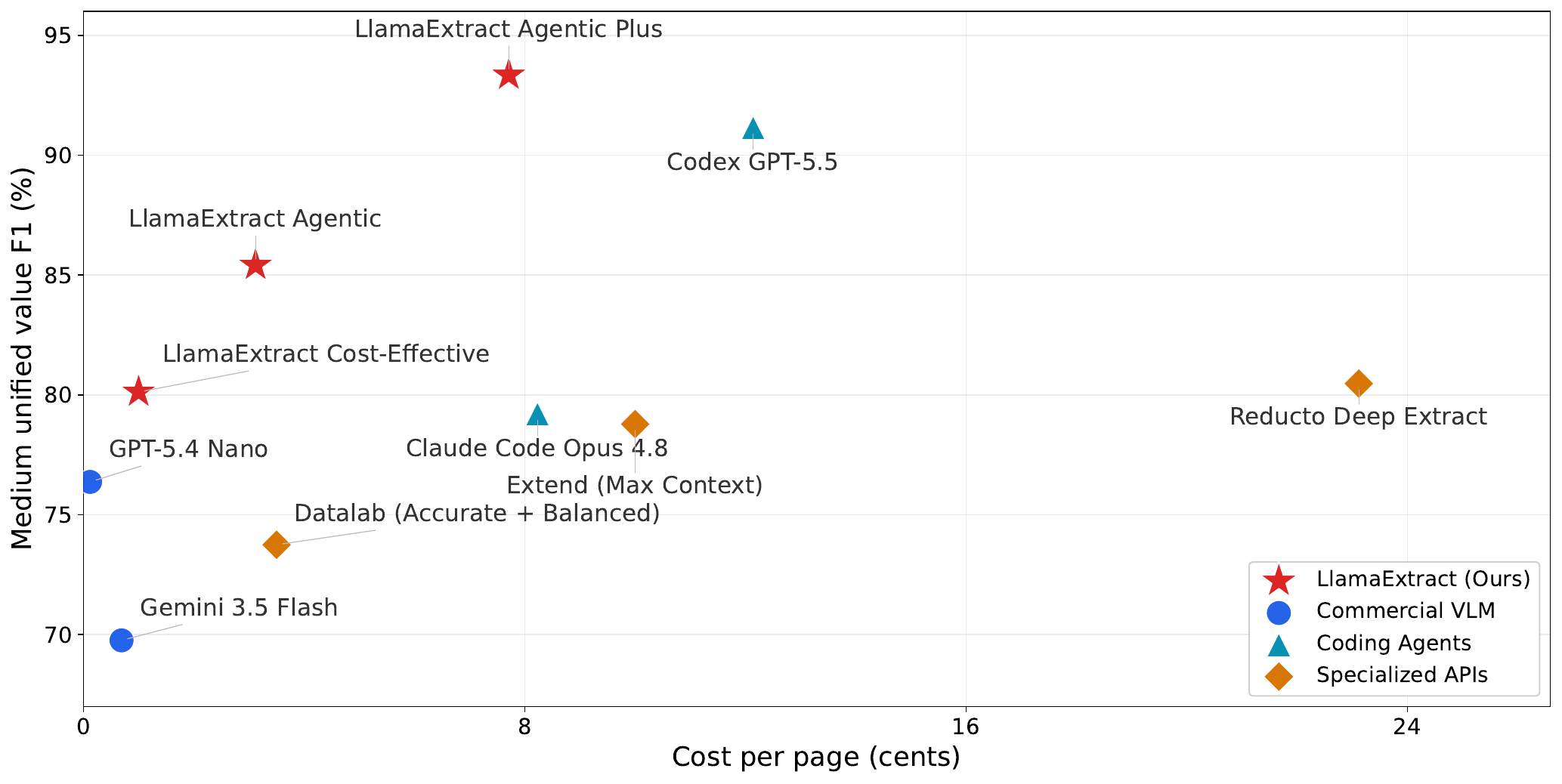}
  \caption{Medium documents (L2)}\label{fig:qc-medium}
\end{subfigure}\\[4pt]
\begin{subfigure}[t]{0.49\textwidth}
  \includegraphics[width=\linewidth]{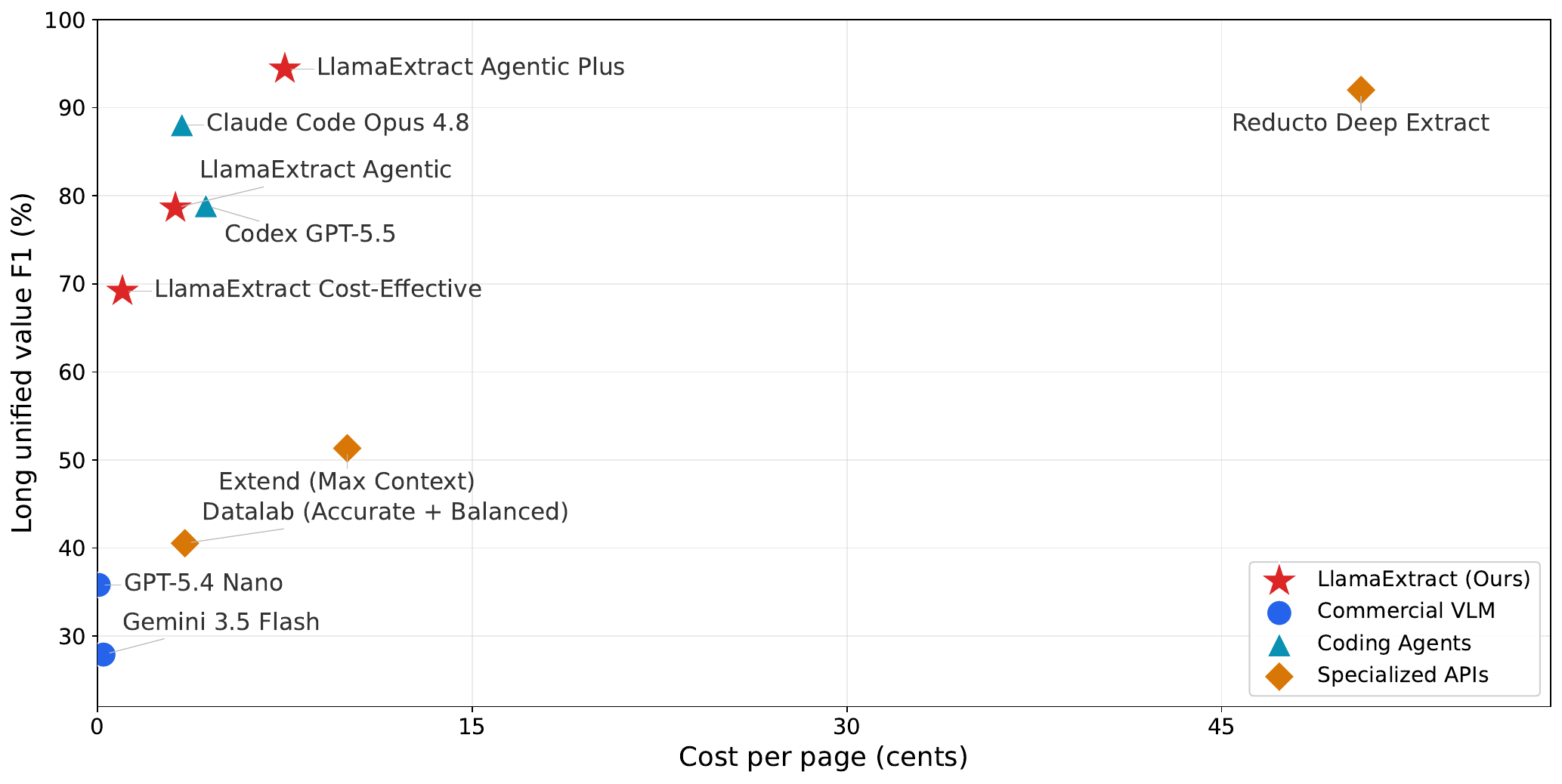}
  \caption{Long documents (L3)}\label{fig:qc-long}
\end{subfigure}
\caption{Unified value F1 versus measured per-page cost by document length (L1/L2/L3). Systems without a measured cost on a slice are omitted. The four OSS pipelines are open-weight and self-hosted.}
\label{fig:qc-per-category}
\end{figure}

\Cref{fig:qc-per-category} decomposes the overall quality--cost tradeoff of
\Cref{sec:cost} by document length. Each panel pairs unified value F1 with mean
per-page cost over the documents in that length slice. \Cref{tab:cost-per-page}
provides the exact costs.

On short and medium documents, the same progression defines the frontier:
GPT-5.4 Nano anchors the lowest-cost end, followed by LlamaExtract
Cost-Effective, Agentic, and Agentic Plus as quality and cost increase.
Agentic Plus remains the high-quality endpoint while costing less than the
coding agents and the most expensive specialized APIs.

Long documents reshape the tradeoff. The token-metered VLMs and coding agents
cost substantially less per page on L3 than on L1, consistent with
per-document overhead being spread across more pages. Their quality does not
scale uniformly, however: the one-shot VLMs degrade sharply, and Codex also
loses ground. Claude preserves more of its short-document quality and becomes a
competitive intermediate point, while Agentic Plus continues to anchor the
high-quality end. Reducto also remains accurate on long documents, but at a
substantially higher per-page cost.

Together, the panels separate two notions of scaling. A system can become
cheaper per page as documents grow while extracting a smaller fraction of the
requested records; robust long-document extraction requires both favorable
cost scaling and stable quality.

\begin{table}[H]
\centering
\footnotesize
\setlength{\tabcolsep}{10pt}
\begin{tabular}{@{}lrrrr@{}}
\toprule
\textbf{System} & \textbf{Overall} & \textbf{L1: short} & \textbf{L2: medium} & \textbf{L3: long} \\
\midrule
\multicolumn{5}{@{}l}{\textit{Specialized APIs}} \\
\quad LlamaExtract Agentic Plus & 8.1 & 8.3 & 7.7 & 7.5 \\
\quad LlamaExtract Agentic & 3.1 & 3.1 & 3.1 & 3.1 \\
\quad LlamaExtract Cost-Effective & 1.0 & 1.0 & 1.0 & 1.0 \\
\quad Datalab (Accurate + Balanced) & 3.5 & 3.5 & 3.5 & 3.5 \\
\quad Extend (Max Context) & 10.0 & 10.0 & 10.0 & 10.0 \\
\quad Reducto Deep Extract & 34.4 & 38.0 & 23.1 & 50.6 \\
\addlinespace[2pt]
\multicolumn{5}{@{}l}{\textit{Coding Agents}} \\
\quad Codex 5.5 & 27.8 & 35.8 & 12.1 & 4.3 \\
\quad Claude Code (Opus 4.8) & 16.2 & 21.4 & 8.2 & 3.4 \\
\addlinespace[2pt]
\multicolumn{5}{@{}l}{\textit{Commercial VLM}} \\
\quad GPT-5.4 Nano & 0.21 & 0.25 & 0.12 & 0.05 \\
\quad Gemini 3.5 Flash & 1.0 & 1.2 & 0.69 & 0.24 \\
\bottomrule
\end{tabular}
\caption{Mean document-level cost per page (\textcent/page), overall and by document length. Each column pools the documents in that slice; Overall pools all scored documents. Flat page-priced systems repeat their list price, while token- and credit-metered systems use recorded or reconstructed consumption (\Cref{app:pricing}). The four self-hosted OSS pipelines have no comparable vendor price and are omitted.}
\label{tab:cost-per-page}
\end{table}

\paragraph{Precision and Recall.}
\label{app:pr-detail}

Precision and recall expose different failures hidden by F1: missing records lower recall, while
duplicated or hallucinated records lower precision. \Cref{tab:pr-length} reports both and the signed
gap $\Delta=P-R$ by document length. The large positive L3 gaps for the commercial VLMs show that
returned values are often correct but many requested records are missing.

\begin{center}
\small
\renewcommand{\arraystretch}{0.88}
\setlength{\tabcolsep}{4pt}
\begin{tabular}{@{}lcccccccccccc@{}}
\toprule
 & \multicolumn{3}{c}{\textbf{Overall}} & \multicolumn{3}{c}{\textbf{L1}} & \multicolumn{3}{c}{\textbf{L2}} & \multicolumn{3}{c}{\textbf{L3}} \\
\cmidrule(lr){2-4}\cmidrule(lr){5-7}\cmidrule(lr){8-10}\cmidrule(lr){11-13}
\textbf{System} & P & R & $\Delta$ & P & R & $\Delta$ & P & R & $\Delta$ & P & R & $\Delta$ \\
\midrule
\multicolumn{13}{@{}l}{\textit{Specialized APIs}} \\
\quad LlamaExtract Agentic Plus & \textbf{95.8} & \textbf{95.4} & 0.4 & \textbf{96.6} & \textbf{96.5} & 0.1 & \textbf{94.0} & \textbf{92.8} & 1.2 & \textbf{94.3} & \textbf{94.5} & -0.2 \\
\quad LlamaExtract Agentic & 90.4 & 89.7 & 0.7 & 91.4 & 92.8 & -1.4 & 89.4 & 84.3 & \cellcolor{resultdroplight}5.1 & 82.5 & 77.2 & \cellcolor{resultdroplight}5.3 \\
\quad LlamaExtract Cost-Effective & 89.2 & 86.1 & 3.1 & 91.2 & 90.6 & 0.6 & 85.5 & 79.2 & \cellcolor{resultdroplight}6.3 & 80.7 & 63.4 & \cellcolor{resultdropmedium}17.3 \\
\quad Datalab (Accurate + Balanced) & 64.7 & 64.5 & 0.2 & 63.1 & 62.9 & 0.2 & 73.9 & 73.6 & 0.3 & 40.5 & 40.5 & 0.0 \\
\quad Extend (Max Context) & 86.0 & 86.8 & -0.8 & 92.0 & 91.9 & 0.1 & 77.4 & 80.9 & -3.5 & 51.2 & 51.5 & -0.3 \\
\quad Reducto Deep Extract & 90.5 & 90.5 & 0.0 & 94.1 & 94.4 & -0.3 & 80.8 & 80.2 & 0.6 & 92.0 & \underline{92.1} & -0.1 \\
\addlinespace[2pt]
\multicolumn{13}{@{}l}{\textit{Coding Agents}} \\
\quad Codex 5.5 & \underline{95.3} & \underline{93.2} & 2.1 & \underline{96.0} & \underline{95.5} & 0.5 & \underline{93.6} & \underline{90.2} & 3.4 & \underline{94.0} & 78.6 & \cellcolor{resultdropmedium}15.4 \\
\quad Claude Code (Opus 4.8) & 87.4 & 87.1 & 0.3 & 90.2 & 90.2 & 0.0 & 79.8 & 78.9 & 0.9 & 89.4 & 87.5 & 1.9 \\
\addlinespace[2pt]
\multicolumn{13}{@{}l}{\textit{OSS}} \\
\quad Gemma4 26B & 67.0 & 66.0 & 1.0 & 81.2 & 80.2 & 1.0 & 41.6 & 40.6 & 1.0 & 12.5 & 11.9 & 0.6 \\
\quad Qwen3.6 35B & 88.9 & 87.2 & 1.7 & 93.2 & 93.3 & -0.1 & 85.5 & 84.3 & 1.2 & 51.0 & 25.7 & \cellcolor{resultdropstrong}25.3 \\
\quad NuExtract3 & 64.2 & 45.0 & \cellcolor{resultdropmedium}19.2 & 66.5 & 51.0 & \cellcolor{resultdropmedium}15.5 & 61.2 & 37.4 & \cellcolor{resultdropmedium}23.8 & 50.1 & 5.8 & \cellcolor{resultdropstrong}44.3 \\
\quad Lift 9B & 78.8 & 77.2 & 1.6 & 87.4 & 87.2 & 0.2 & 65.1 & 62.4 & 2.7 & 37.4 & 24.2 & \cellcolor{resultdroplight}13.2 \\
\addlinespace[2pt]
\multicolumn{13}{@{}l}{\textit{Commercial VLM}} \\
\quad GPT-5.4 Nano & 77.8 & 75.1 & 2.7 & 77.4 & 78.2 & -0.8 & 80.3 & 75.5 & 4.8 & 71.2 & 33.7 & \cellcolor{resultdropstrong}37.5 \\
\quad Gemini 3.5 Flash & 84.5 & 79.5 & 5.0 & 88.2 & 87.7 & 0.5 & 75.1 & 69.3 & \cellcolor{resultdroplight}5.8 & 83.7 & 26.5 & \cellcolor{resultdropstrong}57.2 \\
\bottomrule
\end{tabular}
\captionof{table}{Value precision (P), recall (R), and signed gap $\Delta=P-R$ (percentage points), overall and by document length. Positive $\Delta$ means recall trails precision. Overall weights slices by scored-document count; L1/L2/L3 denote $\le$10, 11--50, and $>$50 pages. Entries are per-document means, so mean F1 is not implied by mean P and R. Bold and underline mark the top two P and R values; red shading marks gaps above 5/15/25 points (darker means larger).}
\label{tab:pr-length}
\end{center}

\par\vspace{0.75\baselineskip}
\subsection{Task-Challenge Analysis}
\label{app:challenge-detail}

\Cref{tab:res-challenge-detail} decomposes the three task-challenge families
reported in \Cref{tab:res-challenge} into their individual sub-tags. Because
these slices can overlap in documents, the rows should be interpreted
individually rather than averaged. The shading highlights where a system falls
below its own overall score, making it possible to distinguish broad weaknesses
from failures concentrated in a particular subtype.

\begin{table}[H]
\centering
\normalsize
\setlength{\tabcolsep}{0pt}
\renewcommand{\arraystretch}{0.9}
\begin{tabular}{@{}l>{\centering\arraybackslash}p{2.48em}>{\centering\arraybackslash}p{2.48em}>{\centering\arraybackslash}p{2.48em}>{\centering\arraybackslash}p{2.48em}>{\centering\arraybackslash}p{2.48em}>{\centering\arraybackslash}p{2.48em}>{\centering\arraybackslash}p{2.48em}>{\centering\arraybackslash}p{2.48em}>{\centering\arraybackslash}p{2.48em}>{\centering\arraybackslash}p{2.48em}>{\centering\arraybackslash}p{2.48em}>{\centering\arraybackslash}p{2.48em}>{\centering\arraybackslash}p{2.48em}>{\centering\arraybackslash}p{2.48em}@{}}
\toprule
 & \multicolumn{6}{c}{\makebox[0pt][c]{\makecell{\textbf{Specialized APIs}}}} & \multicolumn{2}{c}{\makebox[0pt][c]{\makecell{\textbf{Coding}\\\textbf{Agents}}}} & \multicolumn{4}{c}{\makebox[0pt][c]{\makecell{\textbf{OSS}}}} & \multicolumn{2}{c}{\makebox[0pt][c]{\makecell{\textbf{Commercial}\\\textbf{VLM}}}} \\
\cmidrule(lr){2-7}\cmidrule(lr){8-9}\cmidrule(lr){10-13}\cmidrule(lr){14-15}
\textbf{Challenge sub-tag} & \rotatebox{80}{\textbf{LE Agentic Plus}} & \rotatebox{80}{\textbf{LE Agentic}} & \rotatebox{80}{\textbf{LE Cost-Eff.}} & \rotatebox{80}{\textbf{Datalab A+B}} & \rotatebox{80}{\textbf{Extend Max}} & \rotatebox{80}{\textbf{Reducto Deep}} & \rotatebox{80}{\textbf{Codex GPT-5.5}} & \rotatebox{80}{\textbf{CC Opus 4.8}} & \rotatebox{80}{\textbf{Gemma4 26B}} & \rotatebox{80}{\textbf{Qwen3.6 35B-A3B}} & \rotatebox{80}{\textbf{NuExtract3}} & \rotatebox{80}{\textbf{Lift 9B}} & \rotatebox{80}{\textbf{GPT-5.4 Nano}} & \rotatebox{80}{\textbf{Gemini 3.5 Flash}} \\
\midrule
\textbf{Overall} & \textbf{95.6} & 89.5 & 86.8 & 64.5 & 86.3 & 90.4 & \underline{93.6} & 87.1 & 66.2 & 87.3 & 47.9 & 77.3 & 74.9 & 79.8 \\
\midrule
\multicolumn{15}{@{}l}{\textbf{T1: Long-list completeness}} \\
T1.a: single long table & \textbf{97.5} & 87.4 & 83.8 & 78.0 & 85.6 & \underline{96.0} & 92.4 & 95.1 & \cellcolor{resultdropmedium}47.5 & \cellcolor{resultdroplight}74.8 & \cellcolor{resultdroplight}35.9 & \cellcolor{resultdroplight}66.7 & 72.6 & \cellcolor{resultdroplight}74.4 \\
T1.b: cross-page continuation & \textbf{95.8} & \cellcolor{resultdroplight}84.3 & \cellcolor{resultdroplight}79.0 & 78.5 & 85.1 & \underline{94.4} & 89.4 & 92.5 & \cellcolor{resultdropstrong}40.5 & \cellcolor{resultdroplight}73.8 & \cellcolor{resultdroplight}37.6 & \cellcolor{resultdroplight}64.5 & 72.3 & \cellcolor{resultdroplight}73.6 \\
T1.c: repeated complex region & \textbf{93.9} & \cellcolor{resultdroplight}76.2 & \cellcolor{resultdroplight}75.4 & 84.2 & 88.3 & \underline{92.7} & \cellcolor{resultdroplight}84.1 & 91.6 & \cellcolor{resultdroplight}55.4 & 84.3 & \cellcolor{resultdroplight}34.4 & 72.8 & \cellcolor{resultdroplight}62.2 & 87.2 \\
T1.d: pivoted / matrix & \underline{95.0} & 86.1 & 87.2 & 84.4 & 91.4 & \textbf{95.3} & 94.9 & 94.3 & 63.0 & 89.6 & \cellcolor{resultdropstrong}20.9 & 76.4 & 77.7 & 88.4 \\
T1.e: packed / multi-row cell & \textbf{97.2} & 87.3 & \cellcolor{resultdroplight}78.2 & 71.7 & \cellcolor{resultdroplight}75.9 & \underline{95.4} & \cellcolor{resultdroplight}86.8 & 93.9 & \cellcolor{resultdropstrong}37.1 & \cellcolor{resultdropstrong}56.6 & 50.1 & \cellcolor{resultdropstrong}51.3 & \cellcolor{resultdroplight}67.3 & \cellcolor{resultdropstrong}53.4 \\
\midrule
\multicolumn{15}{@{}l}{\textbf{T2: Needle-in-haystack}} \\
T2.a: sparse in narrative & \textbf{95.9} & 92.9 & 88.8 & 79.5 & 93.7 & 93.5 & \underline{94.6} & 90.2 & 67.0 & 92.5 & \cellcolor{resultdropstrong}18.1 & 87.5 & 83.5 & 89.9 \\
T2.b: paraphrased match & \cellcolor{resultdroplight}90.3 & 87.7 & \cellcolor{resultdroplight}80.7 & \cellcolor{resultdroplight}51.1 & 84.2 & \underline{90.8} & \cellcolor{resultdroplight}86.6 & 85.0 & 62.7 & 86.2 & 44.8 & 78.9 & 79.5 & \textbf{93.0} \\
T2.c: dedup across modalities & \textbf{94.8} & 86.4 & 84.8 & 86.8 & 87.6 & \underline{92.0} & 90.2 & \cellcolor{resultdroplight}77.8 & 83.4 & 82.4 & \cellcolor{resultdropstrong}6.6 & 76.3 & 81.8 & 91.0 \\
T2.d: cross-ref / reconciliation & \textbf{92.2} & 85.5 & \cellcolor{resultdroplight}78.2 & 70.3 & 88.2 & \underline{91.8} & 89.9 & 88.5 & \cellcolor{resultdroplight}60.5 & \cellcolor{resultdroplight}80.8 & \cellcolor{resultdropmedium}29.6 & \cellcolor{resultdroplight}72.1 & \cellcolor{resultdroplight}68.0 & 86.6 \\
\midrule
\multicolumn{15}{@{}l}{\textbf{T3: Dense documents}} \\
T3.a: dense form & \textbf{95.4} & 92.1 & 90.3 & \cellcolor{resultdroplight}53.1 & 84.6 & 86.5 & \underline{95.1} & \cellcolor{resultdroplight}81.1 & 76.0 & 93.2 & 59.0 & 82.3 & 75.8 & 78.8 \\
T3.b: receipt / invoice & 97.4 & 92.7 & 95.0 & 65.5 & 96.8 & 97.2 & \underline{98.2} & \textbf{99.2} & 89.7 & 97.1 & 64.8 & 92.5 & 83.8 & 97.1 \\
T3.c: classify then extract & 96.5 & 90.8 & \cellcolor{resultdroplight}74.9 & 75.0 & 93.4 & 95.5 & \textbf{97.5} & \cellcolor{resultdroplight}76.7 & \cellcolor{resultdroplight}52.3 & \cellcolor{resultdropmedium}66.2 & \cellcolor{resultdropstrong}5.4 & \cellcolor{resultdroplight}68.8 & 72.7 & \underline{97.2} \\
T3.d: filer-reviewer separation & \cellcolor{resultdroplight}85.1 & \cellcolor{resultdroplight}82.2 & 82.9 & \cellcolor{resultdroplight}50.4 & \textbf{91.6} & \cellcolor{resultdroplight}85.1 & \underline{89.5} & \cellcolor{resultdropstrong}0.0 & 75.8 & 87.6 & 74.8 & 82.2 & 71.5 & \cellcolor{resultdropstrong}0.0 \\
T3.e: large schema & \cellcolor{resultdroplight}\underline{89.4} & 85.5 & \cellcolor{resultdroplight}80.8 & \cellcolor{resultdropmedium}40.4 & \cellcolor{resultdropstrong}44.6 & \cellcolor{resultdropstrong}42.0 & \textbf{90.3} & \cellcolor{resultdropstrong}10.9 & \cellcolor{resultdropstrong}36.1 & 87.8 & 65.4 & \cellcolor{resultdropstrong}39.1 & \cellcolor{resultdroplight}62.7 & \cellcolor{resultdropstrong}0.0 \\
\bottomrule
\end{tabular}%
\caption{Unified value F1 (\%) for every challenge sub-tag, with the same system columns as \Cref{tab:results-by-dimension}. Within each row, \textbf{bold} and \underline{underlined} mark the highest and second-highest scores. Red shading marks drops of more than 5/15/25 points from each system's overall score (darker means larger). Sub-tag slices overlap in documents; the task-challenge rows of \Cref{tab:res-challenge} score each challenge's document union instead of averaging these rows.}
\label{tab:res-challenge-detail}
\end{table}

\paragraph{Long-list completeness.}
Long-list extraction produces the clearest sustained separation between
systems. LlamaExtract Agentic Plus and Reducto Deep Extract remain strong
across both regular long tables and more difficult packed or multi-row layouts,
whereas several other systems degrade as the row structure becomes less
regular. The contrast across the T1 sub-tags suggests that the central
difficulty is not simply locating individual values, but preserving record
boundaries and associating cells correctly across continuations, repeated
regions, and compact layouts.

\paragraph{Needle-in-haystack.}
The T2 results are less uniform because the sub-tags test different retrieval
behaviors. Sparse narrative retrieval is broadly tractable for the strongest
systems, while paraphrased matching, cross-modal deduplication, and
reconciliation change the system ordering. In particular, Gemini leads on
T2.b, whereas LlamaExtract Agentic Plus and Reducto are strongest on T2.c and
T2.d. The family-level score therefore conceals meaningful differences between
locating a mention, recognizing a paraphrase, and selecting a canonical value.

\paragraph{Dense documents.}
Receipt and invoice extraction is strong across most systems. The scores spread
much more for classify-then-extract, filer-reviewer separation, and large
schemas. Visible density is only part of the problem; schema breadth and the
requested extraction operation matter too.

\medskip
\paragraph{Large-schema rejection.}
Large schemas are a deployment limit for many pipelines: seven systems score
below 50 F1 on T3.e, largely because large-schema rejections result in missing
outputs. Specifically, Gemini returns no output even for the 152-field W-2
schema and scores 0 on this entire challenge, while Claude Code completes W-2 but
fails on the larger W-14 and Form 1040 schemas in our harness. Lift, Gemma4,
Reducto, and Extend return no output for all 18 Form 1040 bundles. However,
Codex GPT-5.5, Qwen3.6 35B-A3B, and all LlamaExtract systems
complete every T3.e document and score above 80.

\subsection{Failure Modes Across Pipelines}
\label{app:failure-analysis}
\Cref{tab:failure-modes} reports each system's extraction success rate and
groups its failures by cause. Capacity or schema limits caused most failures
from Gemini, Claude Code, Qwen, Reducto, and Extend, and many from Lift. Most
failures from Datalab, NuExtract3, and Gemma4 were timeouts. Codex GPT-5.5 and
LlamaExtract Agentic Plus successfully extracted all 370 documents. Missing
outputs receive zero and lower the F1.

\begin{table}[H]
\centering
\small
\setlength{\tabcolsep}{10pt}
\begin{tabular}{@{}lrrrr@{}}
\toprule
\textbf{System} & \textbf{Success} & \makecell{\textbf{Capacity /}\\\textbf{schema}} & \textbf{Timeout} & \textbf{Other} \\
\midrule
\multicolumn{5}{@{}l}{\textit{Specialized APIs}} \\
\quad LlamaExtract Agentic Plus & 370 (100.0\%) & 0 & 0 & 0 \\
\quad LlamaExtract Agentic & 369 (99.7\%) & 0 & 0 & 1 \\
\quad LlamaExtract Cost-Effective & 368 (99.5\%) & 2 & 0 & 0 \\
\quad Datalab (Accurate + Balanced) & 347 (93.8\%) & 0 & 23 & 0 \\
\quad Extend (Max Context) & 342 (92.4\%) & 18 & 10 & 0 \\
\quad Reducto Deep Extract & 352 (95.1\%) & 18 & 0 & 0 \\
\addlinespace[2pt]
\multicolumn{5}{@{}l}{\textit{Coding Agents}} \\
\quad Codex 5.5 & 370 (100.0\%) & 0 & 0 & 0 \\
\quad Claude Code (Opus 4.8) & 336 (90.8\%) & 31 & 2 & 1 \\
\addlinespace[2pt]
\multicolumn{5}{@{}l}{\textit{OSS}} \\
\quad Gemma4 26B & 303 (81.9\%) & 26 & 41 & 0 \\
\quad Qwen3.6 35B & 353 (95.4\%) & 17 & 0 & 0 \\
\quad NuExtract3 & 356 (96.2\%) & 0 & 14 & 0 \\
\quad Lift 9B & 330 (89.2\%) & 17 & 15 & 8 \\
\addlinespace[2pt]
\multicolumn{5}{@{}l}{\textit{Commercial VLM}} \\
\quad GPT-5.4 Nano & 366 (98.9\%) & 4 & 0 & 0 \\
\quad Gemini 3.5 Flash & 327 (88.4\%) & 42 & 0 & 1 \\
\bottomrule
\end{tabular}
\caption{Document success and operational failures across all 370 benchmark documents. Parentheses report the success rate. Capacity/schema includes explicit schema, input/context, file-size, and output/truncation limits, together with otherwise-unclassified failures on T3.e large-schema documents.}
\label{tab:failure-modes}
\end{table}

\subsection{Quality--Cost Scaling Within Model Families}
\label{app:model-family-quality-cost}
\begin{figure}[H]
\centering
\includegraphics[width=\textwidth]{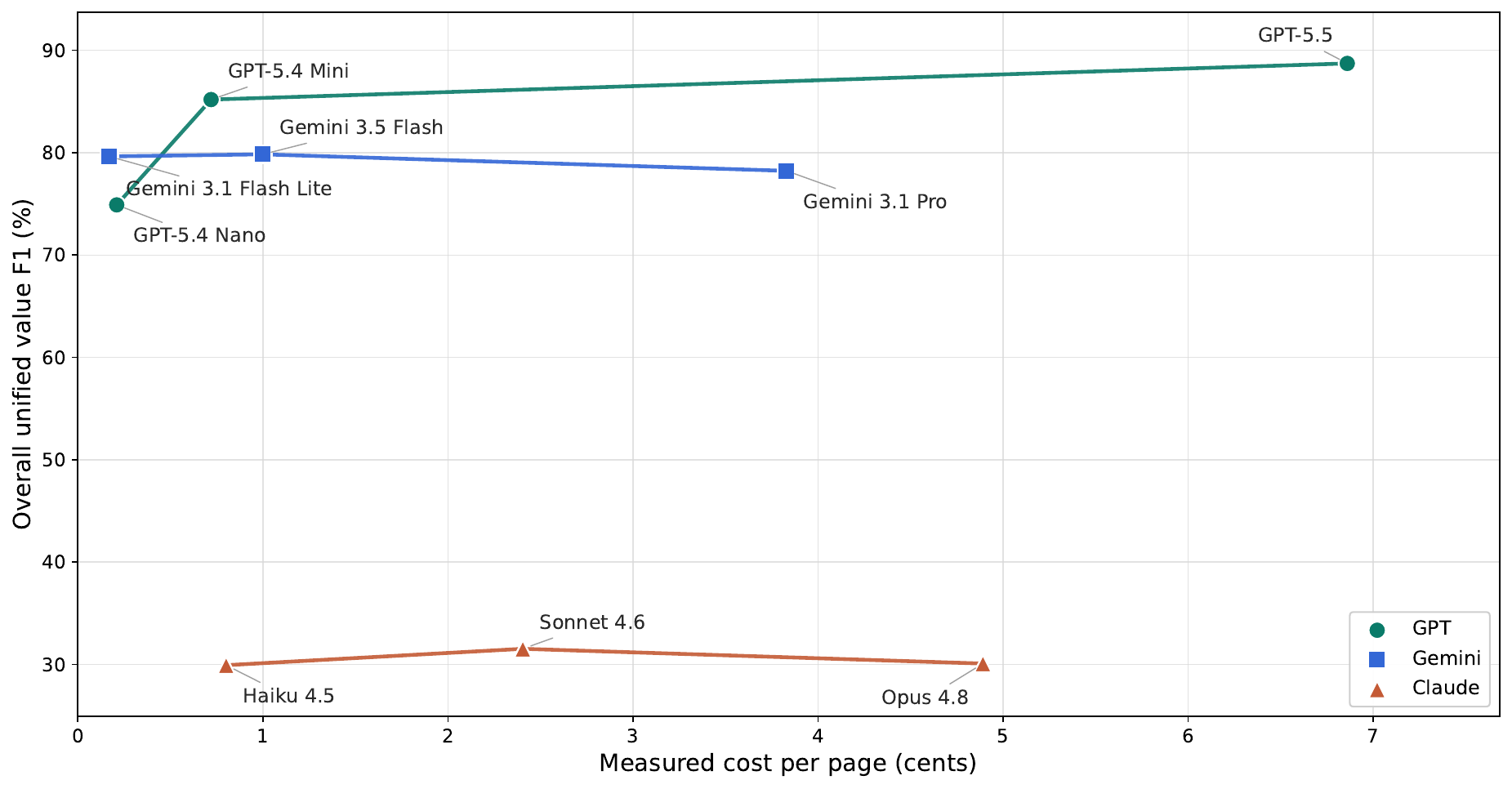}
\caption{Within-family quality--cost scaling for basic, medium, and flagship
models from GPT, Gemini, and Claude. Lines connect the three tiers within each
family. Quality is pooled over the full 370-document benchmark; rejected
documents score zero, while rejections that produce no model response contribute
zero measured extraction cost.}
\label{fig:model-family-quality-cost}
\end{figure}

We compare nine one-shot commercial VLM configurations: basic, medium, and
flagship models from the GPT, Gemini, and Claude families. Every configuration
receives the same document and target schema, and overall unified value F1 is
pooled over the same 370-document benchmark. \Cref{fig:model-family-quality-cost}
plots that quality against measured extraction cost per page; within each
family, the line connects the three model tiers in increasing order.

\paragraph{Quality and cost scaling.}
Scaling behavior differs sharply by family. GPT is the only family with
monotonic quality gains: GPT-5.4 Nano, GPT-5.4 Mini, and GPT-5.5 score 74.9,
85.2, and 88.7 F1 at 0.21, 0.72, and 6.86\,\textcent{} per page. Most of the
gain therefore comes from Nano to Mini; moving from Mini to GPT-5.5 costs
nearly ten times as much for another 3.5 F1 points. Gemini remains essentially
flat as cost rises, moving from 79.6 F1 at 0.17\,\textcent{} per page for Flash
Lite to 79.8 at 1.00\,\textcent{} for Flash and 78.2 at 3.83\,\textcent{} for
Pro. Claude is similarly non-monotonic: Haiku, Sonnet, and Opus score 29.9,
31.5, and 30.1 F1 as cost rises from 0.80 to 2.40 to
4.89\,\textcent{} per page. Within these nine configurations, the quality--cost
frontier is therefore Gemini 3.1 Flash Lite, GPT-5.4 Mini, and GPT-5.5: model
tier alone does not predict extraction quality, and the best operating point
depends on the desired cost--quality tradeoff.

\clearpage
\section{Qualitative Examples}
\label{app:qualitative}
Each example pairs a page crop with a field-by-field table of saved predictions.
For examples with ground-truth boxes, the source panel marks them in blue and
the system panels use \textcolor{qualok}{green} for a correct, grounded value,
\textcolor{qualpartial}{amber} for a correct but ungrounded value,
\textcolor{qualbad}{red} for an incorrect value, and \textcolor{qualnone}{gray}
for a value not returned. A dashed box shows the ground-truth location when the
system provides no usable citation; it is not treated as a predicted location.

Grounding requires a correct value and a citation at IoU~0.5
(\Cref{sec:metrics}). The T2 example has no box ground truth, so it compares
values only. All reported values come from the saved predictions.

\subsection{Long-list completeness (T1)}
\label{app:qual-t1}

\subsubsection*{T1: the first two records of a 250-party service list}

The selected crop contains the first two records from a 17-page service list with 250 parties and 1{,}457 values.

\begin{figure}[H]
\centering
\captionsetup[subfigure]{font=small,skip=2pt,justification=raggedright,singlelinecheck=false}
\begin{subfigure}[t]{1.0\textwidth}
\centering
\includegraphics[width=0.92\textwidth]{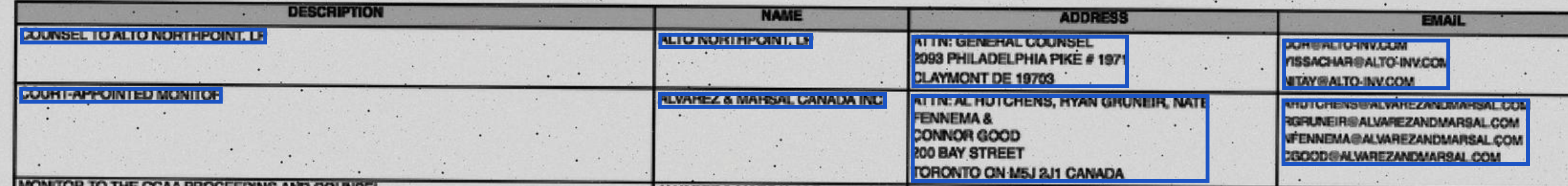}
\caption{\textbf{Ground-\allowbreak{}truth locations for the two records} --- 8 fields}
\label{fig:qual-t1-bbb-service-list-0}
\end{subfigure}
\vspace{2pt}

\begin{subfigure}[t]{0.49\textwidth}
\centering
\includegraphics[width=1.0\textwidth]{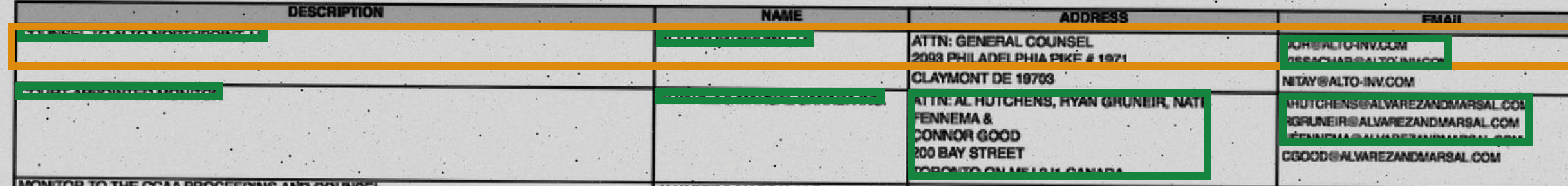}
\caption{\textbf{LlamaExtract Agentic Plus} --- 8/\allowbreak{}8 values correct, 7 grounded at IoU 0.\allowbreak{}5}
\label{fig:qual-t1-bbb-service-list-1}
\end{subfigure}
\hfill
\begin{subfigure}[t]{0.49\textwidth}
\centering
\includegraphics[width=1.0\textwidth]{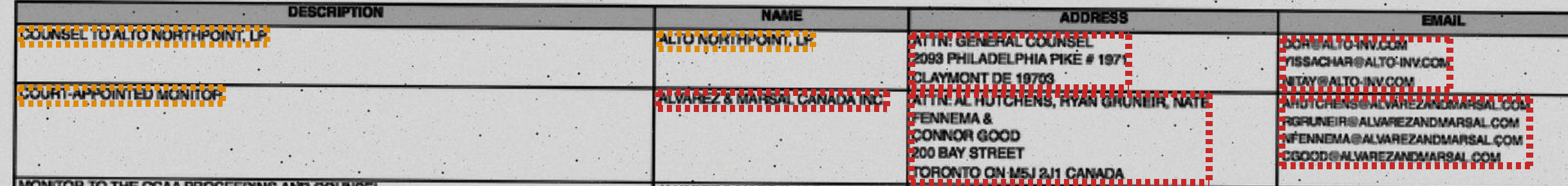}
\caption{\textbf{Lift 9B} --- 3/\allowbreak{}8 values correct; no source evidence returned}
\label{fig:qual-t1-bbb-service-list-2}
\end{subfigure}
\vspace{2pt}

\begin{subfigure}[t]{0.49\textwidth}
\centering
\includegraphics[width=1.0\textwidth]{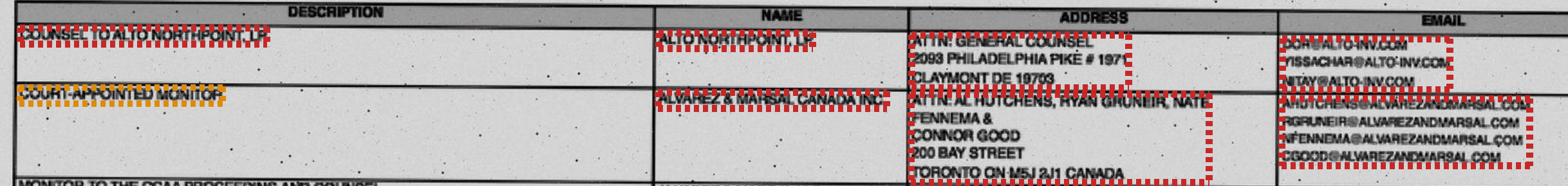}
\caption{\textbf{Codex GPT-\allowbreak{}5.\allowbreak{}5} --- 1/\allowbreak{}8 values correct; no source evidence returned}
\label{fig:qual-t1-bbb-service-list-3}
\end{subfigure}
\hfill
\begin{subfigure}[t]{0.49\textwidth}
\centering
\includegraphics[width=1.0\textwidth]{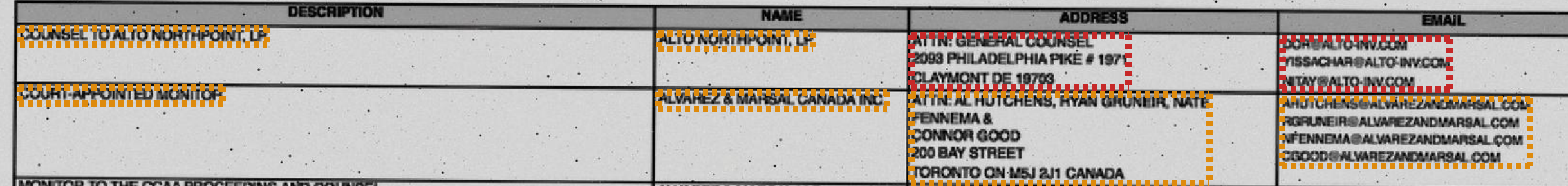}
\caption{\textbf{Extend Max Context} --- 6/\allowbreak{}8 values correct; no citations returned}
\label{fig:qual-t1-bbb-service-list-4}
\end{subfigure}
\vspace{2pt}

\caption{Two records from a degraded bankruptcy service list. LlamaExtract Agentic Plus returns all 8 values and grounds 7. Lift 9B misreads names and addresses; Codex GPT-5.5 removes spaces and corrupts e-mail addresses; Extend Max Context makes two character-level errors. None of the three returns a citation for these values.}
\label{fig:qual-t1-bbb-service-list}
\end{figure}

\begin{center}
\scriptsize
\setlength{\tabcolsep}{3pt}
\begin{tabularx}{\textwidth}{@{}L{1.15}|L{1.0}|L{1.0}|L{1.0}|L{1.0}|L{1.0}@{}}
\toprule
\textbf{Field} & \textbf{Ground truth} & \textbf{LlamaExtract A+} & \textbf{Lift 9B} & \textbf{Codex 5.\allowbreak{}5} & \textbf{Extend Max} \\
\midrule
party 1 · counsel for & COUNSEL TO ALTO NORTHPOINT, LP & \textcolor{qualok}{COUNSEL TO ALTO NORTHPOINT, LP} & \textcolor{qualpartial}{COUNSEL TO ALTO NORTHPOINT, LP} & \textcolor{qualbad}{COUNSELTOALT\allowbreak{}ONORTHPOINT,\allowbreak{}LP} & \textcolor{qualpartial}{COUNSEL TO ALTO NORTHPOINT, LP} \\
\midrule[0.02em]
party 1 · firm & ALTO NORTHPOINT, LP & \textcolor{qualok}{ALTO NORTHPOINT, LP} & \textcolor{qualpartial}{ALTO NORTHPOINT, LP} & \textcolor{qualbad}{ALTONORTHPOI\allowbreak{}NT,\allowbreak{}LP} & \textcolor{qualpartial}{ALTO NORTHPOINT, LP} \\
\midrule[0.02em]
party 1 · address & ATTN: GENERAL COUNSEL\newline 2093 PHILADELPHIA PIKE \# 1971\newline CLAYMONT DE 19703 & \textcolor{qualpartial}{ATTN: GENERAL COUNSEL\newline 2093 PHILADELPHIA PIKE \# 1971\newline CLAYMONT DE 19703} & \textcolor{qualbad}{ATTN: GENERAL COUNSEL 2093 PHILADELPHIA PKWY \# 1971 CLAYMONT DE 19703} & \textcolor{qualbad}{ATTN:\allowbreak{}GENERALCOUNS\allowbreak{}EL\newline 2093PHILADEL\allowbreak{}PHIAPIKE1971\newline CLAYMONTDE19\allowbreak{}703} & \textcolor{qualbad}{ATIN: GENERAL COUNSEL\newline 2093 PHILADELPHIA PIKE \# 1971\newline CLAYMONT DE 19703} \\
\midrule[0.02em]
party 1 · e-\allowbreak{}mail & DOR@\allowbreak{}ALTO-\allowbreak{}INV.\allowbreak{}COM\newline YISSACHAR@\allowbreak{}ALTO-\allowbreak{}INV.\allowbreak{}COM\newline NITAY@\allowbreak{}ALTO-\allowbreak{}INV.\allowbreak{}COM & \textcolor{qualok}{DOR@\allowbreak{}ALTO-\allowbreak{}INV.\allowbreak{}COM\newline YISSACHAR@\allowbreak{}ALTO-\allowbreak{}INV.\allowbreak{}COM\newline NITAY@\allowbreak{}ALTO-\allowbreak{}INV.\allowbreak{}COM} & \textcolor{qualbad}{DORI@\allowbreak{}ALTO-\allowbreak{}INV.\allowbreak{}COM YISSACHAR@\allowbreak{}ALTO-\allowbreak{}INV.\allowbreak{}COM NEWAY@\allowbreak{}ALTO-\allowbreak{}INV.\allowbreak{}COM} & \textcolor{qualbad}{DORALTOHNV.\allowbreak{}COM\newline YISSACHARGAL\allowbreak{}TO-\allowbreak{}INVCOM\newline NITAY@\allowbreak{}ALTO-\allowbreak{}INV.\allowbreak{}COM} & \textcolor{qualbad}{DOR@\allowbreak{}ALTO-\allowbreak{}INV.\allowbreak{}COM\newline YISSACHAR@\allowbreak{}ALTO-\allowbreak{}INV.\allowbreak{}COM\newline NITAY@\allowbreak{}ALTO.\allowbreak{}INV.\allowbreak{}COM} \\
\midrule[0.02em]
party 2 · counsel for & COURT-\allowbreak{}APPOINTED MONITOR & \textcolor{qualok}{COURT-\allowbreak{}APPOINTED MONITOR} & \textcolor{qualpartial}{COURT-\allowbreak{}APPOINTED MONITOR} & \textcolor{qualpartial}{COURT-\allowbreak{}APPOINTEDMON\allowbreak{}ITOR} & \textcolor{qualpartial}{COURT-\allowbreak{}APPOINTED MONITOR} \\
\midrule[0.02em]
party 2 · firm & ALVAREZ \& MARSAL CANADA INC. & \textcolor{qualok}{ALVAREZ \& MARSAL CANADA INC.} & \textcolor{qualbad}{ALVAREZ \& MARSAI CANADA INC.} & \textcolor{qualbad}{ALVAREZ\&MARS\allowbreak{}ALCANADAINC.} & \textcolor{qualpartial}{ALVAREZ \& MARSAL CANADA INC.} \\
\midrule[0.02em]
party 2 · address & ATTN: AL HUTCHENS, RYAN GRUNEIR, NATE FENNEMA \&\newline CONNOR GOOD\newline 200 BAY STREET\newline TORONTO ON M5J 2J1 CANADA & \textcolor{qualok}{ATTN: AL HUTCHENS, RYAN GRUNEIR, NATE FENNEMA \&\newline CONNOR GOOD\newline 200 BAY STREET\newline TORONTO ON M5J 2J1 CANADA} & \textcolor{qualbad}{ATTN: AL HUTCHENS, RYAN GRUNER, NATE PENNEMA \& CONNOR GODIC 200 BAY STREET TORONTO ON M5J 2J1 CANADA} & \textcolor{qualbad}{ATTN:\allowbreak{}ALHUTCHENS,\allowbreak{}RYANGRUNEIR,\allowbreak{}NATE\newline FENNEMA\&\newline CONNOR GOOD\newline 200BAYSTREET\newline TORONTOON·M5\allowbreak{}J2J1CANADA} & \textcolor{qualpartial}{ATTN: AL HUTCHENS, RYAN GRUNEIR, NATE FENNEMA \& CONNOR GOOD\newline 200 BAY STREET\newline TORONTO ON M5J 2J1 CANADA} \\
\midrule[0.02em]
party 2 · e-\allowbreak{}mail & AHUTCHENS@\allowbreak{}ALVAREZANDMA\allowbreak{}RSAL.\allowbreak{}COM\newline RGRUNEIR@\allowbreak{}ALVAREZANDMA\allowbreak{}RSAL.\allowbreak{}COM\newline NFENNEMA@\allowbreak{}ALVAREZANDMA\allowbreak{}RSAL.\allowbreak{}COM\newline CGOOD@\allowbreak{}ALVAREZANDMA\allowbreak{}RSAL.\allowbreak{}COM & \textcolor{qualok}{AHUTCHENS@\allowbreak{}ALVAREZANDMA\allowbreak{}RSAL.\allowbreak{}COM\newline RGRUNEIR@\allowbreak{}ALVAREZANDMA\allowbreak{}RSAL.\allowbreak{}COM\newline NFENNEMA@\allowbreak{}ALVAREZANDMA\allowbreak{}RSAL.\allowbreak{}COM\newline CGOOD@\allowbreak{}ALVAREZANDMA\allowbreak{}RSAL.\allowbreak{}COM} & \textcolor{qualbad}{AHUTCHENS@\allowbreak{}ALVAREZANDMA\allowbreak{}RSAI.\allowbreak{}COM RGRUNER@\allowbreak{}ALVAREZANDMA\allowbreak{}RSAI.\allowbreak{}COM NPENNEMA@\allowbreak{}ALVAREZANDMA\allowbreak{}RSAI.\allowbreak{}COM CGODIC@\allowbreak{}ALVAREZANDMA\allowbreak{}RSAI.\allowbreak{}COM} & \textcolor{qualbad}{AHUTCHENSOAL\allowbreak{}WAREZANOMARS\allowbreak{}ALOOM\newline RGRUNEIRSALV\allowbreak{}AFEZANDMARSA\allowbreak{}LCOM\newline NFENNEMAGALW\allowbreak{}AEZANDMARSAL\allowbreak{}COM\newline CGOODGALWARE\allowbreak{}ZANDMARSALCO\allowbreak{}M} & \textcolor{qualpartial}{AHUTCHENS@\allowbreak{}ALVAREZANDMA\allowbreak{}RSAL.\allowbreak{}COM\newline RGRUNEIR@\allowbreak{}ALVAREZANDMA\allowbreak{}RSAL.\allowbreak{}COM\newline NFENNEMA@\allowbreak{}ALVAREZANDMA\allowbreak{}RSAL.\allowbreak{}COM\newline CGOOD@\allowbreak{}ALVAREZANDMA\allowbreak{}RSAL.\allowbreak{}COM} \\
\bottomrule
\end{tabularx}
\captionof{table}{Exact outputs for the eight fields in \Cref{fig:qual-t1-bbb-service-list}. Multi-line addresses and e-mail blocks are shown line by line.}
\label{tab:qual-t1-bbb-service-list}
\end{center}

\subsection{Needle-in-haystack (T2)}
\label{app:qual-t2}

\subsubsection*{T2: three claims inside a Medi-Cal remittance advice}

The selected block comes from a one-page Medi-Cal remittance advice whose headings and data share the same fixed-width layout.

\begin{figure}[H]
\centering
\includegraphics[width=1.0\textwidth]{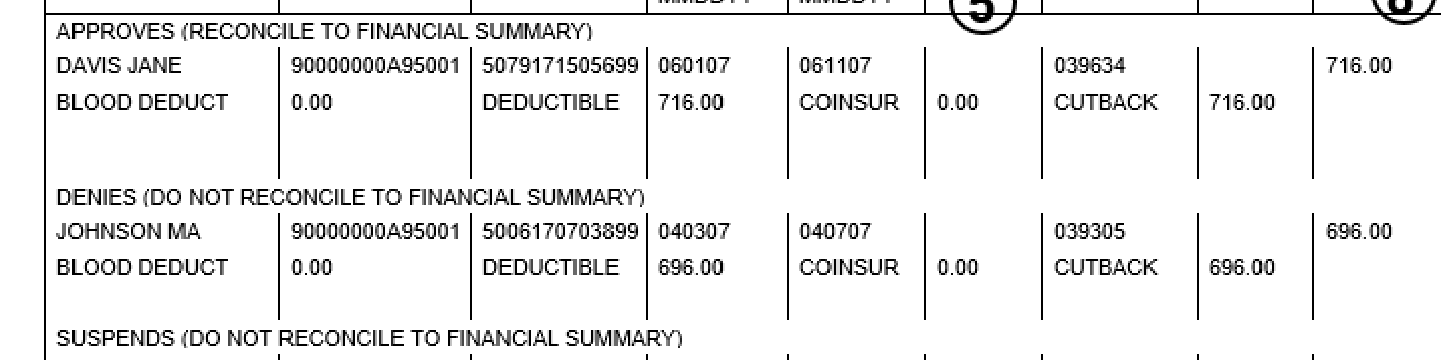}
\caption{Three claims and their service lines. LlamaExtract Agentic Plus returns all 12 selected values. GPT-5.4 Nano treats headings and header fields as data; Claude Code Opus 4.8 omits the allowed amounts; Datalab copies each claim status into a blank service-line status field.}
\label{fig:qual-t2-carad-remittance}
\end{figure}

\begin{center}
\scriptsize
\setlength{\tabcolsep}{3pt}
\begin{tabularx}{\textwidth}{@{}L{1.15}|L{1.0}|L{1.0}|L{1.0}|L{1.0}|L{1.0}@{}}
\toprule
\textbf{Field} & \textbf{Ground truth} & \textbf{LlamaExtract A+} & \textbf{GPT-\allowbreak{}5.\allowbreak{}4 Nano} & \textbf{Claude Code 4.\allowbreak{}8} & \textbf{Datalab} \\
\midrule
claim 1 · status & APPROVES & \textcolor{qualok}{APPROVES} & \textcolor{qualbad}{DENIES (DO NOT RECONCILE TO FINANC…} & \textcolor{qualok}{APPROVES} & \textcolor{qualok}{APPROVES} \\
\midrule[0.02em]
claim 1 · patient name & DAVIS JANE & \textcolor{qualok}{DAVIS JANE} & \textcolor{qualbad}{RECIPIENT NAME} & \textcolor{qualok}{DAVIS JANE} & \textcolor{qualok}{DAVIS JANE} \\
\midrule[0.02em]
claim 1 · recipient id & 90000000A950\allowbreak{}01 & \textcolor{qualok}{90000000A950\allowbreak{}01} & \textcolor{qualbad}{012345678901} & \textcolor{qualok}{90000000A950\allowbreak{}01} & \textcolor{qualok}{90000000A950\allowbreak{}01} \\
\midrule[0.02em]
claim 1 · claim number & 507917150569\allowbreak{}9 & \textcolor{qualok}{507917150569\allowbreak{}9} & \textcolor{qualbad}{039248026} & \textcolor{qualok}{507917150569\allowbreak{}9} & \textcolor{qualok}{507917150569\allowbreak{}9} \\
\midrule[0.02em]
line 1 · allowed \$ & 716 & \textcolor{qualok}{716.\allowbreak{}0} & \textcolor{qualok}{716} & \textcolor{qualnone}{\emph{not returned}} & \textcolor{qualok}{716.\allowbreak{}0} \\
\midrule[0.02em]
line 1 · status & None & \textcolor{qualok}{None} & \textcolor{qualbad}{DENIES (DO NOT RECONCILE TO FINANC…} & \textcolor{qualok}{None} & \textcolor{qualbad}{APPROVES} \\
\midrule[0.02em]
claim 2 · status & DENIES & \textcolor{qualok}{DENIES} & \textcolor{qualbad}{DENIES (DO NOT RECONCILE TO FINANC…} & \textcolor{qualok}{DENIES} & \textcolor{qualok}{DENIES} \\
\midrule[0.02em]
claim 2 · patient name & JOHNSON MA & \textcolor{qualok}{JOHNSON MA} & \textcolor{qualbad}{RECIPIENT NAME} & \textcolor{qualok}{JOHNSON MA} & \textcolor{qualok}{JOHNSON MA} \\
\midrule[0.02em]
claim 2 · claim number & 500617070389\allowbreak{}9 & \textcolor{qualok}{500617070389\allowbreak{}9} & \textcolor{qualbad}{040707} & \textcolor{qualok}{500617070389\allowbreak{}9} & \textcolor{qualok}{500617070389\allowbreak{}9} \\
\midrule[0.02em]
line 2 · allowed \$ & 696 & \textcolor{qualok}{696.\allowbreak{}0} & \textcolor{qualbad}{716} & \textcolor{qualnone}{\emph{not returned}} & \textcolor{qualok}{696.\allowbreak{}0} \\
\midrule[0.02em]
line 2 · status & None & \textcolor{qualok}{None} & \textcolor{qualbad}{DENIES (DO NOT RECONCILE TO FINANC…} & \textcolor{qualok}{None} & \textcolor{qualbad}{DENIES} \\
\midrule[0.02em]
claim 3 · status & SUSPENDS & \textcolor{qualok}{SUSPENDS} & \textcolor{qualbad}{SUSPENDS (DO NOT RECONCILE TO FINA…} & \textcolor{qualok}{SUSPENDS} & \textcolor{qualok}{SUSPENDS} \\
\bottomrule
\end{tabularx}
\captionof{table}{Exact outputs for 12 of the page's 45 fields. This document has no box ground truth, so grounding is not scored.}
\label{tab:qual-t2-carad-remittance}
\end{center}

\subsection{Dense documents (T3)}
\label{app:qual-t3}

\subsubsection*{T3: the location and operations lines of a Railroad Commission Form P-18}

The selected crop contains location and operations fields from a scanned Form P-18, where most values align to dotted leaders rather than ruled cells.

\begin{figure}[H]
\centering
\captionsetup[subfigure]{font=small,skip=2pt,justification=raggedright,singlelinecheck=false}
\begin{subfigure}[t]{1.0\textwidth}
\centering
\includegraphics[width=1.0\textwidth]{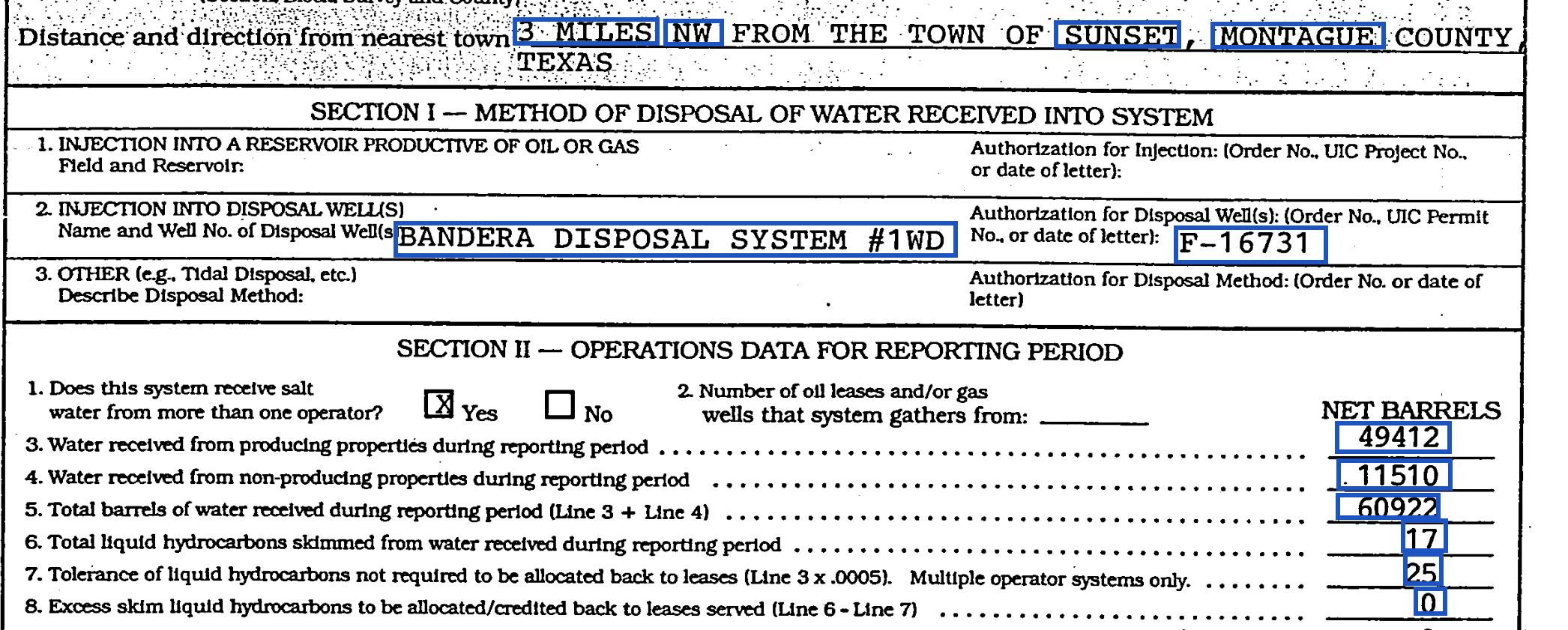}
\caption{\textbf{Ground-\allowbreak{}truth locations in the selected blocks} --- 12 fields}
\label{fig:qual-t3-p18-skim-report-0}
\end{subfigure}
\vspace{2pt}

\begin{subfigure}[t]{0.49\textwidth}
\centering
\includegraphics[width=1.0\textwidth]{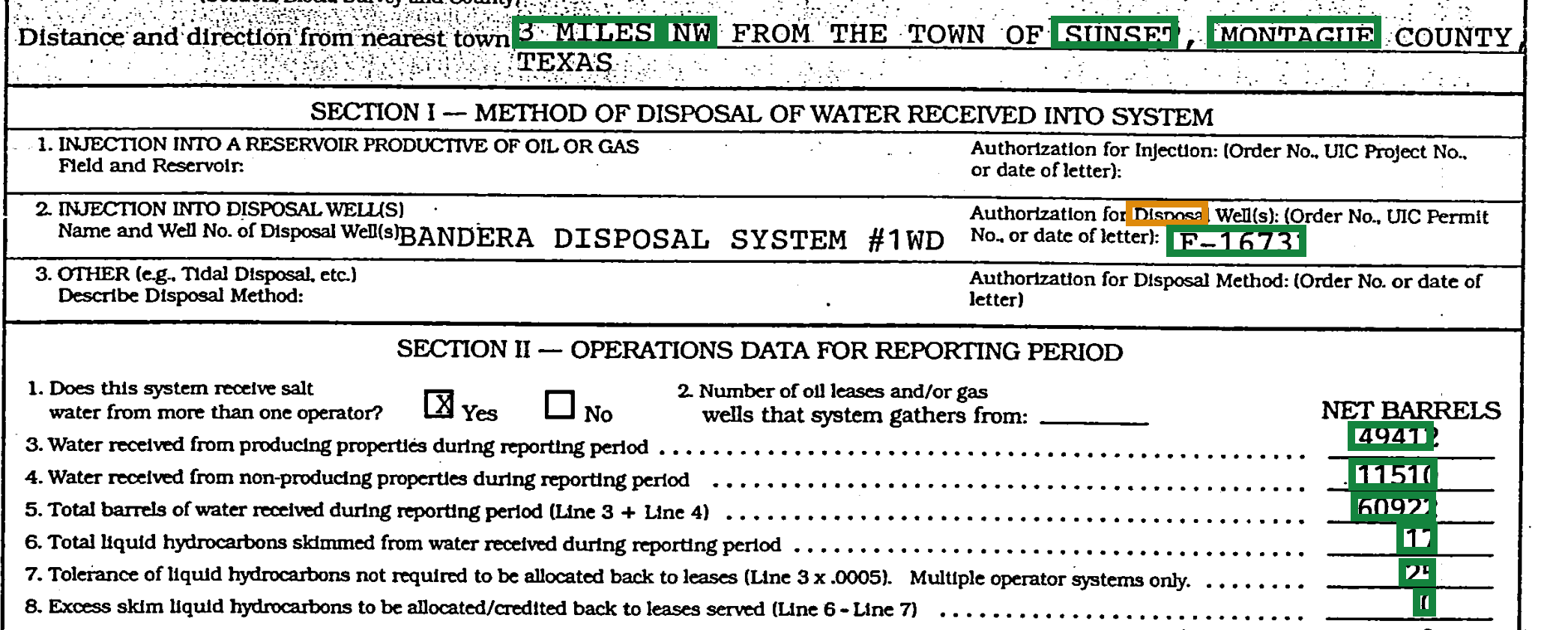}
\caption{\textbf{LlamaExtract Agentic Plus} --- 12/\allowbreak{}12 values correct, 11 grounded at IoU 0.\allowbreak{}5}
\label{fig:qual-t3-p18-skim-report-1}
\end{subfigure}
\hfill
\begin{subfigure}[t]{0.49\textwidth}
\centering
\includegraphics[width=1.0\textwidth]{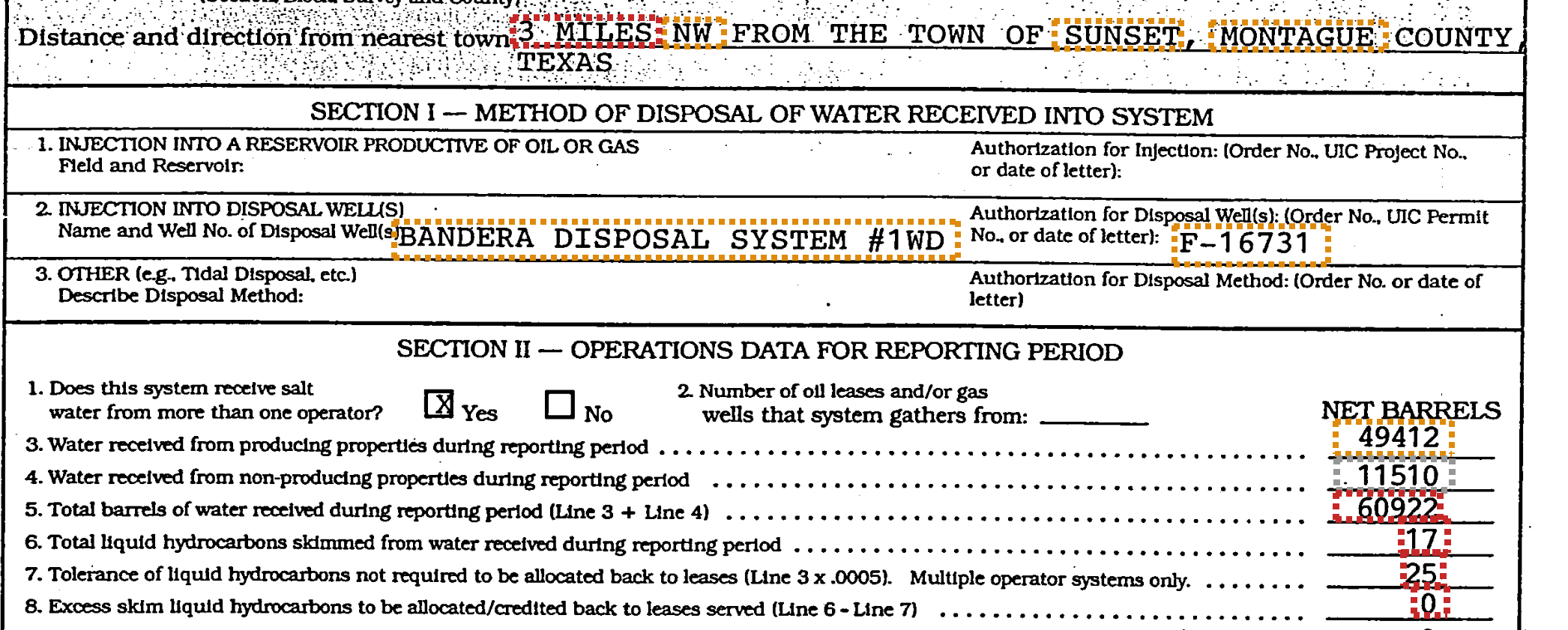}
\caption{\textbf{Gemma4 26B} --- 6/\allowbreak{}12 values correct; no source evidence returned; 1 not returned (gray)}
\label{fig:qual-t3-p18-skim-report-2}
\end{subfigure}
\vspace{2pt}

\begin{subfigure}[t]{0.49\textwidth}
\centering
\includegraphics[width=1.0\textwidth]{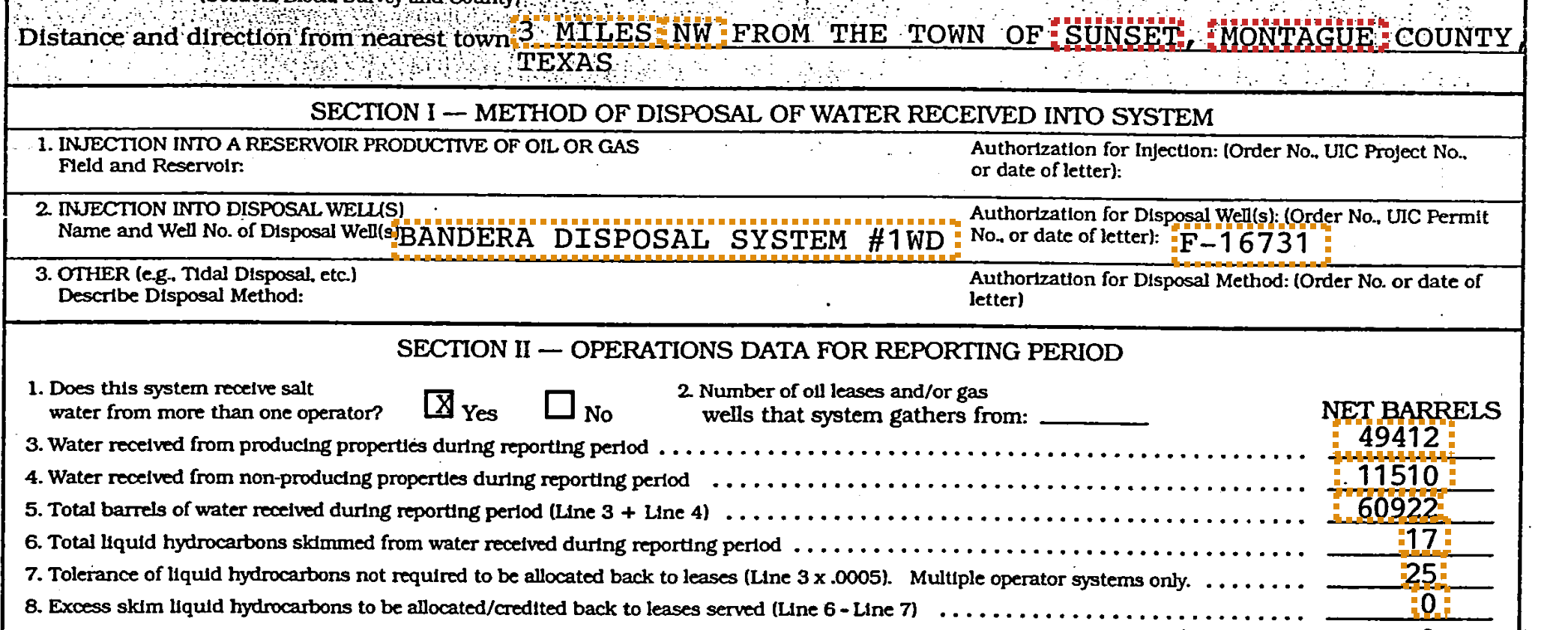}
\caption{\textbf{Codex GPT-\allowbreak{}5.\allowbreak{}5} --- 10/\allowbreak{}12 values correct; no source evidence returned}
\label{fig:qual-t3-p18-skim-report-3}
\end{subfigure}
\hfill
\begin{subfigure}[t]{0.49\textwidth}
\centering
\includegraphics[width=1.0\textwidth]{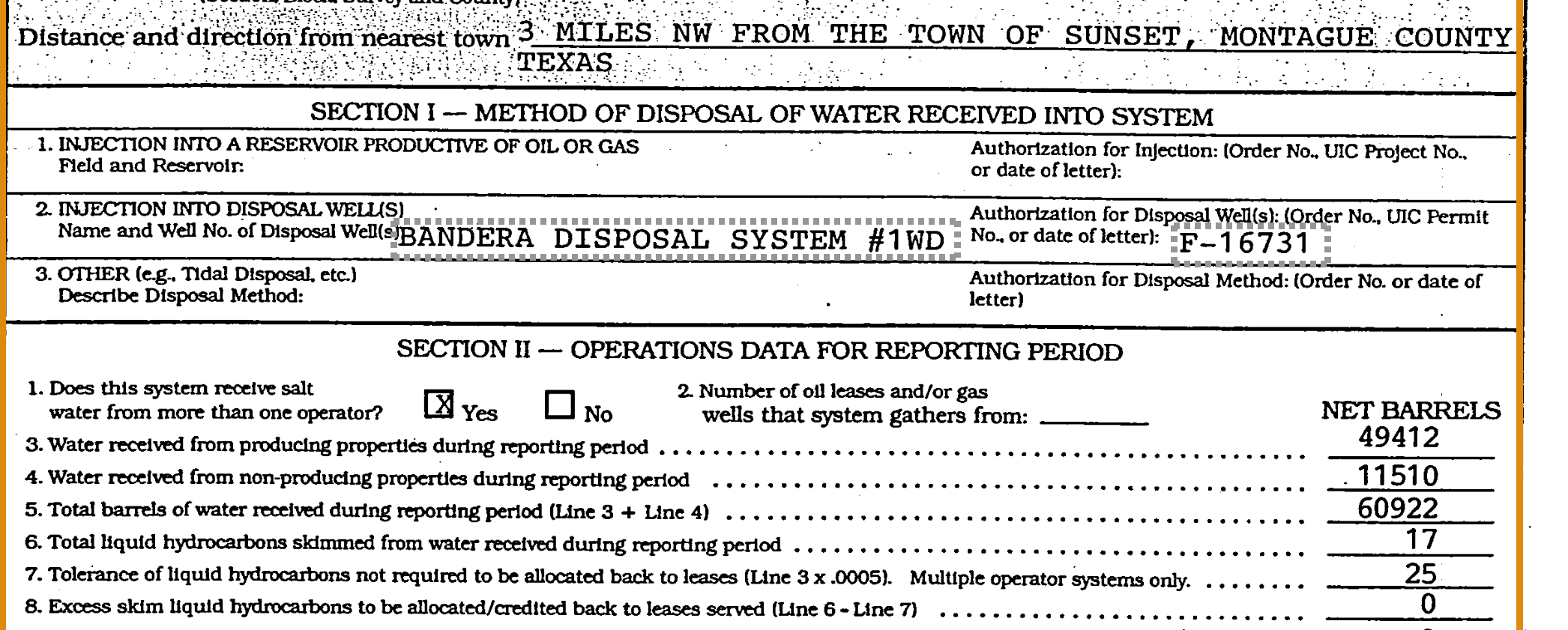}
\caption{\textbf{Datalab Accurate + Balanced} --- 10/\allowbreak{}12 values correct, 0 grounded at IoU 0.\allowbreak{}5; 2 not returned (gray)}
\label{fig:qual-t3-p18-skim-report-4}
\end{subfigure}
\vspace{2pt}

\caption{Location and operations fields from a typewritten Form P-18. LlamaExtract Agentic Plus returns all 12 values and grounds 11. Gemma4 26B shifts values down the dotted-leader column; Codex GPT-5.5 merges the location phrase into the town and county fields; Datalab omits the disposal well and permit number. None of the three grounds a value in this crop.}
\label{fig:qual-t3-p18-skim-report}
\end{figure}

\begin{center}
\scriptsize
\setlength{\tabcolsep}{3pt}
\begin{tabularx}{\textwidth}{@{}L{1.15}|L{1.0}|L{1.0}|L{1.0}|L{1.0}|L{1.0}@{}}
\toprule
\textbf{Field} & \textbf{Ground truth} & \textbf{LlamaExtract A+} & \textbf{Gemma4 26B} & \textbf{Codex 5.\allowbreak{}5} & \textbf{Datalab} \\
\midrule
distance from town & 3 MILES & \textcolor{qualok}{3 MILES} & \textcolor{qualbad}{3} & \textcolor{qualpartial}{3 MILES} & \textcolor{qualpartial}{3 MILES} \\
\midrule[0.02em]
direction from town & NW & \textcolor{qualok}{NW} & \textcolor{qualpartial}{NW} & \textcolor{qualpartial}{NW} & \textcolor{qualpartial}{NW} \\
\midrule[0.02em]
nearest town & SUNSET & \textcolor{qualok}{SUNSET} & \textcolor{qualpartial}{SUNSET} & \textcolor{qualbad}{SUNSET, MONTAGUE COUNTY, TEXAS} & \textcolor{qualpartial}{SUNSET} \\
\midrule[0.02em]
county & MONTAGUE & \textcolor{qualok}{MONTAGUE} & \textcolor{qualpartial}{MONTAGUE} & \textcolor{qualbad}{MONTAGUE CO.\allowbreak{}, TX} & \textcolor{qualpartial}{MONTAGUE} \\
\midrule[0.02em]
disposal well name & BANDERA DISPOSAL SYSTEM \#1WD & \textcolor{qualpartial}{BANDERA DISPOSAL SYSTEM \#1WD} & \textcolor{qualpartial}{BANDERA DISPOSAL SYSTEM \#1WD} & \textcolor{qualpartial}{BANDERA DISPOSAL SYSTEM \#1WD} & \textcolor{qualnone}{\emph{not returned}} \\
\midrule[0.02em]
UIC permit no. & F-\allowbreak{}16731 & \textcolor{qualok}{F-\allowbreak{}16731} & \textcolor{qualpartial}{F-\allowbreak{}16731} & \textcolor{qualpartial}{F-\allowbreak{}16731} & \textcolor{qualnone}{\emph{not returned}} \\
\midrule[0.02em]
water · producing & 49412 & \textcolor{qualok}{49412} & \textcolor{qualpartial}{49412} & \textcolor{qualpartial}{49412} & \textcolor{qualpartial}{49412} \\
\midrule[0.02em]
water · non-\allowbreak{}producing & 11510 & \textcolor{qualok}{11510} & \textcolor{qualnone}{\emph{not returned}} & \textcolor{qualpartial}{11510} & \textcolor{qualpartial}{11510} \\
\midrule[0.02em]
water · total received & 60922 & \textcolor{qualok}{60922} & \textcolor{qualbad}{11510} & \textcolor{qualpartial}{60922} & \textcolor{qualpartial}{60922} \\
\midrule[0.02em]
hydrocarbons skimmed & 17 & \textcolor{qualok}{17} & \textcolor{qualbad}{60922} & \textcolor{qualpartial}{17} & \textcolor{qualpartial}{17} \\
\midrule[0.02em]
tolerance & 25 & \textcolor{qualok}{25} & \textcolor{qualbad}{17} & \textcolor{qualpartial}{25} & \textcolor{qualpartial}{25} \\
\midrule[0.02em]
excess back to leases & 0 & \textcolor{qualok}{0} & \textcolor{qualbad}{25} & \textcolor{qualpartial}{0} & \textcolor{qualpartial}{0} \\
\bottomrule
\end{tabularx}
\captionof{table}{Exact outputs for 12 of the form's 76 fields.}
\label{tab:qual-t3-p18-skim-report}
\end{center}

\end{document}